\pdfoutput=1

\documentclass[11pt]{article}
\usepackage[final]{acl}
\usepackage{adjustbox}
\usepackage{booktabs}
\usepackage{times}
\usepackage{latexsym}
\usepackage{algorithm}
\usepackage{algorithmic}
\usepackage{amsmath}
\usepackage{multirow}
\usepackage{multicol}
\usepackage{array}
\usepackage{soul}
\usepackage[T1]{fontenc}

\usepackage[utf8]{inputenc}

\usepackage{microtype}

\usepackage{inconsolata}

\usepackage{graphicx}
\definecolor{deepgreen}{rgb}{0.0, 0.5, 0.0}

\title{CoT-RAG: Integrating Chain of Thought and Retrieval-Augmented Generation to Enhance Reasoning in Large Language Models}

\author{
  \textbf{Feiyang Li\textsuperscript{1}},
  \textbf{Peng Fang\textsuperscript{1,2}\thanks{Corresponding author}},
  \textbf{Zhan Shi\textsuperscript{1,2}},
  \textbf{Arijit Khan\textsuperscript{3,4}},\\
  \textbf{Fang Wang\textsuperscript{1,2}},
  \textbf{Weihao Wang\textsuperscript{5}},
  \textbf{Xin Zhang\textsuperscript{5}},
  \textbf{Yongjian Cui\textsuperscript{5}},\\
  \textsuperscript{1}School of Computer Science and Technology, Huazhong University of Science and Technology, Wuhan, China \\
  \textsuperscript{2}Wuhan National Laboratory for Optoelectronics, Huazhong University of Science and Technology, Wuhan, China \\
  \textsuperscript{3}Department of Computer Science, Bowling Green State University, Ohio, USA \\
  \textsuperscript{4}Department of Computer Science, Aalborg University, Aalborg, Denmark \\
  \textsuperscript{5}Huawei Technologies Co., Ltd, Shenzhen, China\\
}

\begin{document}
\maketitle
\begin{abstract}
Chain-of-thought (CoT) reasoning boosts large language models' (LLMs) performance on complex tasks but faces two key limitations: a lack of reliability when solely relying on LLM-generated reasoning chains and lower reasoning performance from natural language prompts compared with code prompts. To address these issues, we propose CoT-RAG, a novel reasoning framework with three key designs: (i) \textit{Knowledge Graph-driven CoT Generation}, 
featuring knowledge graphs to modulate reasoning chain generation of LLMs, thereby enhancing reasoning  credibility; (ii) \textit{Learnable Knowledge Case-aware RAG}, which incorporates retrieval-augmented generation (RAG) into knowledge graphs to retrieve relevant sub-cases and sub-descriptions, providing LLMs with learnable information; (iii) \textit{Pseudo-Program Prompting Execution}, which promotes greater logical rigor by guiding LLMs to execute reasoning tasks as pseudo-programs. Evaluations on nine public datasets spanning three reasoning tasks reveal significant accuracy gains—ranging from 4.0\% to 44.3\%--over state-of-the-art methods. Furthermore, tests on four domain-specific datasets demonstrate exceptional accuracy and efficient execution, underscoring its practical applicability and scalability. Our code and data are available at \url{https://github.com/hustlfy123/CoT-RAG}.
\end{abstract}

\section{Introduction}

Large language models (LLMs) have garnered significant attention in both academia and industry due to their exceptional performance in natural language processing (NLP) \citep{kabir-etal-2024-benllm,chiang-lee-2023-large}, such as machine translation \citep{zhu-etal-2024-multilingual,10.1145/3639631.3639658}, text summarization \citep{li-etal-2024-active} and sentiment analysis \citep{bai-etal-2024-compound}. 
Nonetheless, they exhibit notable limitations in complex tasks requiring arithmetic, commonsense, and symbolic reasoning \citep{wei2022chain,gao2023pal}.
To overcome these challenges, chain-of-thought (CoT) reasoning has been introduced, wherein LLMs explicitly generate intermediate reasoning steps prior to reaching a conclusion \citep{chu-etal-2024-navigate}. Several CoT variants have emerged, including Manual-CoT \citep{wei2022chain}, Zero-shot-CoT \citep{kojima2022large}, PoT \citep{chen2023program}, and PS \citep{wang-etal-2023-plan}.

\begin{figure*}[t]
  \centering
  \includegraphics[width=0.9\textwidth]{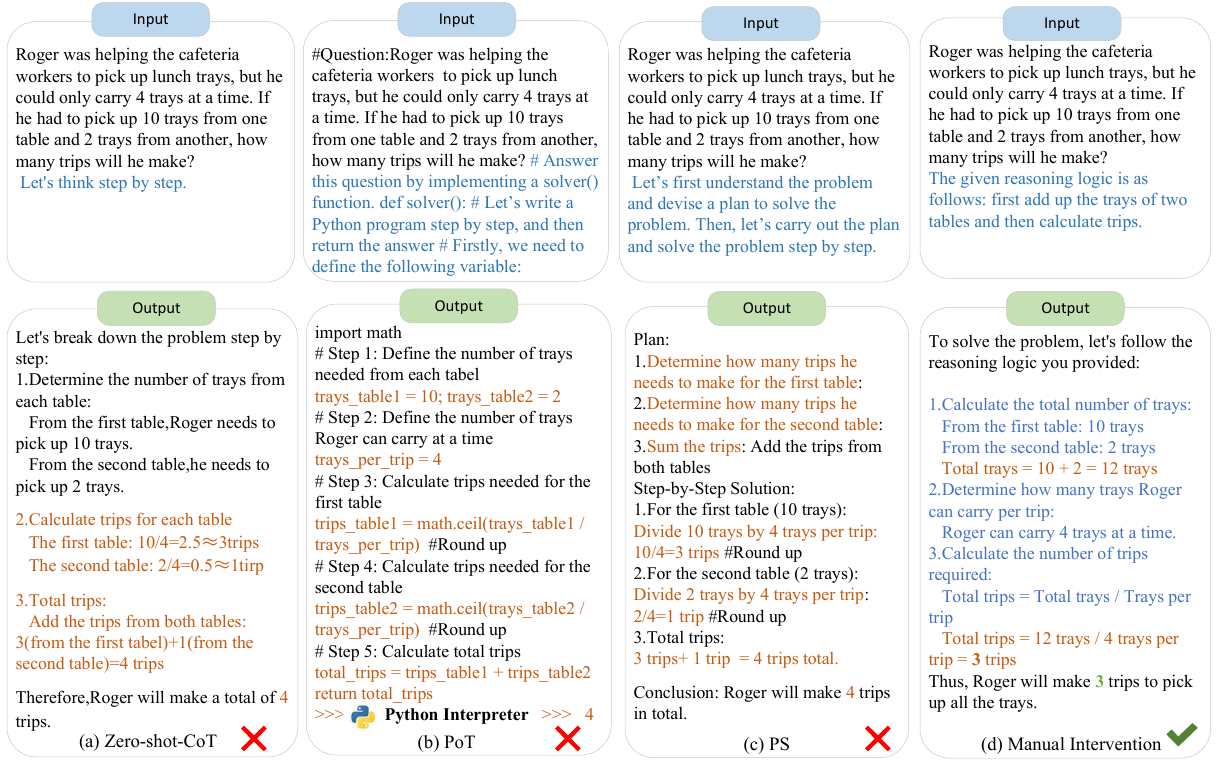}
  \caption{Example inputs and outputs of GPT-4o mini with (a) Zero-shot-CoT \citep{kojima2022large}, (b) PoT \citep{chen2023program}, (c) PS \citep{wang-etal-2023-plan} and (d) Manual Intervention on MultiArith \citep{roy-roth-2015-solving}.}
  \label{Chains}
\end{figure*}

Although CoT improves performance on multi-step reasoning tasks for LLMs, existing methods still face two main challenges:

{\bf (1) The low reliability of relying solely on LLMs to generate reasoning chains}. 
Most existing CoT reasoning methods rely on prompting strategies, such as prompt format \citep{gao2023pal,chen2023program} and prompt planning \citep{yao2023react,wang-etal-2023-plan}, to guide LLMs in generating reasoning chains. However, due to the inherent black-box nature \citep{kim-etal-2024-learning} and hallucination issues \citep{manakul-etal-2023-selfcheckgpt} of LLMs, the generated reasoning steps often contain logical errors or factual inaccuracies \citep{10.1145/3637528.3672022}.
For instance, Manual-CoT, Zero-shot-CoT, and PS achieve average accuracies of only 48.4\%, 38.9\%, and 42.5\%, respectively, on the AQuA arithmetic reasoning dataset \citep{ling-etal-2017-program}, highlighting their generally low performance \citep{wang-etal-2023-plan}.
In vertical domains, e.g., law, medicine, and finance--where errors may compromise human life and critical assets--the inherent unreliability of LLM outputs introduces unquantifiable risks. Consequently, solely depending on LLM-generated reasoning cannot ensure reliable and safe outcomes.

{\bf (2) The poorer reasoning performance from natural language prompts compared with code prompts.} 
LLMs typically generate reasoning chains in natural language (NL) to describe intermediate steps.
While these NL-based reasoning chains are intuitive for humans, they exhibit lower reasoning accuracies compared with code prompts \citep{gao2023pal,NEURIPS2023_b5c8c1c1}.
\citet{lyu-etal-2023-faithful} demonstrate that on the GSM8K arithmetic reasoning dataset \citep{2021arXiv211014168C}, Manual-CoT and LtM \citep{zhou2023leasttomost}, which rely on NL-based reasoning chains, achieve average accuracies of 46.6\% and 42.3\%, respectively, exhibiting inadequate correctness; while Faithful CoT \citep{lyu-etal-2023-faithful} that employs a two-step process where LLMs first generate a symbolic reasoning chain (e.g., in Python) and then execute this chain using an external interpreter (e.g., a Python interpreter), achieves an average accuracy of 64.2\%. This underscores that LLMs which use NL prompts achieve inferior reasoning performance compared with code prompts.

To address these challenges, we propose CoT-RAG, an enhanced reasoning framework with three key components. 
{\bf Knowledge Graph-driven CoT Generation:}
To ensure human safety and reduce potential risks in vertical domains, our approach places domain experts at the helm while LLMs augment their capabilities. In particular, experts supply a one-time, coarse-grained decision tree (DT) that encapsulates the underlying reasoning logic for the domain, independent of individual user queries. The LLMs then convert this DT into a detailed knowledge graph (KG) to enhance comprehension. For each user query, the KG is employed to generate reasoning chains by LLMs, thereby improving the process's controllability, reliability, and adaptability to specific domains.
{\bf Learnable Knowledge Case-aware RAG:}
CoT-RAG integrates Retrieval-Augmented Generation into the knowledge graph, retrieving relevant sub-cases and sub-descriptions that supply learnable data to mitigate LLMs' inherent logical errors and factual biases. Moreover, the interactive framework between LLMs and the KG enables dynamic updates to both the graph structure and the case repository.
{\bf Pseudo-Program Prompting Execution:}
CoT-RAG employs pseudo-program prompting as an alternative to NL prompts, which directs LLMs to execute reasoning tasks via pseudo-programmatic chains, retaining the versatility of NL prompts while improving logical coherence in complex tasks.

We evaluated CoT-RAG on nine public datasets spanning three reasoning tasks. Results indicate that it outperforms existing methods, with accuracy improvements from 4.0\% to 44.3\%. Additionally, tests on four domain-specific datasets confirm its high accuracy and efficient execution, underscoring its scalable cross-domain performance.





\section{Motivation}
\subsection{Reasoning Chain Generation}

Generally, CoT reasoning relies on prompt strategies 
to guide LLMs in generating reasoning chains.
However, excessive reliance on LLMs may lead to erroneous reasoning steps, which pose incalculable risks and potential losses in fields related to human life and property due to the tendency of LLMs to make reasoning errors in vertical domains \citep{10.1109/MNET.2024.3435752,article}.

To examine this limitation, we compare methods that exclusively depend on LLM-generated reasoning chains, including Zero-shot-CoT \citep{kojima2022large}, PoT \citep{chen2023program}, and PS \citep{wang-etal-2023-plan}, against manual intervention method (i.e., providing explicit reasoning logic to LLMs).
As shown in Figure \ref{Chains}, 
we input a mathematical problem from the MultiArith dataset \citep{roy-roth-2015-solving} into GPT-4o mini \citep{openai2024gpt4omini}, where Zero-shot-CoT, PoT, and PS produce incorrect answers. In contrast, manual intervention in the reasoning logic (Figure \ref{Chains}(d)) allows the LLM to generate correct results.
Similarly, as reported by \citet{wang-etal-2023-plan}, Manual-CoT, Zero-shot-CoT, and PS achieve average accuracies of 74.8\%, 64.5\%, and 68.7\% on commonsense reasoning tasks, and 48.4\%, 38.9\%, and 42.5\% on the AQuA arithmetic reasoning dataset \citep{ling-etal-2017-program}, indicating the generally lower performance of such methods.
These findings further support our observations: Solely relying on LLM-generated reasoning chains is insufficient to ensure reliable results, highlighting the necessity of manual intervention in the reasoning process.
While some works \citep{gou2024critic,weng-etal-2023-large,paul-etal-2024-refiner,madaan2023selfrefine} have integrated verification and refinement into LLMs to reduce errors, they still depend solely on LLMs for evaluation and verification, which results in low accuracy \citep{chu-etal-2024-navigate}.

\subsection{Reasoning Chain Execution}
LLMs commonly produce reasoning chains in natural language (NL) to outline intermediate steps. Although these NL-based chains are human-friendly, they exhibit lower reasoning accuracies \citep{NEURIPS2023_b5c8c1c1}. As stated in \citet{lyu-etal-2023-faithful}, the average accuracies of Manual-CoT and LtM \citep{zhou2023leasttomost}, which execute NL reasoning chains on GSM8K \citep{2021arXiv211014168C}, are 46.6\% and 42.3\%, respectively, exhibiting their low performance; while Faithful CoT \citep{lyu-etal-2023-faithful} requires the LLM to first generate a symbolic reasoning chain (e.g., in Python) and then execute it using an external interpreter, achieving an average accuracy of 64.2\%. 
Therefore, methods like ProgPrompt \citep{singh2022progprompt}, Code as Policies \citep{10160591}, and AdaPlanner \citep{NEURIPS2023_b5c8c1c1} use code prompts instead of NL prompts to reduce ambiguities and improve inference performance. However, Code prompts exhibit three primary limitations. {\bf Complexity:} They use intricate programming symbols and function calls that are often unintelligible to non-programmers.
{\bf Scope:} They struggle with general or domain-specific reasoning outside of mathematical contexts.
{\bf Language Restriction:} They are confined exclusively to the Python code style. 
Therefore, it is essential to develop a prompting methodology that combines the broad applicability of natural language prompts with the logical precision of code prompts, all while maintaining clarity.


\begin{figure*}[t]
\centering
  \includegraphics[width=0.9\textwidth]{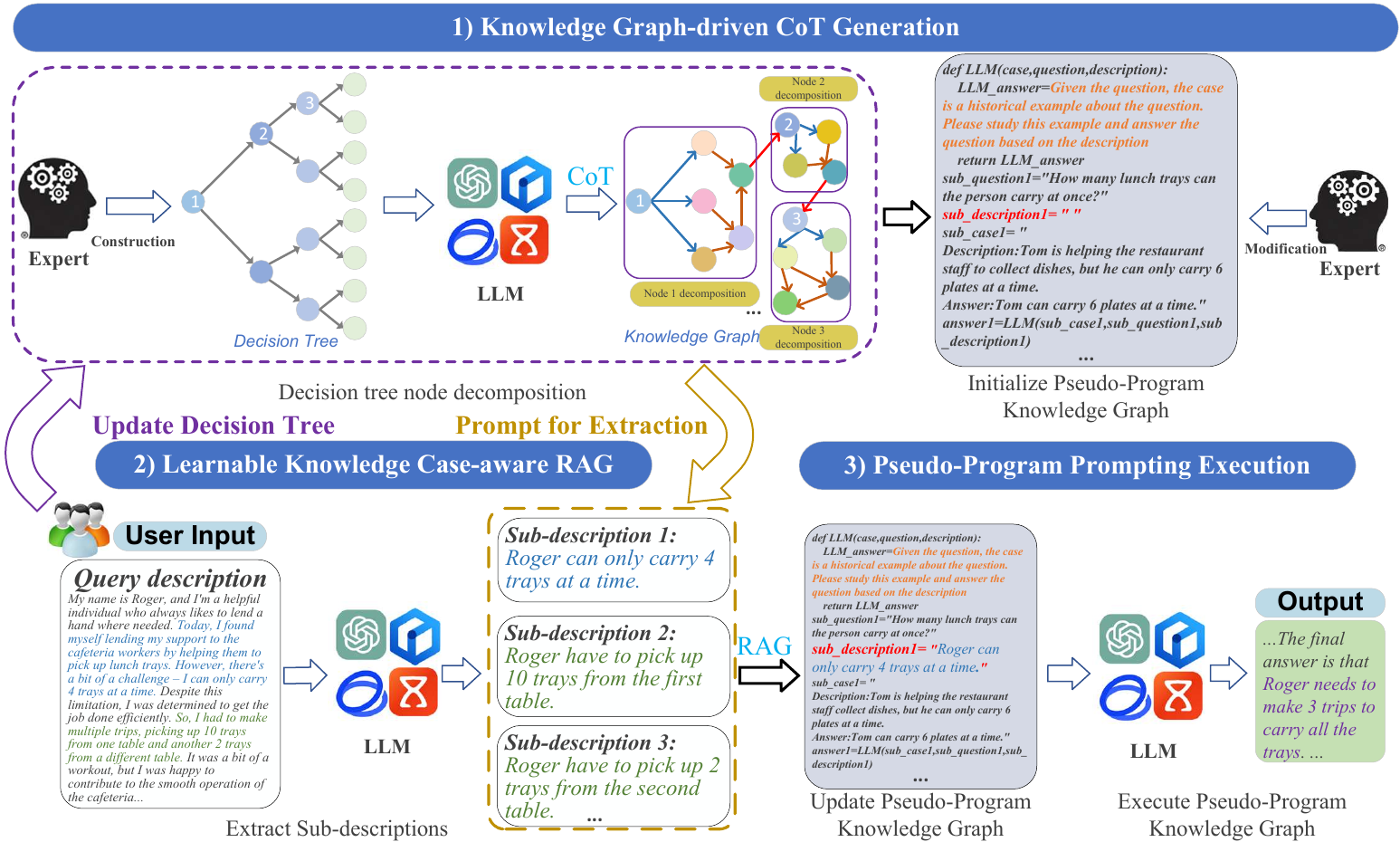}
  \caption{An overview of our CoT-RAG framework.}
  \label{CoT-RAG}
  \vspace{-3mm}
\end{figure*}
\section{Methodology}


\textbf{Overview.} To address the aforementioned concerns, we design CoT-RAG, a novel reasoning framework with three stages (Figure \ref{CoT-RAG}). First, experts construct and input a coarse-grained decision tree that represents the reasoning logic of problems in a specific domain, where experts only need to build it once, then the LLM transforms it into a knowledge graph for a deeper understanding of its internal logic in the \textit{Knowledge Graph-driven CoT Generation} phase (§\ref{stage1}). Next, during the \textit{Learnable Knowledge Case-aware RAG} phase (§\ref{stage2}), users input query descriptions related to this domain and the LLM extracts sub-descriptions to update the knowledge graph for accurately generating results in the next stage. Third, the LLM uses  \textit{Pseudo-Program Prompting Execution} (§\ref{stage3}) to process the updated knowledge graph and produces the ultimate result. Specific algorithm demonstration and time complexity analysis are given in Appendix \ref{sec:appendix—5}, and related notations are shown in Appendix \ref{sec:appendix—6}, which displays significant differences between our decision trees and knowledge graphs w.r.t. their conventional notions.


\begin{figure*}[t]
\centering
  \includegraphics[width=0.9\textwidth]{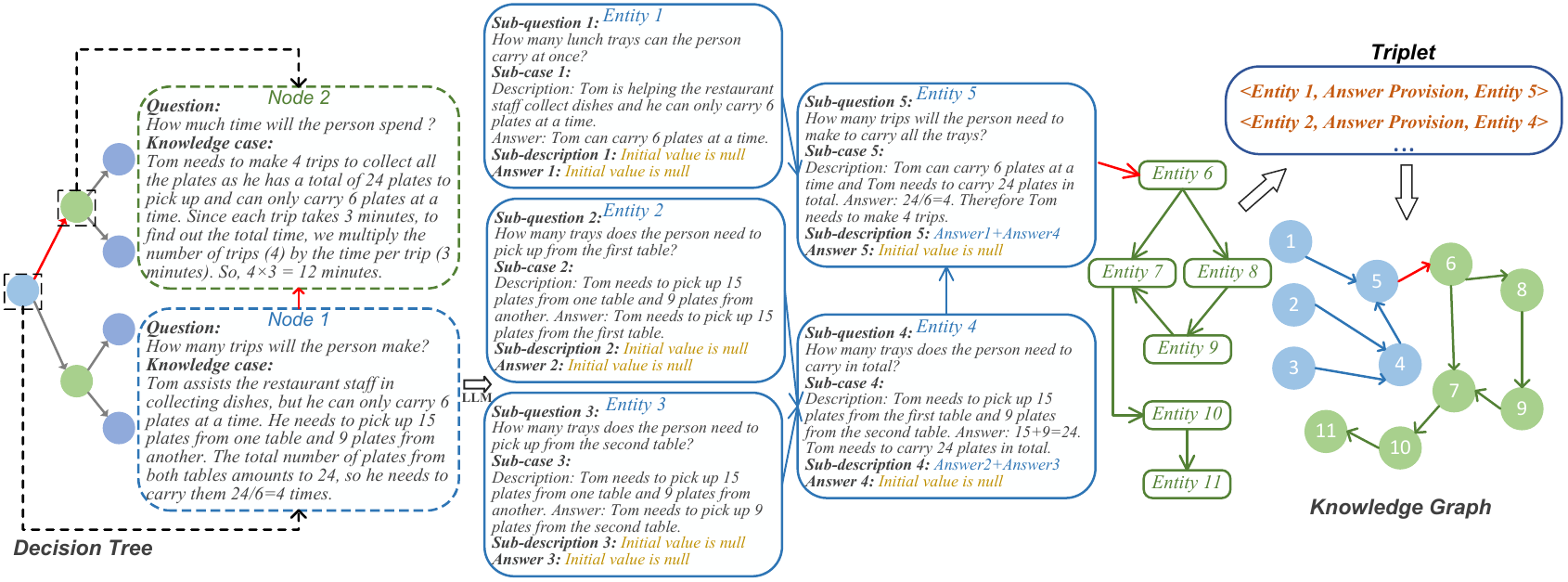}
  \caption{An example of decomposing a decision tree into a knowledge graph.}
  \label{KG}
  \vspace{-3mm}
\end{figure*}

\subsection{Knowledge Graph-driven CoT Generation}
\label{stage1}
A decision tree (DT) is a tree-structured algorithm for classification and regression, where the internal nodes ``test'' a condition, branches represent the test outcome, and the leaf nodes denote the final results \citep{magerman-1995-statistical}. 
DTs offer robust logical coherence and interpretability, enabling human intervention \citep{10562290,kalra2024can}. They are widely used in critical areas such as education \citep{10.1145/3447490.3447504}, finance \citep{10.1145/3482632.3482733,CHOU2024111550}, and medicine \citep{he-etal-2024-generative,10.1145/3675094.3680116}, which motivates us to utilize a manually crafted DT that represents the reasoning logic of problems in a specific domain, to modulate the inference processes of LLMs. Distinct from traditional DTs, each node in our DT contains a \textit{Question} and a \textit{Knowledge case}, which represents some descriptions from user input processed by the node (e.g., Figure \ref{KG}), information supplemented artificially (e.g., Appendix \ref{sec:appendix-2} Table \ref{CSQA}), or relevant considerations (e.g., Appendix \ref{sec:appendix-2} Table \ref{CoinFlip}) to assist LLMs in responding the \textit{Question} of the node. Each branch implies that the output of the parent node is fed to the child node.

However, designing a ``fine-grained'' DT manually consumes a lot of financial and human resources. To alleviate this situation, LLMs can efficiently decompose the ``coarse-grained'' DT nodes provided by experts, leveraging their proficiency in decomposition \citep{huang-etal-2024-qdmr}. Furthermore, \textit{Knowledge case} can assist LLMs in learning how to decompose each node of the DT,  mitigating incorrect results caused by the tendency of LLMs to make errors in vertical fields \citep{10.1109/MNET.2024.3435752,article}. Nevertheless, the complex relationships among the new nodes after the decomposition of the DT need to be captured. Considering that knowledge graphs (KGs), featuring clear delineation and interpretability, precisely represent intricate inter-individual relationships and facilitate inference \citep{sun2024thinkongraph}, we introduce them to represent decomposed DTs. As depicted in \textit{Stage 1} of Figure \ref{CoT-RAG}, LLMs transform each node in the coarse-grained DT built by experts into several entities and generate a highly transparent KG, optimizing the decision-making process.

We provide an example in Figure \ref{KG} to show the DT node decomposition process. 
The \textit{node 1}: "\textit{How many trips will the person make?}" is decomposed by the LLM into five entities, each corresponding to four attributes: (i) \textit{Sub-question}, a simplified component of the original complex \textit{Question}; (ii) \textit{Sub-case}, which is a concise case corresponding to \textit{Sub-question}, derived from the LLM-decomposed  \textit{Knowledge case}; (iii) \textit{Sub-description}, a text description corresponding to \textit{Sub-question} and \textit{Sub-case}, its initial value is null and will be assigned by extracting from users' input query descriptions in § \ref{stage2} or obtained from \textit{Answer} of other entities; and (iv) \textit{Answer}, which refers to the LLM's output for matching \textit{Sub-question}, its initial value is null and will be assigned in § \ref{stage3}.
Edges between entities represent \textit{Answer Provision} relationships, for example, the triplet <\textit{Entity 1, Answer Provision, Entity 5}> indicates that the \textit{Answer} of \textit{Entity 1} is provided to the \textit{Sub-description} of \textit{Entity 5}. The \textit{Answer} of \textit{Entity 5} represents the \textit{node 1}’s inference result. Moreover, as \textit{node 2} in the DT is the child of  \textit{node 1}, \textit{Entity 5} points to \textit{Entity 6}, which is the entity  after \textit{node 2} is decomposed by the LLM.

\subsection{Learnable Knowledge Case-aware RAG}
\label{stage2}
§ \ref{stage1} pertains to the introductory phase of the entire CoT-RAG framework. Subsequently, it comes to the application phase targeting users, that is,\textbf{ how to extract brief yet significant information from relatively long query descriptions input by users}. Unlike traditional vector-based retrieval in RAG, we utilize LLM-based retrieval \citep{shen-etal-2024-retrieval} for long query descriptions, combining \textit{Sub-question} and \textit{Sub-case} of each entity as a prompt generated in the previous stage, to extract relevant descriptions that are assigned to the corresponding \textit{Sub-description}, which has higher accuracy and shorter runtime (refer to § \ref{results} and Appendix \ref{sec:appendix—8}).  

We present an example to illustrate specific \textit{Sub-description} extraction process. On the premise that the KG generated from Figure \ref{KG} is taken as the output of \textit{Stage 1} of CoT-RAG, the user inputs a query description about \textit{Roger}, as depicted in \textit{Stage 2} of Figure \ref{CoT-RAG}, and the LLM uses this KG (including \textit{Sub-question} and \textit{Sub-case} of \textit{Entities 1, 2, and 3}) as a prompt to extract the corresponding descriptions as \textit{Sub-description}. Specifically, \textit{Sub-description 1} is "\textit{Roger can only carry 4 trays at a time}", \textit{Sub-description 2} is "\textit{Roger has to pick up 10 trays from the first table}", and \textit{Sub-description 3} is "\textit{Roger has to pick up 2 trays from the second table}". These \textit{Sub-description} and \textit{Sub-case} will be employed by the LLM to precisely answer the corresponding \textit{Sub-question} of each entity in § \ref{stage3}.

Furthermore, if the query descriptions entered by users are new to the DT, they have the capacity to dynamically update the \textit{Knowledge case} within the DT, which facilitates the LLM in generating a more comprehensive and application-oriented knowledge graph, thereby augmenting the flexibility of the knowledge graph.

\subsection{Pseudo-Program Prompting Execution}
\label{stage3}
The first two designs enhance the credibility and interpretability of the reasoning chains generated by LLMs. However, there remains a critical issue: \textbf{how can reasoning chains be represented to ensure the rigorous logical execution of LLMs?} Inspired by QDMRPS \citep{huang-etal-2024-qdmr} and AdaPlanner \citep{NEURIPS2023_b5c8c1c1}, we propose Pseudo-program Prompting (PsePrompting), which allows the LLM to represent the knowledge graph reasoning chain as pseudo-programs, referred to as the pseudo-program knowledge graph (PKG). Its initial version created by the LLM of \textit{Stage 1} is shown in Table \ref{initial} of Appendix \ref{sec:appendix—1}, where \textit{Entities 1, 2, and 3’s} \textit{Sub-description} are filled based on user input in \textit{Stage 2} ( Table \ref{complete} in Appendix \ref{sec:appendix—1}). The final result of the PKG executed by the LLM in \textit{Stage 3} is depicted in Table \ref{result} of Appendix \ref{sec:appendix—1}. The LLM processes each entity of the PKG sequentially, it first retrieves the \textit{Sub-case} and \textit{Sub-description} for learning, then generates the \textit{Answer}. We can observe that PsePrompting features a simple logical structure and is easily understandable. Additionally, it demonstrates broad applicability for addressing queries that demand domain-specific syntax or reasoning paradigms (Appendix \ref{sec:appendix-2}), eliminates reliance on external interpreters, and supports extension to programming languages such as C++ and Java (Appendix \ref{sec:appendix-3}).
\label{sec:Methodology}
\section{Experimental setup}
\subsection{Datasets}
\textbf{General domains: }Our CoT-RAG is evaluated on nine benchmark datasets from three categories of reasoning problems: \textbf{Arithmetic Reasoning:} AQUA \citep{ling-etal-2017-program}  , GSM8K \citep{2021arXiv211014168C}, MultiArith \citep{roy-roth-2015-solving}, and SingleEq \citep{koncel-kedziorski-etal-2015-parsing}; \textbf{Commonsense Reasoning:} HotpotQA \citep{wolfson-etal-2020-break}, CSQA \citep{talmor-etal-2019-commonsenseqa}, and SIQA \citep{sap-etal-2019-social}; 
\textbf{Symbolic Reasoning:} Last Letter Concatenation \citep{wei2022chain}, and Coin Flip \citep{wei2022chain}. 

\noindent\textbf{Vertical domains: } To further demonstrate the scalability of our method across different vertical domains, we evaluate it on four open-source datasets from the legal, financial, and logic fields: LawBench (LaB) \citep{fei-etal-2024-lawbench}, LegalBench (LeB), CFBenchmark (CFB) \citep{2023arXiv231105812L}, and AGIEval (AGI) \citep{zhong-etal-2024-agieval}.

Additionally, following GraphRAG \citep{2024arXiv240416130E} and Graph-CoT \citep{jin-etal-2024-graph}, we employ an LLM to adapt the datasets to satisfy our testing needs, where the specific details and descriptions of datasets are shown in Appendix \ref{sec:appendix-9}. 
\begin{table*}[ht]
  \centering
  \scriptsize
  \begin{tabular}{ccccccccccc}
    \hline
    Method & AQuA & GSM8K & MultiArith & SingEq& HotpotQA & CSQA & SIQA& Letter& Coin& Average\\
    \hline
    \multicolumn{11}{c}{\textit{Without external knowledge and exemplars}} \\  
    \hline
Zero-shot&	42.6&	77.8&	95.9&	86.8&	80.1&	74.5&	77.3&	35.8&	76.7&	71.9\\
Zero-shot-CoT&	43.4&	78.3&	96.7&	88.5&	81.4&	75.6&	78.0&	34.5&	75.6&	72.4\\
PS&	50.1&	82.8&	96.9&	89.4&	83.0&	\underline{74.2}&	\underline{76.3}&	44.7&	79.5&	75.2\\
QDMRPS&	47.3&	83.8&	95.2&	90.7&	86.7&	76.6&	77.8&	36.5&	76.3&	74.5\\
\hline
    \multicolumn{11}{c}{\textit{With exemplars}} \\  
    \hline
Manual-CoT&	54.3&	85.8&	97.2&	92.3&	85.7&	79.6&	82.4&	39.6&	79.2&	77.3\\
Auto-CoT&	47.8&	82.4&	97.5&	91.6&	86.1&	76.4&	80.6&	41.0&	81.2&	76.1\\
Complex-CoT& 	51.7&	83.5&	96.6&	92.8&	82.8&	76.9&	78.5&	37.7&	81.9&   75.8\\ 
Iter-CoT&	51.6&	80.0&	97.8&	93.4&	64.8&	76.9&	77.3&	41.8&	77.5&   73.5\\
ZEUS&	51.9&	88.4&	97.3&	92.8&	84.9&	77.4&	81.7&	42.8&	82.5&   77.7\\
Pattern-CoT&	52.8&	85.3&	97.7&	91.3&	82.5&	76.3&	78.9&	41.4&	83.7&   76.7\\
    \hline
    \multicolumn{11}{c}{\textit{With external knowledge}} \\  
    \hline
KD-CoT&	22.3&	68.4&	\underline{76.0}&	62.3&	79.9&	85.6&	90.8&	\underline{24.6}&	58.3&   63.1\\
IRCoT&	20.7&	\underline{65.6}&	78.3&	65.1&	87.5&	82.8&	87.9&	28.2&	54.5&   63.4\\
KG-CoT&	\underline{12.3}&	66.8&	78.6&	\underline{61.5}&	\underline{73.5}&	88.9&	92.1&	26.7&	\underline{53.2}&   \underline{61.5}\\
    \hline
    \multicolumn{11}{c}{\textit{With both external knowledge and exemplars}} \\  
    \hline

\textbf{CoT-RAG (ours)}&	\textbf{65.7}&	\textbf{94.7}&	\textbf{98.5}&	\textbf{98.7}&	\textbf{98.4}&	\textbf{97.9}&	\textbf{98.7}&	\textbf{54.6}&	\textbf{94.7}&	\textbf{89.1}\\

    \hline
  \end{tabular}
  \caption{\label{ERNIE-Speed-128K}
    Accuracy on nine datasets from three categories of reasoning tasks using ERNIE-Speed-128K. Throughout the tables in this paper, the best results are highlighted in bold, and the poorest results are underlined.}
  
\end{table*}
\begin{table}[ht]
  \centering
  \scriptsize
  \begin{tabular}{cccccc}
    \hline
    Method & LaB & LeB &  CFB &  AGI & Average\\ 
    \hline
\multicolumn{6}{c}{\textit{Graph-form LLM-based RAG methods}} \\  
    \hline
GraphRAG&	94.8&	97.5& 73.1& 54.6&80.0\\
KG-CoT&	89.3& 93.6&	72.8& 32.6&72.1\\
ToG&	86.7&	90.2&	68.3& 64.2&77.4\\
PoG&	93.8&	91.7& 89.5& 45.3&80.1\\
RRKG&	91.3&	92.4& 74.7& 27.4&71.5\\
RoG& 90.4& 88.1&  88.7& 67.5&83.7\\
Graph-CoT& \underline{54.7}&  \underline{68.2}&  \underline{63.1}& \underline{24.7}&\underline{52.7}\\
ToG-2& 92.5&  93.7& 76.6& 70.8&83.4\\
AtomR& 83.6&  82.5& 77.3& 37.5&70.2\\
\hline
\multicolumn{6}{c}{\textit{Variants of CoT-RAG}} \\  
    \hline
    CoT-RAG (IndexFlatL2)&	92.7&	93.7& 85.2& 67.8 &84.9\\
CoT-RAG (IndexFlatIP)&	91.5&	91.9& 87.1& 72.4	&85.7\\
CoT-RAG (IndexIVFFlat)&	93.6&	92.1&	86.8& 75.3&87.0\\
CoT-RAG (IndexLSH)&	90.8&	92.5&  88.1& 74.6&86.5\\
CoT-RAG (IndexPQ)&	93.1&	92.9&	87.4& 73.9&86.8\\
CoT-RAG (IndexIVFPQ)&	93.6&	93.1&	87.8& 76.2&87.7\\
CoT-RAG (Zero-expert)&	93.6&	94.7& 86.3& 74.8&87.4\\
\textbf{CoT-RAG (ours)}&	\textbf{99.3}&		\textbf{98.6}&	\textbf{94.7}& \textbf{88.3}&\textbf{95.2}\\

    \hline
  \end{tabular}
  \caption{\label{accuracy_vertical_domains}
    Accuracy on four datasets from vertical domains using GPT-4o mini}. 
  
\end{table}
\subsection{Baselines}
\label{baselines}
\textbf{General domains: }We compare our CoT-RAG with following methods that focus on CoT: (1) \textbf{Manual-CoT} \citep{wei2022chain} uses a “thought chain” prompt to guide LLMs through a step-by-step solution process, leading to a detailed answer. (2) \textbf{Zero-shot-CoT} \citep{kojima2022large} encourages LLMs to generate reasoning steps automatically by appending “Let’s think step by step” to the question. (3) \textbf{Complex-CoT} \citep{fu2023complexitybased} represents a simple and effective example selection scheme for multi-step
reasoning. (4) \textbf{Auto-CoT} \citep{zhang2023automatic} automates the generation of high-quality prompts, improving efficiency and accuracy in reasoning tasks. (5) \textbf{PS} \citep{wang-etal-2023-plan} first generates a coarse task plan, followed by a fine-grained solution process. (6) \textbf{KD-CoT} \citep{2023arXiv230813259W} modifies reasoning traces in CoT via a retriever to interact with external knowledge stored in an unstructured knowledge base. (7) \textbf{IRCoT} \citep{trivedi-etal-2023-interleaving} interleaves retrieval with steps in a CoT, in turn using retrieved results to improve CoT. Notice that both KD-CoT and IRCoT retrieve external knowledge from an unstructured knowledge base, which is different from our highly-structured and logical DTs and KGs. (8) \textbf{QDMRPS} \citep{huang-etal-2024-qdmr} decomposes problems into QDMR-based directed acyclic graphs and reasons step-by-step based on dependencies. (9) \textbf{Iter-CoT} \citep{sun-etal-2024-enhancing}  prompts LLMs to self-correct their errors in reasoning chains by leveraging iterative bootstrapping. (10) \textbf{KG-CoT} augments LLMs \citep{ijcai2024p734} via a graph reasoning model
that generates explicit reasoning paths over factual KGs, e.g., Freebase, which is different from our specialized DTs and KGs, as well as our KG-focused logical reasoning. (11) \textbf{ZEUS} \citep{kumar-etal-2025-enhancing} improves CoT prompting by utilizing uncertainty estimates to select effective demonstrations without needing access to model parameters. (12) \textbf{Pattern-CoT} \citep{Zhang_Wang_Wu_Wang_2025} employs reasoning patterns to enhance CoT prompting effectiveness. (13) In addition, we evaluate the intrinsic capabilities of the LLM, where we only deal with the original problem input to the LLM without using any additional methods, namely \textbf{Zero-shot}. Methods like PoT \citep{chen2023program} and Faithful CoT \citep{lyu-etal-2023-faithful}, which cannot handle complex non-mathematics reasoning tasks, especially in the field of commonsense reasoning, are excluded from this comparison. 

\noindent \textbf{Vertical domains: } We utilize the Faiss \citep{faiss} vector database to replace the LLM-based retrieval in CoT-RAG with vector-based retrieval, where six variants are set up according to different indexes \citep{faissindex}, namely CoT-RAG (IndexFlatL2), CoT-RAG (IndexFlatIP), CoT-RAG (IndexIVFFlat), CoT-RAG (IndexLSH), CoT-RAG (IndexPQ), and CoT-RAG (IndexIVFPQ). We also compare the scalability of CoT-RAG under the condition of zero-expert, that is, CoT-RAG (Zero-expert), where LLMs replace experts to generate decision trees. 
Additionally, besides KG-CoT (§ \ref{baselines}), we also incorporate eight state-of-the-art graph-form LLM-based RAG methods: \textbf{RoG} \citep{luo2024reasoning} first generates KG-grounded relation paths as plans, then uses them to retrieve reasoning paths from KGs for LLMs to reason reliably; \textbf{Graph-CoT} \citep{jin-etal-2024-graph} enhances LLMs by encouraging them to conduct iterative reasoning on the graph; \textbf{ToG} \citep{sun2024thinkongraph} employs LLMs  for iterative beam search on KGs, finding the best paths and returning top results.; \textbf{ToG-2} \citep{ma2025thinkongraph} utilizes KGs to connect documents via entities, which facilitates deep and knowledge-guided context retrieval, thereby enhancing the reasoning ability of LLMs; \textbf{RRKG} \citep{ji-etal-2024-retrieval} combines explainable knowledge graphs with LLMs to enhance complex reasoning capabilities; \textbf{AtomR} \citep{2024arXiv241116495X} is a framework that enables LLMs to conduct accurate heterogeneous knowledge reasoning at the atomic level; \textbf{GraphRAG} \citep{2024arXiv240416130E} combines the advantages of RAG and query-focused summarization, and it uses knowledge graphs to store source document data for efficient global question answering; \textbf{PoG} \citep{10.1145/3696410.3714892} enhances LLM reasoning by integrating knowledge reasoning paths from KGs.

Moreover, we evaluate these methods across five NL-based LLMs: ERNIE-Speed-128K \citep{BaiduERNIE}, ERNIE-3.5-128K \citep{BaiduERNIE}, GLM-4-flash \citep{Zhipuaiglm-4-flash}, GPT-4o mini \citep{openai2024gpt4omini}, and GPT-4o \citep{openai2024gpt4o}. Following \citet{wei2022chain,kojima2022large}, we invoke LLMs that are not fine-tuned via API and  set the temperature to 0 to ensure deterministic outputs. However, there is one difference, that is, we set the max tokens to 1000 to accommodate the need for a longer context. Moreover, in line with \citet{wang-etal-2023-plan}, for Manual-CoT and Auto-CoT, we typically use 8 demonstration examples across most tasks, 4 examples for AQuA and Last Letter tasks, and 7 examples for CSQA.  Complex-CoT selects exemplars with most complex rationales as demonstrations. The remaining baselines use default settings. In our proposed CoT-RAG, each DT node’s \textit{Knowledge case} contains only one demonstration example (e.g., \textit{Node 1} of Figure \ref{KG}).

\section{Experimental Results}
\label{results}
Table \ref{ERNIE-Speed-128K} presents our results on ERNIE-Speed-128K, with additional results from other LLMs available in Appendix \ref{sec:appendix—4}. The experimental data demonstrate that CoT-RAG significantly improves reasoning accuracy across all datasets compared to existing CoT techniques, ranging from 4.0\% to 44.3\%. Specifically illustrated by representative results, compared to Manual-CoT, Zero-shot-CoT, Auto-CoT, PS, QDMRPS, KD-CoT, Iter-CoT and KG-CoT, the average accuracy across datasets increases by 4.0\%-15.2\%, 7.3\%-23.0\%, 5.0\%-17.1\%, 5.7\%-18.5\%, 4.8\%-19.5\%, 13.7\%-41.1\%, 4.2\%-21.3\%, and 13.8\%-44.3\%, respectively. The overall improvement in accuracy can be attributed to the comprehensive optimizations of our method in terms of credible CoT generation and rigorous instruction execution. 

Table \ref{accuracy_vertical_domains} indicates that, compared with the baselines, our CoT-RAG has increased accuracy by 8.9\% to 80.6\%. Specifically, compared to CoT-RAG methods based on different vector indexes, the average accuracy of our method has increased by 8.6\%-12.1\%. In particular, we observe that information loss and high running time may occur when using vectors to store and retrieve knowledge. Furthermore, the average accuracy of CoT-RAG (Zero-expert) is 7.8\% lower than our CoT-RAG with expert involvement, which further highlights the indispensable role of experts in vertical domain applications. Other 
graph-form LLM-based RAG methods, including KG-CoT, ToG, PoG, RRKG, etc., target factual knowledge graph question answering (KGQA) and do not cover logical reasoning in vertical domains, showing a relatively low accuracy. In particular, these methods have a very poor accuracy on the AGIEval dataset with high logical requirements. Additionally, it is clearly observable that the average accuracy of CoT-RAG variants significantly outperforms graph-form LLM-based RAG methods. The above experimental analysis further justifies the strong scalability and sufficient novelty of our CoT-RAG. 
The relevant analysis regarding ablation results, robustness study, the relationship between accuracy and reasoning complexity, the runtime and resource usage can be found in Appendix \ref{sec:appendix—7}. 

\section{Related Work}
\subsection{Chain of Thought Reasoning}
To utilize LLMs' reasoning capabilities, \citet{wei2022chain} propose Chain-of-Thought prompting, adding reasoning steps before the answer to improve performance. Subsequent works have enhanced CoT reasoning in areas including prompt format \citep{gao2023pal,chen2023program,lyu-etal-2023-faithful}, selection \citep{yao2023tree,yu2024thought,kumar-etal-2025-enhancing,Zhang_Wang_Wu_Wang_2025}, ensemble \citep{weng-etal-2023-large,li-etal-2023-making,hu-etal-2024-rankprompt,fu2023complexitybased}, decomposition \citep{zhou2023leasttomost,khot2023decomposed,press-etal-2023-measuring,huang-etal-2024-qdmr}, model fine-tuning \citep{liao-etal-2025-skintern,2025arXiv250203373Y}, and planning \citep{wang-etal-2023-plan,sun-etal-2024-pearl,wang2023describe}. \citet{chen2023program} design PoT, separating computation and reasoning with code-trained LLMs. \citet{lyu-etal-2023-faithful} introduce Faithful CoT, combining natural and symbolic languages to enhance interpretability. \citet{kojima2022large} propose Zero-shot-CoT, automating reasoning steps with minimal human input. \citet{wang-etal-2023-plan} address errors in Zero-shot-CoT by introducing PS prompting, which refines coarse task plans. \citet{huang-etal-2024-qdmr} use QDMR-based graphs for structured problem decomposition. \citet{kumar-etal-2025-enhancing} design ZEUS to improve CoT prompting by utilizing uncertainty estimates to select effective demonstrations. \citet{Zhang_Wang_Wu_Wang_2025} introduce Pattern-CoT, which employs reasoning patterns to enhance CoT prompting effectiveness. However, these methods solely rely on the reasoning and evaluation of the LLM itself, leading to low reliability. We refer readers to the survey \citep{chu-etal-2024-navigate} for more related works.
\subsection{RAG Reasoning with Knowledge Graph}
Retrieval-Augmented Generation (RAG) enhances LLMs by using relevant content retrieved from knowledge sources, aiming to mitigate the black-box nature and hallucination issues and improve text generation quality \citep{10.1145/3676957,10.1145/3626772.3661377}. Previous works have explored different ways in which LLMs leverage retrieved or generated text as external knowledge to boost reasoning \citep{NEURIPS2020_6b493230,JMLR:v24:23-0037,sun2023recitationaugmented,hagstrom-etal-2023-effect}. Recent research has further enhanced RAG reasoning by incorporating knowledge graphs \citep{luo2024reasoning, ji-etal-2024-retrieval,ma2025thinkongraph}. \citet{sun2024thinkongraph} introduce ToG, an innovative framework that enhances LLMs through interaction with knowledge graphs and expanding reasoning paths via beam search. \citet{2024arXiv240416130E} propose GraphRAG, which combines the strengths of RAG and query-focused summarization, using knowledge graphs to store source document data for efficient global question answering. \citet{saleh-etal-2024-sg} design SG-RAG, a zero-shot method, utilizing structured information from knowledge graphs to enable LLMs to accurately answer multi-hop questions. \citet{ma2025thinkongraph} propose ToG-2, which utilizes KGs to connect documents via entities, facilitating deep and knowledge-guided context retrieval and enhancing the reasoning ability of LLMs. Our work incorporates RAG to enhance the reasoning of LLMs based on the generated knowledge graphs.
\section{Conclusion}
To address the issues of low reliability in reasoning chains generated solely by LLMs and poorer reasoning performance from natural language prompts compared with code prompts, we propose CoT-RAG, a novel reasoning framework with three key designs: (i) \textit{Knowledge Graph-driven CoT Generation}, 
which introduces knowledge graphs to regulate LLMs' reasoning chain generation, thereby enhancing reasoning reliability; (ii) \textit{Learnable Knowledge Case-aware RAG}, intergrating RAG into knowledge graphs to retrieve relevant sub-cases and sub-descriptions for learnable information; (iii) \textit{Pseudo-Program Prompting Execution}, which inspires LLMs to execute logically reasoning tasks in pseudo-programs.
Evaluations across nine public datasets demonstrate that CoT-RAG outperforms existing methods, particularly in four domain-specific datasets, validating its powerful cross-domain scalability.

\section{Limitations}
Although the CoT-RAG framework demonstrates effectiveness, it faces two key limitations. First, its implementation relies on LLMs with advanced program understanding and execution capabilities, thus publicly available LLMs, especially those with a smaller scale (e.g., 7B or 13B parameters), fall short on these requirements, necessitating our use of proprietary LLMs. This constraint limits the framework's generalizability and precludes the evaluation of how LLMs' parameter sizes impact CoT-RAG performance. Second, the construction of decision trees is influenced by the expert's domain-specific knowledge and background. In future, we plan to explore more automated methods in decision tree design. For example, in the literature of vertical domains (e.g., court proceeding records, medical case history, etc.), there are already descriptions of knowledge cases. We shall explore automated coarse-grained DT construction from such domain knowledge available in the form of text. Moreover, we will explore more widely applicable prompting techniques to adapt to small language models.

\section{Acknowledgements}
This work was supported by the National Key Research and Development Program of China (No. 2023YFB4502701), the National Natural Science Foundation of China (Grant Nos.U22A2027, 82090044 and 62402187), the China Postdoctoral
Science Foundation (Nos. GZB20240243 and 2024M751009), and the Postdoctoral
Project of Hubei Province (No.2024HBBHCXA024). Arijit Khan acknowledges support from the Novo Nordisk Foundation Grant NNF22OC0072415.

\bibliography{custom}

\begin{algorithm*}[t]
\caption{\large CoT-RAG: Enhanced Reasoning Framework Based on CoT and RAG}
\label{alg:cot-rag}

\begin{algorithmic}[1]
\STATE \textbf{Function} KnowledgeGraphDrivenCoTGeneration(\textit{field}, \textit{ExpertModify}):
\STATE \quad\quad// \textit{field} depends on the actual scenario of CoT-RAG application
\STATE \quad\quad $\text{decision\_tree} \leftarrow \text{ExpertBuildDecisionTree}(\textit{field})$
\STATE \quad\quad$\text{knowledge\_graph} \leftarrow \text{LLMDecomposeDecisionTree}(\textit{decision\_tree})$
\STATE \quad\quad$\text{initial\_PKG} \leftarrow \text{LLMGeneratePKG}(\textit{knowledge\_graph})$ 
\STATE \quad\quad\textbf{if} \textit{ExpertModify} is \text{true} \textbf{then}
    \STATE \quad\quad\quad\quad$\text{initial\_PKG} \leftarrow \text{ExpertModifyKnowledgeGraph}(\textit{initial\_PKG})$
\STATE \quad\quad\textbf{end}
\STATE \quad\quad\textbf{return} $\text{initial\_PKG}$

\STATE
\STATE \textbf{Function} LearnableKnowledgeCaseAwareRAG(\textit{initial\_PKG}, \textit{decision\_tree}, \textit{is\_useful}):
\STATE \quad\quad// Get the query description entered by the user
\STATE \quad\quad$\text{query\_description} \leftarrow \text{GetUserQueryDescription}()$
\STATE \quad\quad\textbf{if} \textit{is\_useful} is \text{true} \textbf{then}
    \STATE \quad\quad\quad\quad$\text{updated\_decision\_tree} \leftarrow \text{Update}(\textit{query\_description}, \textit{decision\_tree})$
\STATE \quad\quad\textbf{end}
    \STATE \quad\quad$\text{updated\_PKG} \leftarrow \text{LLMExtractSubDescriptions}(\textit{initial\_PKG}, \textit{query\_description})$
\STATE \quad\quad\textbf{return} $\text{updated\_PKG}$

\STATE
\STATE \textbf{Function} PseudoProgramPromptingExecution(\textit{updated\_PKG}):
\STATE \quad\quad// Generate final results
\STATE \quad\quad$\text{final\_result} \leftarrow \text{LLMGenerateFinalResult}(\textit{updated\_PKG})$
\STATE \quad\quad\textbf{return} $\text{final\_result}$
\end{algorithmic}
\end{algorithm*}  

\begin{table*}[ht]
  \centering
  
  \begin{tabular}{m{1.5cm}<{\centering}|m{14cm}}
    \hline
    
     Notation &  Description \\
    \hline
    KG(s) & A \small \textbf{K}nowledge \textbf{G}raph(s) is a semantically structured network, which uses triplets to describe cross-domain entity relationships and attribute values \citep{mendes-etal-2024-application,elhammadi-etal-2020-high}. It integrates general facts, domain-specific knowledge (such as in mathematics, law, finance, and biomedical), and common-sense knowledge, providing a knowledge base and reasoning foundation for complex tasks like cross-domain intelligent decision-making and comprehensive knowledge-based question-answering \citep{schneider-etal-2022-decade,arsenyan-etal-2024-large}. \textbf{Distinct from traditional KGs, the nodes in our KGs include attributes such as Sub-question/Sub-case. The edges represent reasoning dependencies (i.e., “Answer Provision” relationships) rather than factual relationships as in the classic KGs and place a pronounced emphasis on the "structuring of reasoning logic"  As depicted in Figure \ref{KG}, the triple <Entity 1, Answer Provision, Entity 5> indicates that the answer of Entity 1 is provided to the sub-description of Entity 5, which aids in the reasoning and answer-generation processes of Entity 5. This relationship clarifies the information flow and dependency among different entities within the KG. It enables LLMs to leverage relevant information during the reasoning process, thereby enhancing the controllability and reliability of reasoning.}\\
    \hline
    DT(s) &A \small \textbf{D}ecision \textbf{T}ree(s) is a tree-structured machine-learning algorithm that finds extensive application in classification and regression tasks \citep{wu-etal-2020-dtca,10.1145/3447490.3447504}. Its fundamental principle entails recursively partitioning a dataset into progressively smaller subsets, thereby giving rise to a dendritic structure. In this construct, each internal node signifies a test of a characteristic or attribute, each branch represents a particular value of that characteristic, and each leaf node represents a classification result or a regression value  \citep{he-etal-2024-generative,magerman-1995-statistical}. \textbf{Different from the concept described above, the decision tree in this paper refers to a tree-shaped structure constructed by experts, where there are well-defined logical relationships between parent and child nodes. The parent nodes represent more general or antecedent questions. The information and reasoning results from parent nodes are passed on to child nodes, facilitating the resolution of problems at the child node. For example, in the decision tree in Figure \ref{KG} about transporting dishes in a restaurant, the parent node could be “Calculate the number of times restaurant staff transport dishes,” while a child node might be “Calculate the total time for transporting dishes.” Branching conditions are determined by domain-specific logic defined by experts. Parent nodes point to child nodes to construct a complete reasoning path, enabling LLMs to reason step-by-step following this structured logic.}\\
    \hline
    RAG & \small \textbf{R}etrieval-\textbf{A}ugmented \textbf{G}eneration incorporates a retrieval mechanism within the text-generation process \citep{2024arXiv241012837G}. It empowers LLMs to retrieve pertinent information from external knowledge repositories, and integrate it into the generated outputs, endowing the generated content with enhanced accuracy, pertinence, and factual fidelity. RAG can be generally categorized into three types: Documen-based Retrieval RAG, Graph-based Retrieval RAG, and Multi-modal Retrieval RAG \citep{2024arXiv240219473Z,2024arXiv240808921P}. \textbf{Distinct from the above-mentioned vector-based retrieval RAG, the RAG in this paper harnesses the formidable retrieval capabilities inherent in LLMs. It retrieves relevant content from the knowledge cases of the decision tree and the query descriptions input by users, reducing runtime and enhancing accuracy (Appendix \ref{sec:appendix—7}).}
    \\ \hline
  \end{tabular}
  \caption{\label{Notation}
     Frequently-Used Notations.
  }
\end{table*}

\begin{table*}[ht]
  \centering
  \scriptsize
  \renewcommand{\arraystretch}{1.2} 
  \begin{tabular}{ccccccccccc}
    \hline
    Method & AQuA & GSM8K & MultiArith & SingEq& HotpotQA & CSQA & SIQA& Letter& Coin& Average\\
    \hline
        \multicolumn{11}{c}{\textit{Without external knowledge and exemplars}} \\  
    \hline
Zero-shot& 82.4& 92.6& 96.8& 97.7& 87.2& 83.3& 87.1& 92.2& 98.7& 90.9\\
Zero-shot-CoT&	81.8&	93.3&	96.3&	97.3&	87.7&	\underline{82.7}&	87.8&	92.7&	98.0&	90.8\\
PS&	82.1&	92.8&	97.7&	98.1&	89.9&	84.9&	86.5&	97.3&	99.9&	92.1\\
QDMRPS&	83.5&	94.3&	98.5&	98.8&	90.8&	83.9&	\underline{85.9}&	96.7&	98.3&	92.3\\
\hline
    \multicolumn{11}{c}{\textit{With exemplars}} \\  
    \hline
Manual-CoT&	90.2&	95.3&	98.1&	98.0&	92.3&	89.0&	93.9&	97.7&	\textbf{100}&	94.9\\
Auto-CoT&	84.3&	94.6&	97.3&	98.6&	90.0&	85.3&	90.1&	97.0&	99.8&	93.0\\
Complex-CoT& 	90.8&	93.9&	97.2&	98.0&	91.3&	86.8&	89.3&	96.8&	98.8&   93.7\\
Iter-CoT&	92.6&	96.3&	97.6&	98.4&	84.8&	89.8&	92.3&	94.5&	99.5&   94.0\\
ZEUS&	91.8&	95.0&	97.9&	98.3&	91.8&	88.6&	89.3&	98.3&	98.7&   94.4\\
Pattern-CoT&	86.7&	94.6&	97.9&	98.3&	88.3&	85.8&	88.6&	94.3&	99.5&   92.7\\
    \hline
    \multicolumn{11}{c}{\textit{With external knowledge}} \\  
    \hline
KD-CoT&	72.4&	\underline{82.8}&	85.3&	86.6&	\underline{84.2}&	94.7&	96.2&	87.3&	\underline{90.2}&   \underline{86.6}\\
IRCoT&	76.6&	83.5&	\underline{82.9}&	88.2&	86.5&	92.8&	95.9&	\underline{85.8}&	93.7&   87.3\\

KG-CoT&	\underline{67.2}&	84.6&	88.0&	\underline{84.9}&	84.7&	93.6&	96.8&	89.5&	92.3&   86.8\\
    \hline
    \multicolumn{11}{c}{\textit{With both external knowledge and exemplars}} \\  
     
    \hline

\textbf{CoT-RAG (ours)}&	\textbf{99.0}&	\textbf{99.5}&	\textbf{99.9}&	\textbf{99.6}&	\textbf{99.4}&	\textbf{99.6}&	\textbf{98.3}&	\textbf{99.9}&	\textbf{100}&	\textbf{99.5}\\

    \hline
  \end{tabular}
  \caption{\label{GPT-4o-mini}
    Accuracy on nine datasets from three categories of reasoning tasks using GPT-4o mini.}
  
\vspace{-3mm}  
\end{table*}
\begin{table*}[t]
  \centering
  \scriptsize
  \renewcommand{\arraystretch}{1.2} 
  \begin{tabular}{ccccccccccc}
    \hline
    Method & AQuA & GSM8K & MultiArith & SingEq& HotpotQA & CSQA & SIQA& Letter& Coin& Average\\
    \hline
     \multicolumn{11}{c}{\textit{Without external knowledge and exemplars}} \\  
    \hline
Zero-shot& 78.8& 91.3& 96.5& 93.3& 84.8& \underline{80.3}& 87.1& 91.8& \underline{81.4}& 87.3\\
Zero-shot-CoT&	80.3&	92.0&	96.7&	92.9&	84.3&	80.8&	\underline{86.6}&	92.2&	81.9&	87.5\\
PS&	83.4&	93.6&	97.3&	95.6&	86.6&	84.9&	87.0&	96.8&	94.6&	91.1\\
QDMRPS&	85.7&	94.9&	98.2&	94.3&	89.3&	86.3&	89.5&	97.3&	92.7&	92.0\\
\hline
    \multicolumn{11}{c}{\textit{With exemplars}} \\  
    \hline
Manual-CoT&	87.9&	94.5&	98.6&	96.4&	90.5&	89.3&	92.1&	98.3&	95.7&	93.7\\
Auto-CoT&	82.6&	95.3&	98.2&	95.5&	91.4&	87.3&	91.5&	96.4&	96.8&	92.8\\
Complex-CoT& 	84.3&	94.2&	97.7&	96.1&	87.8&	88.2&	90.6&	96.6&	95.2&   92.3\\
Iter-CoT&	85.8&	95.4&	98.5&	97.2&	87.8&	86.7&	92.6&	97.8&	96.9&   93.2\\
ZEUS&	85.8&	94.9&	98.8&	96.1&	88.8&	85.2&	90.4&	96.6&	93.7&   92.3\\
Pattern-CoT&	83.2&	94.7&	98.2&	95.8&	87.6&	85.4&	89.3&	95.8&	96.2&   91.8\\
    \hline
    \multicolumn{11}{c}{\textit{With external knowledge}} \\  
    \hline

KD-CoT&	64.3&	84.3&	\underline{82.9}&	82.2&	\underline{81.8}&	93.5&	92.5&	79.8&	88.6&   83.3\\
IRCoT&	\underline{57.0}&	\underline{81.5}&	84.5&	84.7&	82.6&	93.7&	94.2&	\underline{75.4}&	91.0& \underline{82.7}\\
KG-CoT&	61.7&	85.7&	84.3&	\underline{82.0}&	83.1&	93.8&	92.4&	76.7&	90.5&   83.4\\
    \hline
    \multicolumn{11}{c}{\textit{With both external knowledge and exemplars}} \\ 
    \hline

\textbf{CoT-RAG (ours)}&	\textbf{96.3}&	\textbf{99.8}&	\textbf{99.4}&	\textbf{99.1}&	\textbf{98.3}&	\textbf{99.8}&	\textbf{99.7}&	\textbf{99.9}&	\textbf{99.7}&	\textbf{99.1}\\

    \hline
  \end{tabular}
  \caption{\label{ERNIE-3.5-128K}
   Accuracy on nine datasets from three categories of reasoning tasks using ERNIE-3.5-128K.}

\end{table*}
\begin{table*}[ht]
  \centering
  \scriptsize
  \renewcommand{\arraystretch}{1.2} 
  \begin{tabular}{ccccccccccc}
    \hline
    Method & AQuA & GSM8K & MultiArith & SingEq& HotpotQA & CSQA & SIQA& Letter& Coin& Average\\
    \hline
    \multicolumn{11}{c}{\textit{Without external knowledge and exemplars}} \\  
    \hline
Zero-shot& 57.4& 86.3& 97.5& 97.2& 87.7& 84.2& 86.6& 55.7& 79.3& 81.3\\
Zero-shot-CoT&	59.0&	86.8&	97.8&	97.0&	87.3&	\underline{83.3}&	\underline{86.4}&	54.8&	79.8&	81.4\\
PS&	61.9&	94.7&	98.2&	97.8&	89.9&	85.3&	87.9&	60.4&	83.3&	84.4\\
QDMRPS&	57.1&	93.9&	98.0&	98.3&	91.2&	85.0&	86.7&	55.1&	84.5&	83.3\\
    \hline
    \multicolumn{11}{c}{\textit{With exemplars}} \\  
    \hline
Manual-CoT&	62.3&	93.8&	97.9&	98.1&	90.9&	87.1&	91.6&	60.8&	89.9&	85.8\\
Auto-CoT&	57.8&	92.0&	98.4&	97.9&	92.4&	84.2&	89.3&	58.3&	88.7&	84.3\\
Complex-CoT& 	60.6&	92.8&	97.9&	98.5&	88.7&	86.4&	90.4&	55.3&	83.6&   83.8\\
Iter-CoT&	64.6&	94.7&	97.7&	97.6&	82.4&	86.8&	92.8&	57.5&	92.2&   85.1\\
ZEUS&	61.5&	90.7&	98.2&	98.6&	88.9&	84.8&	89.6&	62.7&	91.3&   85.1\\
Pattern-CoT&	59.7&	92.8&	98.1&	97.5&	86.2&	84.8&	86.9&	55.8&	88.4&   83.4\\
    \hline
    \multicolumn{11}{c}{\textit{With external knowledge}} \\  
    \hline
KD-CoT&	38.8&	76.0&	83.5&	80.8&	\underline{74.8}&	89.7&	92.5&	29.8&	68.4&   70.5\\
IRCoT&	\underline{35.1}&	78.6&	85.8&	78.3&	93.5&	91.6&	93.8&	35.6&	76.2&   74.3\\

KG-CoT&	35.4&	\underline{72.2}&	\underline{78.6}&	\underline{75.6}&	82.5&	91.5&	92.2&	\underline{25.4}&	\underline{64.0}&     \underline{68.6}\\

    \hline
    \multicolumn{11}{c}{\textit{With both external knowledge and exemplars}} \\  
    \hline
\textbf{CoT-RAG (ours)}&	\textbf{72.3}&	\textbf{99.2}&	\textbf{99.5}&	\textbf{99.8}&	\textbf{98.6}&	\textbf{98.8}&	\textbf{98.1}&	\textbf{64.3}&	\textbf{98.7}&	\textbf{92.2}\\

    \hline
  \end{tabular}
  \caption{\label{GLM-4-flash}
    Accuracy on nine datasets from three categories of reasoning tasks using GLM-4-flash.}

\end{table*}
\begin{table*}[ht]
  \centering
  \scriptsize
  \renewcommand{\arraystretch}{1.2} 
  \begin{tabular}{ccccccccccc}
    \hline
    Method & AQuA & GSM8K & MultiArith & SingEq& HotpotQA & CSQA & SIQA& Letter& Coin& Average\\
    \hline
    \multicolumn{11}{c}{\textit{Without external knowledge and exemplars}} \\  
    \hline
Zero-shot& 86.3& 94.2& 97.7& 97.4& 89.6& 83.2& 89.4& 94.5& 98.6& 92.3\\
Zero-shot-CoT&	86.2&	94.5&	97.3&	97.8&	89.3&	\underline{83.9}&	\underline{89.4}&	94.3&	98.4&	92.3\\
PS&	87.4&	94.8&	98.7&	98.5&	89.4&	86.8&	91.9&	97.1&	98.9&	93.7\\
QDMRPS&	89.0&	95.3&	99.2&	99.5&	90.4&	89.0&	92.3&	97.5&	98.8&	94.6\\
    \hline
\multicolumn{11}{c}{\textit{With exemplars}} \\  
    \hline
Manual-CoT&	91.4&	96.5&	99.0&	99.3&	91.5&	88.9&	92.6&	97.6&	\textbf{100}&	95.2\\
Auto-CoT&	88.7&	95.6&	98.8&	98.8&	91.8&	87.3&	91.7&	96.8&	99.2&	94.3\\
Complex-CoT& 	89.0&	95.3&	97.7&	98.2&	89.7&	85.8&	90.9&	96.8&	98.6&   93.6\\
Iter-CoT&	93.2&	96.1&	99.2&	98.7&	89.5&	90.5&	91.7&	97.3&	99.4&   95.1\\
ZEUS&	92.7&	95.8&	98.2&	98.6&	92.0&	86.7&	91.4&	98.6&	99.2&   94.8\\
Pattern-CoT&	88.3&	95.8&	98.6&	98.3&	86.4&	84.7&	90.4&	96.0&	99.4&   93.1\\
    \hline
    \multicolumn{11}{c}{\textit{With external knowledge}} \\  
    \hline
KD-CoT&	\underline{75.3}&	\underline{81.9}&	87.4&	88.3&	84.8&	93.6&	96.7&	85.2&	\underline{91.5}&   \underline{87.2}\\
IRCoT&	75.9&	84.2&	\underline{83.6}&	86.4&	\underline{83.8}&	93.2&	94.9&	89.7&	92.8&   \underline{87.2}\\

KG-CoT&	77.6&	83.2&	86.8&	\underline{83.6}&	85.9&	93.2&	97.1&	\underline{82.6}&	93.6&   87.1\\
\hline
    \multicolumn{11}{c}{\textit{With both external knowledge and exemplars}} \\  
    \hline

\textbf{CoT-RAG (ours)}&	\textbf{96.5}&	\textbf{98.7}&	\textbf{99.7}&	\textbf{99.0}&	\textbf{98.8}&	\textbf{99.5}&	\textbf{99.6}&	\textbf{99.9}&	\textbf{100}&	\textbf{99.1}\\

    \hline
  \end{tabular}
  \caption{\label{GPT-4o}
    Accuracy on nine datasets from three categories of reasoning tasks using GPT-4o.}

\end{table*}

\begin{table*}[ht]
  \centering
  \scriptsize
  \renewcommand{\arraystretch}{1.2} 
  \begin{tabular}{ccccccccccc}
    \hline
    Method & AQuA & GSM8K & MultiArith & SingEq& HotpotQA & CSQA & SIQA& Letter& Coin& Average\\
    \hline
    \multicolumn{11}{c}{\textit{Without external knowledge and exemplars}} \\  
    \hline
Zero-shot& \textbf{2.58}& \textbf{2.29}& \textbf{1.53}& \textbf{1.42}& \textbf{2.21}& \textbf{1.25}& \textbf{1.42}& \textbf{1.64}& \textbf{1.38}& \textbf{1.75}\\
Zero-shot-CoT& 3.36&	3.03&	1.87&	1.69&	2.43&	1.34&	1.50&	1.78&	1.66&	2.07\\
PS& 5.27&	4.76&	4.43&	3.61&	6.24&	4.52&	4.57&	2.94&	3.24&	4.40\\
QDMRPS& 5.84&	5.73&	4.86&	4.17&	6.78&	5.83&	4.96&	3.37&	4.43&	5.11\\
    \hline
\multicolumn{11}{c}{\textit{With exemplars}} \\  
    \hline
Manual-CoT& 2.63&	2.42&	2.09&	1.73&	3.14&	2.67&	1.97&	2.02&	1.98&	2.29\\
Auto-CoT& 4.04&	3.87&	3.27&	2.08&	3.68&	3.18&	2.14&	2.60&	2.27&	3.01\\
Complex-CoT& 	3.86&	3.31&	2.93&	2.65&	4.18&	2.74&	2.43&	2.18&	2.09&   2.93\\
Iter-CoT&	4.05&	3.79&	3.11&	2.85&	3.94&	3.71&	3.65&	2.67&	2.48&   3.36\\
ZEUS&	7.23&	6.74&	6.52&	6.81&	7.17&	5.61&	5.95&	4.43&	4.61&   6.12\\
Pattern-CoT&	3.64&	3.37&	3.16&	2.94&	3.88&	2.91&	2.61&	2.26&	2.14&   2.99\\
    \hline
    \multicolumn{11}{c}{\textit{With external knowledge}} \\  
    \hline
KD-CoT&	7.34&	7.89&	6.96&	7.61&	8.83&	\underline{8.54}&	\underline{8.37}&	\underline{6.32}&	5.86&   7.52\\
IRCoT&	\underline{8.67}&	\underline{8.12}&	\underline{8.37}&	\underline{7.94}&	\underline{9.06}&	8.32&	9.43&	7.16&	\underline{7.53}&   \underline{8.29}\\

KG-CoT&	8.03&	7.44&	7.11&	7.63&	8.46&	8.26&	7.81&	5.94&	5.81&   7.39\\
\hline
    \multicolumn{11}{c}{\textit{With both external knowledge and exemplars}} \\ 

    \hline

\textbf{CoT-RAG (ours)}& 4.25&	4.21&	4.12&	3.22&	4.97&	4.18&	3.84&	3.24&	2.64&	3.85\\

    \hline
  \end{tabular}
  \caption{\label{Runtime}
    Runtime (sec.) on nine datasets from three categories of reasoning tasks using GPT-4o mini. }

\end{table*}
\appendix

\section{CoT-RAG Algorithm and Time Complexity Analysis}
\label{sec:appendix—5}
 As shown in Algorithm \ref{alg:cot-rag}, the CoT-RAG framework involves three  stages: 1) Knowledge Graph-driven CoT Generation (Lines 1-9), 2) Learnable Knowledge Case-aware RAG (Lines 11-18), and 3) Pseudo-Program Prompting Execution (Lines 20-23). First, experts construct and input a coarse-grained decision tree related to a field (Line 3), and then ask the LLM to decompose the decision tree into a fine-grained, highly structured knowledge graph (Line 4), which is represented by the Pseudo-Program Prompting we proposed (Line 5), referred to as the initialized pseudo-program knowledge graph (\textit{initial\_PKG}). This can further be modified by the experts (Lines 6-8).  Next, users enter query descriptions related to this field (Line 13), which can help update the decision tree benefiting from their usefulness (Lines 14-16). According to the \textit{initial\_PKG}, the LLM subsequently extracts the corresponding sub-descriptions from the query description to update the \textit{initial\_PKG}, namely \textit{updated\_PKG} (Lines 17). Finally, the LLM executes \textit{updated\_PKG} to output the final result (Lines 20-23).

The time complexity of the CoT-RAG algorithm is determined by three aforementioned stages. In \textit{Stage 1}, the algorithm decomposes a coarse-grained decision tree into a fine-grained knowledge graph. The time complexity of this process is \textit{O(N)}, where \textit{N} represents the number of nodes in the decision tree. In \textit{Stage 2}, the algorithm extracts sub-descriptions from user input and updates the knowledge graph. The time complexity of this stage is \textit{O(M)}, with \textit{M} denoting the number of entities in the knowledge graph. In \textit{Stage 3}, the algorithm inputs the updated knowledge graph into LLMs to generate the final result, with a time complexity of \textit{O(M)} as well. Overall, because the number of entities \textit{M} in the knowledge graph has a linear relationship with decision tree nodes' number \textit{N} in our experiments, the total time complexity of the algorithm is \textit{O(N)}.  

\section{Frequently-Used Notations}
\label{sec:appendix—6}
Table \ref{Notation} shows the frequently used notations in this paper.

\section{Analysis}
\label{sec:appendix—7}
We also perform an extensive analysis of CoT-RAG for a deeper insight into the function of each component (\S \ref{sec:Ablation Results}), the robustness of exemplars (\S \ref{sec:Robustness Results}), the relationship between accuracy and reasoning complexity (\S \ref{sec:Complexity}), and the runtime and resource usage (\S \ref{sec:appendix—8}). GPT-4o mini is chosen as the benchmark model for all ensuing analyses. To ensure transparency and reproducibility, we have released
the codebase, datasets, and manually designed decision trees at \url{https://github.com/hustlfy123/CoT-RAG}, enabling the research community to replicate and extend our findings.  
\subsection{Ablation Results w.r.t. CoT-RAG Components}
\label{sec:Ablation Results}
The CoT-RAG framework enhances LLMs' reasoning accuracy. To quantify the contribution of each component to accuracy improvement, we conduct an ablation analysis by removing different parts of the framework. We test four variants on arithmetic reasoning datasets (AQuA, GSM8K) and commonsense reasoning datasets (HotpotQA, CSQA):
\textbf{No node decomposition:} Decision tree nodes are not decomposed into knowledge graphs;
\textbf{No RAG:} Removal of entity \textit{Sub-case} in the knowledge graph;
\textbf{No PsePrompting:} Replace Pseudo-program Prompting with NL prompts;
\textbf{No expert inspection:} The initial PKG is not inspected by an expert.

\begin{figure}[ht]
\centering
  \includegraphics[width=\columnwidth]{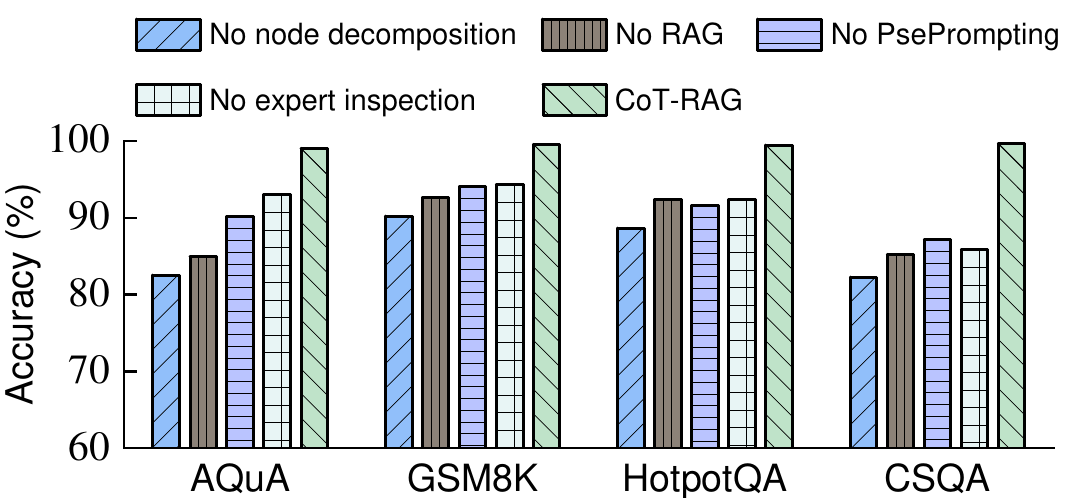}
  \caption{Ablation study results: accuracy when we remove different parts of CoT-RAG.}
  \label{Abation_study}
\end{figure}
Figure \ref{Abation_study} presents the experimental results, showing that accuracy decreases for all variants across datasets. Notably, removing node decomposition leads to the most significant drop in accuracy, highlighting its crucial role in the framework. In GSM8K, HotpotQA, and CSQA, RAG and expert inspection have similar contributions. However, in AQuA, RAG's contribution is evidently higher, probably due to the difficulty of the dataset's mathematical tasks. In Table \ref{ERNIE-Speed-128K}, AQuA's lower accuracy suggests its more challenging tasks, which may require knowledge retrieval. PsePrompting's contribution is moderate, probably because the average complexity of these datasets is low and thus inadequate to manifest its advantages.
\subsection{Robustness Results}
\label{sec:Robustness Results}
Considering that robustness is a key performance indicator, we further conduct a comprehensive robustness analysis on CoT-RAG. In addition to the original CoT-RAG (Table \ref{GPT-4o-mini}), we test several variants to assess the impact of different factors. Specifically, we evaluate the effect of PsePrompting language by using C++ and Java to generate PKG (Appendix \ref{sec:appendix-3}). We also investigate the influence of knowledge cases by manually replacing all \textit{Knowledge case} in the decision tree twice and comparing inference accuracy. Following \citet{wei2022chain}, two other co-authors of this paper (A and B) assume expert roles to contribute to the decision tree design, allowing us to compare the impact of different experts on our method.
\begin{figure}[ht]
\centering
  \includegraphics[width=\columnwidth]{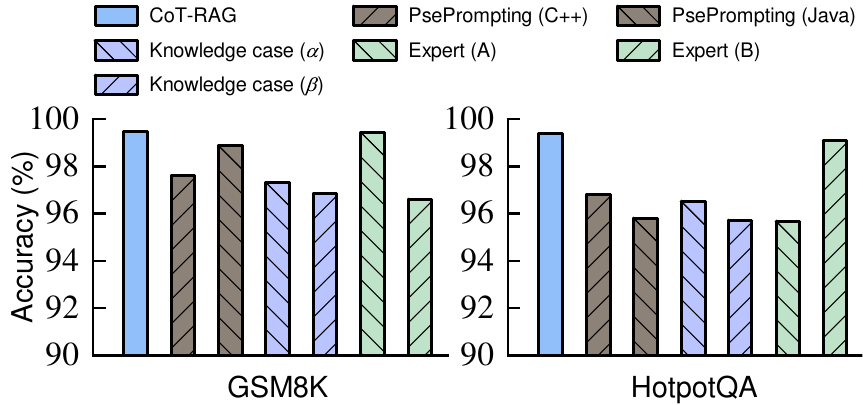}
  \caption{Robustness of CoT-RAG.}
  \label{Robustness}
\end{figure}

Figure \ref{Robustness} presents the experimental results across the GSM8K and HotpotQA datasets for all variants. Analysis shows that compared to the original CoT-RAG, the accuracy error for all variants does not exceed 4\%, demonstrating the strong robustness of CoT-RAG. Furthermore, the participation of Expert A and Expert B caused noticeable accuracy fluctuations, indicating that the alignment between an expert's domain knowledge and the CoT-RAG framework can influence the quality of decision trees and final inference outcomes. This observation highlights the synergy between expert knowledge and the framework. 
\subsection{Accuracy Results w.r.t. Varying Reasoning Complexity}
\label{sec:Complexity}

\begin{figure}[t]
\centering
  \includegraphics[width=0.8\columnwidth]{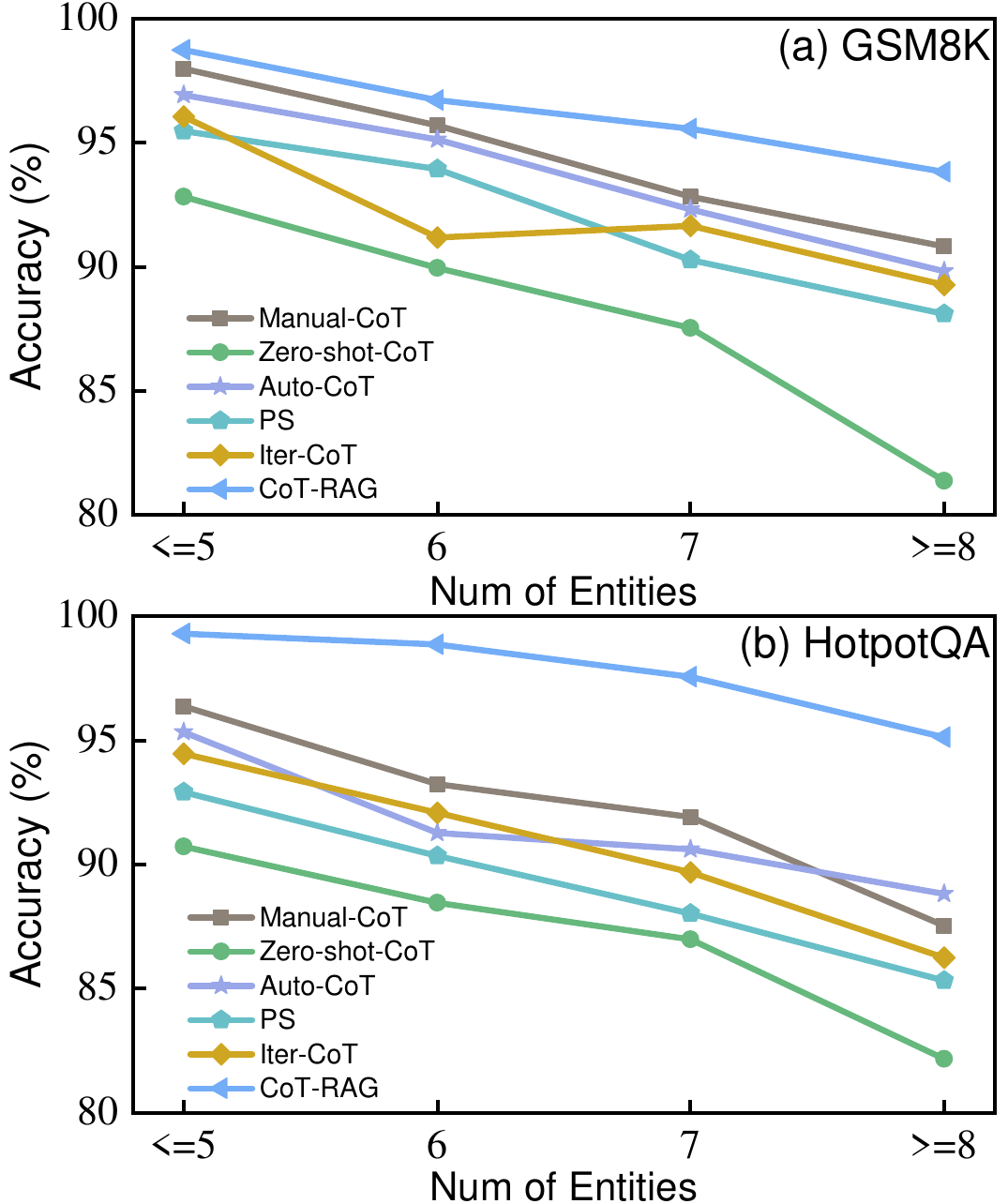}
  \caption{Accuracy of problems that can be decomposed into different numbers of entities.}
  \label{num_entities}
\end{figure}

To further showcase the superiority of our method, we contrast the average accuracies of diverse reasoning complexities on GSM8K (arithmetic reasoning) and HotpotQA (commonsense reasoning). Reasoning complexity is gauged by the entity counts post-original decision tree decomposition. Considering that the maximum number of entities split from the two datasets is 9, we compared the accuracies with four levels of complexity: $\leq$ 5, 6, 7, and $\geq$ 8. According to Table \ref{ERNIE-Speed-128K}, we select representative baselines for comparison: Zero-shot- CoT, PS, Manual-CoT, Auto-CoT, and Iter-CoT. As shown in Figure \ref{num_entities}, for problems with over 8 entities, on GSM8K, CoT-RAG's average accuracy outperforms Manual-CoT by 3.4\%, Zero-shot-CoT by 15.3\%, Auto-CoT by 4.5\%, PS by 6.5\%, and Iter-CoT by 5.1\%. On HotPotQA, the gains are 8.7\%, 15.8\%, 7.1\%, 11.5\%, and 10.3\% respectively, attesting to our method's efficacy in complex reasoning tasks. Moreover, as the number of the entity increases, the superiority of CoT-RAG over the competitors enlarges progressively.

\subsection{Runtime and Resource Usage Results}
\label{sec:appendix—8}
We conduct more comparisons with CoT approaches on average runtime for answering each question in each dataset using GPT-4o mini, where our running time only includes \textit{Stage 2} and \textit{Stage 3} of our CoT-RAG, that is, CoT inference time to answer a user’s question. Table \ref{Runtime} reveals that our method generally outperforms PS, QDMRPS, KD-CoT, IRCoT, KG-CoT, and ZEUS in terms of running time. However, it shows slightly higher runtime compared to Manual-CoT, Zero-shot, Zero-shot-CoT, and Auto-CoT, due to longer text processing workflow. Overall, it ranks in the middle range regarding the running time. 

Additionally, we also compare with the average running time and token resource consumption of the graph-form LLM-based RAG methods. Table \ref{consumption}  indicates that, compared with GraphRAG, CoT-RAG achieves an overall improvement, with 29.2\% reduction in average runtime, and 33.4\% decrease in average token consumption. Compared with the rest of the baselines, although CoT-RAG consumes more tokens, it shows an improvement in the average running time, in the range 13.9\%-63.0\%. Furthermore, Table \ref{accuracy_vertical_domains} indicates that, compared with the baselines, our CoT-RAG has increased accuracy by 8.9\% to 80.6\%, which evidently demonstrates that the benefits far outweigh the costs. 

However, the running time mentioned above does not incorporate \textit{Stage 1} of our CoT-RAG, which consists of two components: the construction of DTs by experts and the decomposition of DTs into KGs by LLMs. Therefore, we conduct a further exploration of the time it takes Expert A and Expert B to design DTs, and subsequently, the running time of LLMs to convert these DTs into detailed KGs, across four vertical domain datasets as presented in Table \ref{Runtimeofstage1}. Analysis demonstrates that the time cost of DT design is significantly influenced by an expert's expertise, while LLM-based KG conversion time is lower. Although \textit{Stage 1} consumes a great deal of time, it only needs to be constructed once in an offline manner. When subsequently faced with thousands of user questions in real time, the running time of our CoT-RAG (\textit{Stage 2} + \textit{Stage 3}) is lower than that of all the baseline methods (Table \ref{consumption}). The more questions from users there are, the more obvious our advantages will be.

Furthermore, in practical applications, experts can conduct desensitization preprocessing (i.e., de-identifying, masking, or replacing sensitive data) when constructing decision trees. Therefore, there is \textbf{no risk of data leakage} during the process in which LLMs decompose the
decision trees to construct knowledge graphs.
\begin{table*}[ht]
  \centering
  \footnotesize
  \renewcommand{\arraystretch}{1.2} 
  \begin{tabular}{ccccccccc}
    \hline
    \multirow{2}{1cm}{Method} &\multicolumn{2}{c}{LawBench}&\multicolumn{2}{c}{LegalBench} & \multicolumn{2}{c}{CFBenchmark} & \multicolumn{2}{c}{AGIEval} \\ 
  &Runtime (s)	&Token	&Runtime (s)	&Token &Runtime (s)	&Token &Runtime (s)	&Token   \\ 
    \hline
\multicolumn{9}{c}{\textit{Graph-form LLM-based RAG methods}} \\  
    \hline
GraphRAG&5.85&	\underline{2883}&4.72&	\underline{1850}&	5.08	&\underline{2026}	&4.89	&\underline{2817}\\
KG-CoT& 6.56&	\textbf{265}& 6.93&	\textbf{235}&	6.84&	\textbf{286}&		7.27&	\textbf{316}\\
ToG&\underline{10.4}&	843&\underline{11.7}&	876&		\underline{9.66}&	728&		\underline{13.5}&	774\\
PoG&8.70&	547&11.2&	529&		9.43&	486&		8.92&	516\\
RRKG&6.84&	583&6.31&	542&		7.24&	647&		7.67&	628\\
RoG& 6.27& 628 &5.92& 583& 5.38& 516 & 6.64& 674 \\
Graph-CoT& 7.24& 1368 &7.82& 1237& 6.73& 1297 &  8.13& 1312 \\
ToG-2& 9.53& 1097&10.9& 987& 8.62& 1128&  10.2& 1035 \\
AtomR& 6.34& 1762&6.92& 1531&  7.08& 1621&  7.86& 1687\\
\hline
\multicolumn{9}{c}{\textit{Variants of CoT-RAG}} \\  
    \hline
    CoT-RAG (IndexFlatL2)& 4.67&	896&4.89&	875&	5.12	&1028	&5.23	&1257\\
CoT-RAG (IndexFlatIP)&4.52&	923&4.96&	815&	4.73	&985	&5.32	&1316\\
CoT-RAG (IndexIVFFlat)&4.04&	908&4.23&	892&	4.09	&1028	&4.53	&1342\\
CoT-RAG (IndexLSH)&	4.49&	988&4.37&	842&4.66	&979	&4.67	&1268\\
CoT-RAG (IndexPQ)&4.32&	1002&4.46&	878&	4.86	&996	&4.89	&1225\\
CoT-RAG (IndexIVFPQ)&4.12&	975&4.25&	866&4.42	&1005	&4.62	&1243\\
CoT-RAG (Zero-expert)&3.14&	1532&\textbf{3.27}&	1346&		3.98&	1653&		4.31&	1826\\
\textbf{CoT-RAG (ours)}&\textbf{3.02}&	1557&3.46&	1298&\textbf{3.95}	&1643	&\textbf{4.11}	&1861\\

    \hline
  \end{tabular}
  \caption{\label{consumption}
    Runtime and token consumption on four datasets from vertical domains  using GPT-4o mini.}
  
\end{table*}
\begin{table*}[ht]
  \centering
  \small
  \renewcommand{\arraystretch}{1.2} 
  \begin{tabular}{ccccccccc}
    \hline
    
    \multirow{2}{1cm}{Expert} & \multicolumn{2}{c}{LawBench} & \multicolumn{2}{c}{LegalBench} & \multicolumn{2}{c}{CFBenchmark} & \multicolumn{2}{c}{AGIEval} \\ 
   &DT design	&DT to KG	&DT design	&DT to KG	&DT design	&DT to KG &DT design	&DT to KG  \\ 
    \hline
    A& 262.3& 5.7& 232.7& 4.3& 179.2& 3.8& 454.8& 7.2\\
    B& 130.8& 6.2& 469.1& 6.9& 58.8& 3.2& 574.3& 6.4\\

    \hline
  \end{tabular}
  \caption{\label{Runtimeofstage1}
    Average running time (sec.) for Experts A and B in decision tree design of various vertical domains and converting DTs to KGs using GPT-4o mini.}
  
\end{table*}
\section{Accuracy Results with Varying LLMs}
\label{sec:appendix—4}
Table \ref{GPT-4o-mini}, Table \ref{ERNIE-3.5-128K}, Table \ref{GLM-4-flash}, and Table \ref{GPT-4o} are the experimental results of GPT-4o mini, ERNIE-3.5-128K, GLM-4-flash, and GPT-4o, respectively.
\section{Example Outputs of the three stages of CoT-RAG}
\label{sec:appendix—1}
Tabel \ref{initial}, Table \ref{complete}, and Table \ref{result} are example outputs for the three stages of CoT-RAG, respectively.

\section{Examples of pseudo-program knowledge graphs from different datasets}
\label{sec:appendix-2}
Tables \ref{AQuA} to \ref{AGIEvalresult} are examples of pseudo-program knowledge graphs and their output results of GPT-4o mini for each dataset, where we can evidently observe that the pseudo-program format exhibits generality and scalability, enabling it to meet the requirements of various domains to great extent. In this study, the pseudo-program is designed based on reasoning chains within knowledge graphs. Its core logic involves guiding LLMs to conduct reasoning through well-defined steps. Although it is not designed with domain-specific syntax, due to its simple logical structure and comprehensibility, it can be flexibly adapted to different domain-specific reasoning paradigms. This is achieved by adjusting elements such as sub-questions, sub-cases, and sub-descriptions of nodes.

Consider the two pseudo-program prompts in Table \ref{LegalBench} and Table \ref{CFBenchmark} as examples. They belong to the law and finance, respectively. We observe that the variable names of these two pseudo-program prompt words are all "sub\_questionX", "sub\_caseX", "sub\_descriptionX", and "answerX", where "X" represents a specific number, and the logical connections are all made around these variable names. Therefore, when the application field shifts from law to finance, one only needs to modify the corresponding variable contents, the number of variables, and the logical relationships in Table \ref{LegalBench}, and then it can be transformed into Table \ref{CFBenchmark}, enabling fully automatic domain adaptation.
\section{Pseudo-program knowledge graph in different programming language forms}
\label{sec:appendix-3}
Tables \ref{C++} to \ref{Javaresult} are examples of pseudo-program knowledge graphs in both C++ and Java languages, along with their output results.
\section{Descriptions of Datasets}
\label{sec:appendix-9}
\textbf{General domains: }\textbf{Arithmetic Reasoning:} (1) the AQUA \citep{ling-etal-2017-program} dataset of algebraic word problems with NL rationales, (2) the GSM8K \citep{2021arXiv211014168C} dataset of high quality linguistically diverse grade school math word problems created by human problem writers, (3) the MultiArith \citep{roy-roth-2015-solving} dataset of mathematical problems that necessitates numerous inference steps for resolution, (4) the SingleEq \citep{koncel-kedziorski-etal-2015-parsing} dataset of algebraic problems calls for the solution of equations; \textbf{Commonsense Reasoning:} (5) the HotpotQA \citep{wolfson-etal-2020-break} dataset of commonsense questions based on Wikipedia that requires reasoning using multiple supporting problems, (6) the CSQA \citep{talmor-etal-2019-commonsenseqa} benchmark dataset of multiple-choice questions that require different types of commonsense knowledge to obtain the correct answers, (7) the SIQA \citep{sap-etal-2019-social} dataset of questions which focus on inferring people's behavior and its social impact; 
\textbf{Symbolic Reasoning:} (8) the Last Letter Concatenation \citep{wei2022chain} dataset of questions requiring the last letters of words in a name to be concatenated (e.g., “Donald Trump”→“dp”), and (9) the Coin Flip \citep{wei2022chain} dataset of questions on whether a coin is still heads up after it is flipped or not flipped based on steps given in the questions. 

\noindent\textbf{Vertical domains: }\textbf{LawBench (LaB)} \citep{fei-etal-2024-lawbench}: A Chinese legal benchmark including tasks such as entity recognition, reading comprehension, and crime amount calculation; \textbf{LegalBench (LeB)} \citep{guha2023legalbench}: A American legal benchmark featuring 162 legal reasoning tasks; \textbf{CFBenchmark (CFB)} \citep{2023arXiv231105812L}: A Chinese financial benchmark assessing the performance of LLMs in finance; \textbf{AGIEval (AGI)} \citep{zhong-etal-2024-agieval}: A benchmark for evaluating human cognitive task performance, where we select the dataset with a focus on logic-based question answering. 

However, in our experiments, the problem categories in existing open-source question-answering datasets are too diverse to directly meet our testing requirements. Thus, following GraphRAG \citep{2024arXiv240416130E} and Graph-CoT \citep{jin-etal-2024-graph}, we utilize five LLMs in § \ref{baselines} to generate datasets that suit our testing needs. Specifically, we first select 200 questions with distinct reasoning logics from each open-source dataset. Then, we prompt each LLM to generate four new questions for each selected question. These new questions had the same reasoning logic but different content. As a result, each domain-specific dataset yields a collection of 200 question sets, each containing 21 questions (one original question plus twenty new questions generated by the five LLMs) with the same reasoning logic, namely a total of 4200 questions for each dataset, whose scale is far larger than the 125 questions per dataset of GraphRAG \citep{2024arXiv240416130E}, the 2579 questions per dataset on average of KD-CoT \citep{2023arXiv230813259W}, the 947 questions per dataset on average of Iter-CoT \citep{sun-etal-2024-enhancing}, the 618 questions per dataset on average of Pattern-CoT \citep{Zhang_Wang_Wu_Wang_2025} , the 500 questions per dataset on average of IRCoT \citep{trivedi-etal-2023-interleaving}, the 174 questions per domain on average of Graph-CoT \citep{jin-etal-2024-graph}, and the 1334 questions per dataset on average of PoG \citep{10.1145/3696410.3714892}, evincing the substantial magnitude and adequacy of our datasets. The specific prompt is shown in Table \ref{prompt} and the generated datasets are available at \url{https://github.com/hustlfy123/CoT-RAG}.

\begin{table*}[t]
  \centering
  \small
  \begin{tabular}{p{\linewidth}}
    \hline
    def LLM(case,question,description): \\
    \quad LLM\_answer=\textcolor{orange}{Given the question, the case is an example about the question. Please study this example and answer the question based on the description} \\
    \quad return LLM\_answer \\
    \\
    \textcolor{blue}{sub\_question1}="How many lunch trays can the person carry at once?"\\
    \textcolor[rgb]{0.1333, 0.5451, 0.1333}{sub\_case1}="  \\
    Description: Tom is helping the restaurant staff collect dishes and he can only carry 6 plates at a time.\\
    Answer: Tom can carry 6 plates at a time."\\
    \textcolor[rgb]{1, 0, 0}{sub\_description1}=" "\\
    \textcolor[rgb]{1, 0.078, 0.57}{answer1}=LLM(sub\_case1,
    sub\_question1,sub\_description1)\\
    \\
    \textcolor{blue}{sub\_question2}="How many trays does the person need to pick up from the first table?"\\
    \textcolor[rgb]{0.1333, 0.5451, 0.1333}{sub\_case2}="\\
    Description: Tom needs to pick up 15 plates from one table and 9 plates from another.\\
    Answer: Tom needs to pick up 15 plates from the first table."\\
    \textcolor[rgb]{1, 0, 0}{sub\_description2}=" "\\
    \textcolor[rgb]{1, 0.078, 0.57}{answer2}=LLM(sub\_case2,sub\_question2,sub\_description2)\\
    \\
    \textcolor{blue}{sub\_question3}="How many trays does the person need to pick up from the second table?"\\
    \textcolor[rgb]{0.1333, 0.5451, 0.1333}{sub\_case3}="\\
    Description: Tom needs to pick up 15 plates from one table and 9 plates from another.\\
    Answer: Tom needs to pick up 9 plates from the second table."\\
    \textcolor[rgb]{1, 0, 0}{sub\_description3}=" "\\
    \textcolor[rgb]{1, 0.078, 0.57}{answer3}=LLM(sub\_case3,sub\_question3,sub\_description3)\\
    \\
    \textcolor{blue}{sub\_question4}="How many trays does the person need to carry in total?"\\
    \textcolor[rgb]{0.1333, 0.5451, 0.1333}{sub\_case4}="\\
    Description: Tom needs to pick up 15 plates from the first table and 9 plates from the second table.\\
    Answer: 15+9=24. Therefore Tom needs to carry 24 plates in total."\\
    \textcolor[rgb]{0.72, 0.52, 0.04}{\#Answering sub\_question4 requires relying on the answers of sub\_question2 and sub\_question3, namely answer2 and answer3}\\
    \textcolor[rgb]{1, 0, 0}{sub\_description4}=\textcolor[rgb]{1, 0.078, 0.57}{answer2}+\textcolor[rgb]{1, 0.078, 0.57}{answer3}\\
    \textcolor[rgb]{1, 0.078, 0.57}{answer4}=LLM(sub\_case4,sub\_question4,sub\_description4)\\
    \\
    \textcolor{blue}{sub\_question5}="How many trips will the person need to make to carry all the trays?"\\
    \textcolor[rgb]{0.1333, 0.5451, 0.1333}{sub\_case5}="\\
    Description: Tom can carry 6 plates at a time and Tom needs to carry 24 plates in total.\\
    Answer: 24/6=4. Therefore Tom needs to carry 24 plates in total."\\
    \textcolor[rgb]{0.72, 0.52, 0.04}{\#Answering sub\_question5 requires relying on the answers of sub\_question1 and sub\_question4, namely answer1 and answer4}\\
    \textcolor[rgb]{1, 0, 0}{sub\_description5}=\textcolor[rgb]{1, 0.078, 0.57}{answer1}+\textcolor[rgb]{1, 0.078, 0.57}{answer4}\\
    \textcolor[rgb]{1, 0.078, 0.57}{answer5}=LLM(sub\_case5,sub\_question5,sub\_description5)\\
    \\
    \textcolor[rgb]{0.72, 0.52, 0.04}{\#The final answer}\\
    \textcolor[rgb]{0, 0.39, 0}{final\_answer}=\textcolor[rgb]{1, 0.078, 0.57}{answer5}\\
    print(final\_answer)\\
    \hline
  \end{tabular}
  \caption{\label{initial}
    The initialized pseudo-program knowledge graph on MultiArith, where entity 1, 2, and 3’s \textit{Sub-description} are \textbf{null} values. (Question: Roger was helping the cafeteria workers pick up lunch trays, but he could only carry 4 trays at a time. If he had to pick up 10 trays from one table and 2 trays from another, how many trips will he make?)
  }
\end{table*}

\begin{table*}[t]
  \centering
  \small
  \begin{tabular}{p{\linewidth}}
    \hline
    \textcolor[rgb]{0.72, 0.52, 0.04}{\#Given a program text, your role is the LLM function,which has three parameters: case, question, and description. The description is textual, the question requires you to answer based on the description, and the case involves processed scenarios or supplementary information or related considerations to assist you in answering the question. You need to strictly follow the program logic to execute. During execution, the LLM function will be called multiple times, which means you will answer corresponding questions based on different cases and descriptions. Please output the final result of this program text in natural language.}\\
    \\
    def LLM(case,question,description): \\
    \quad LLM\_answer=\textcolor{orange}{Given the question, the case is an example about the question. Please study this example and answer the question based on the description} \\
    \quad return LLM\_answer \\
    \\
    \textcolor{blue}{sub\_question1}="How many lunch trays can the person carry at once?"\\
    \textcolor[rgb]{0.1333, 0.5451, 0.1333}{sub\_case1}="  \\
    Description: Tom is helping the restaurant staff collect dishes, but he can only carry 6 plates at a time.\\
    Answer: Tom can carry 6 plates at a time."\\
    \textcolor[rgb]{1, 0, 0}{sub\_description1}=" \textcolor[rgb]{0.11, 0.56, 1}{Roger can only carry 4 trays at a time.} "\\
    \textcolor[rgb]{1, 0.078, 0.57}{answer1}=LLM(sub\_case1,
    sub\_question1,sub\_description1)\\
    \\
    \textcolor{blue}{sub\_question2}="How many trays does the person need to pick up from the first table?"\\
    \textcolor[rgb]{0.1333, 0.5451, 0.1333}{sub\_case2}="\\
    Description: Tom needs to pick up 15 plates from one table and 9 plates from another.\\
    Answer: Tom needs to pick up 15 plates from the first table."\\
    \textcolor[rgb]{1, 0, 0}{sub\_description2}="\textcolor[rgb]{0.11, 0.56, 1}{Roger have to pick up 10 trays from the first table.} "\\
    \textcolor[rgb]{1, 0.078, 0.57}{answer2}=LLM(sub\_case2,sub\_question2,sub\_description2)\\
    \\
    \textcolor{blue}{sub\_question3}="How many trays does the person need to pick up from the second table?"\\
    \textcolor[rgb]{0.1333, 0.5451, 0.1333}{sub\_case3}="\\
    Description: Tom needs to pick up 15 plates from one table and 9 plates from another.\\
    Answer: Tom needs to pick up 9 plates from the second table."\\
    \textcolor[rgb]{1, 0, 0}{sub\_description3}="\textcolor[rgb]{0.11, 0.56, 1}{Roger have to pick up 2 trays from the second table.} "\\
    \textcolor[rgb]{1, 0.078, 0.57}{answer3}=LLM(sub\_case3,sub\_question3,sub\_description3)\\
    \\
    \textcolor{blue}{sub\_question4}="How many trays does the person need to carry in total?"\\
    \textcolor[rgb]{0.1333, 0.5451, 0.1333}{sub\_case4}="\\
    Description: Tom needs to pick up 15 plates from the first table and 9 plates from the second table.\\
    Answer: 15+9=24. Therefore Tom needs to carry 24 plates in total."\\
    \textcolor[rgb]{0.72, 0.52, 0.04}{\#Answering sub\_question4 requires relying on the answers of sub\_question2 and sub\_question3, namely answer2 and answer3}\\
    \textcolor[rgb]{1, 0, 0}{sub\_description4}=\textcolor[rgb]{1, 0.078, 0.57}{answer2}+\textcolor[rgb]{1, 0.078, 0.57}{answer3}\\
    \textcolor[rgb]{1, 0.078, 0.57}{answer4}=LLM(sub\_case4,sub\_question4,sub\_description4)\\
    \\
    \textcolor{blue}{sub\_question5}="How many trips will the person need to make to carry all the trays?"\\
    \textcolor[rgb]{0.1333, 0.5451, 0.1333}{sub\_case5}="\\
    Description: Tom can carry 6 plates at a time and Tom needs to carry 24 plates in total.\\
    Answer: 24/6=4. Therefore Tom needs to carry 24 plates in total."\\
    \textcolor[rgb]{0.72, 0.52, 0.04}{\#Answering sub\_question5 requires relying on the answers of sub\_question1 and sub\_question4, namely answer1 and answer4}\\
    \textcolor[rgb]{1, 0, 0}{sub\_description5}=\textcolor[rgb]{1, 0.078, 0.57}{answer1}+\textcolor[rgb]{1, 0.078, 0.57}{answer4}\\
    \textcolor[rgb]{1, 0.078, 0.57}{answer5}=LLM(sub\_case5,sub\_question5,sub\_description5)\\
    \\
    \textcolor[rgb]{0.72, 0.52, 0.04}{\#The final answer}\\
    \textcolor[rgb]{0, 0.39, 0}{final\_answer}=\textcolor[rgb]{1, 0.078, 0.57}{answer5}\\
    print(final\_answer)\\
    \hline
  \end{tabular}
  \caption{\label{complete}
    The updated pseudo-program knowledge graph on MultiArith, where entity 1, 2, and 3’s \textit{Sub-description} are \textbf{assigned} values. (Question: Roger was helping the cafeteria workers pick up lunch trays, but he could only carry 4 trays at a time. If he had to pick up 10 trays from one table and 2 trays from another, how many trips will he make?)
  }
\end{table*}

\begin{table*}[t]
  \centering
  \small
  \begin{tabular}{p{\linewidth}}
    \hline
    
To derive the final answer step by step, we will process the given sub-questions, descriptions, cases, and collect the responses according to the logic of the program.\\
\\
1. Subquestion 1:\\
  - Question: How many lunch trays can the person carry at once?\\
  - Description: Roger can only carry 4 trays at a time.\\
  - Case: The example describes Tom, who can carry 6 plates at a time.\\
  - Answer: Roger can carry 4 trays at a time.\\
\\
2. Subquestion 2:\\
  - Question: How many trays does the person need to pick up from the first table?\\
  - Description: Roger have to pick up 10 trays from the first table.\\
  - Case: The example describes Tom needing to pick up 15 plates from one table and 9 plates from another.\\
  - Answer: Roger needs to pick up 10 trays from the first table.\\
\\
3. Subquestion 3:\\
  - Question: How many trays does the person need to pick up from the second table?\\
  - Description: Roger have to pick up 2 trays from the second table.\\
  - Case: The example describes Tom needing to pick up 15 plates from one table and 9 plates from another.\\
  - Answer: Roger needs to pick up 2 trays from the second table.\\
\\
4. Subquestion 4:\\
  - Question: How many trays does the person need to carry in total?\\
  - Description: Based on answers from subquestions 2 and 3, Roger's total tray count is 10 (from the first table) + 2 (from the second table).\\
  - Case: The example explains how Tom calculates his total number of plates.\\
  - Answer: 10 + 2 = 12. Therefore, Roger needs to carry 12 trays in total.\\
\\
5. Subquestion 5:\\
  - Question: How many trips will the person need to make to carry all the trays?\\
  - Description: Based on answers from subquestions 1 and 4, Roger can carry 4 trays at a time and he needs to carry 12 trays in total.\\
  - Case: The example outlines how Tom calculates the number of trips needed.\\
  - Answer: 12 (total trays) / 4 (trays per trip) = 3 trips. Therefore, Roger needs to make 3 trips to carry all the trays.\\
\\
Finally, by compiling all the answers, we find that:\\
\\
Final Answer: Roger needs to make 3 trips to carry all the trays.\\

    \hline
  \end{tabular}
  \caption{\label{result}
    The output of executing the pseudo-program knowledge graph using GPT-4o mini in Table \ref{complete}.
  }
\end{table*}

\begin{table*}[t]
  \centering
  \small
  \begin{tabular}{p{\linewidth}}
    \hline
\textcolor[rgb]{0.72, 0.52, 0.04}{\#Given a program text, your role is the LLM function,which has three parameters: case, question, and description. The description is textual, the question requires you to answer based on the description, and the case involves processed scenarios or supplementary information or related considerations to assist you in answering the question. You need to strictly follow the program logic to execute. During execution, the LLM function will be called multiple times, which means you will answer corresponding questions based on different cases and descriptions. Please output the final result of this program text in natural language.}\\
\\
def LLM(case,question,description):\\
    \quad LLM\_answer=\textcolor{orange}{Given the question, the case is an example about the question. Please study this example and answer the question based on the description.}\\
    \quad return LLM\_answer\\
\\
\textcolor{blue}{sub\_question1}="Assuming the speed of the first person is v kilometers per hour, what is the speed of the second person in kilometers per hour?"\\
\textcolor[rgb]{0.1333, 0.5451, 0.1333}{sub\_case1}="  
Description: Cyclist A's rate is 20\% faster than Cyclist B's.\\
Answer: Given that Cyclist A has a speed 20\% faster than Cyclist B, the speed of Cyclone A is v+0.2v=1.2v kilometers per hour."\\
\textcolor[rgb]{1, 0, 0}{sub\_description1}="\textcolor[rgb]{0.11, 0.56, 1}{Friend P's rate is 15\% faster than Friend Q's.}"\\
\textcolor[rgb]{1, 0.078, 0.57}{answer1}=LLM(sub\_case1,sub\_question1,sub\_description1)\\
\\
\textcolor{blue}{sub\_question2}="How to express the time t (expressed in an equation containing v) when two people start from both ends and walk towards each other at the same time?"\\
\textcolor[rgb]{0.1333, 0.5451, 0.1333}{sub\_case2}="\\
Description:Two cyclists plan to ride along a 60-km route, starting at opposite ends of the route at the same time. The speed of Cyclist A is 1.2v, and the speed of Cyclist B is v.\\
Answer: Two people are walking towards each other, with a total distance of 60 kilometers. According to the formula for the encounter problem, "Encounter time=total distance ÷ sum of two people's speeds", the sum of two people's speeds is v+1.2v=2.2v kilometers per hour, so the encounter time t=60/2.2v=300/11v hours"\\
\textcolor[rgb]{0.72, 0.52, 0.04}{\#Answering sub\_question2 requires relying on the answer of sub\_question1 , namely answer1}\\
\textcolor[rgb]{1, 0, 0}{sub\_description2}="\textcolor[rgb]{0.11, 0.56, 1}{Two friends plan to walk along a 43-km trail, starting at opposite ends of the trail at the same time.}"+\textcolor[rgb]{1, 0.078, 0.57}{answer1}\\
\textcolor[rgb]{1, 0.078, 0.57}{answer2}=LLM(sub\_case2,sub\_question2,sub\_description2)\\
\\
\textcolor{blue}{sub\_question3}="How many kilometers will the first person walk when they meet?"\\
\textcolor[rgb]{0.1333, 0.5451, 0.1333}{sub\_case3}="\\
Description: The speed of Cyclist A is 1.2v and  the encounter time t=60/2.2v=300/11v hours.\\
Answer: The speed of Cyclone A is 1.2v kilometers per hour, and the encounter time is 300/11v hours. According to the distance=speed x time, the distance traveled by Cyclone A is S=1.2v x 300/11v=360/11=32.73 kilometers."\\
\textcolor[rgb]{0.72, 0.52, 0.04}{\#Answering sub\_question3 requires relying on the answers of sub\_question1 and sub\_question2, namely answer1 and answer2}\\
\textcolor[rgb]{1, 0, 0}{sub\_description3}=\textcolor[rgb]{1, 0.078, 0.57}{answer1}+\textcolor[rgb]{1, 0.078, 0.57}{answer2}\\
\textcolor[rgb]{1, 0.078, 0.57}{answer3}=LLM(sub\_case3,sub\_question3,sub\_description3)\\
\\
\textcolor[rgb]{0.72, 0.52, 0.04}{\#The final answer}\\
\textcolor[rgb]{0, 0.39, 0}{final\_answer}=\textcolor[rgb]{1, 0.078, 0.57}{answer3}\\
print(final\_answer)\\

    \hline
  \end{tabular}
  \caption{\label{AQuA}
    The updated pseudo-program knowledge graph on AQuA. (Question: Two friends plan to walk along a 43-km trail, starting at opposite ends of the trail at the same time. If Friend P's rate is 15\% faster than Friend Q's, how many kilometers will Friend P have walked when they pass each other?)
  }
\end{table*}

\begin{table*}[t]
  \centering
  \small
  \begin{tabular}{p{\linewidth}}
    \hline
Let's break down the program logic step by step and determine the final answer.\\
\\
Step-by-Step Breakdown:\\
\\
1. First Call to LLM (sub\_question1):\\
   - Question: Assuming the speed of the first person is  v kilometers per hour, what is the speed of the second person in kilometers per hour?\\
   - Case: Cyclist A's rate is 20\% faster than Cyclist B's.\\
   - Description: Friend P's rate is 15\% faster than Friend Q's.\\
   \\
   Answer: \\
   - Since Friend P's rate is 15\% faster than Friend Q's, the speed of Friend P is v + 0.15v = 1.15v  kilometers per hour.\\
   \\
   answer1 = "1.15v"\\
\\
2. Second Call to LLM (sub\_question2):\\
   - Question: How to express the time  t  (expressed in an equation containing  v ) when two people start from both ends and walk towards each other at the same time?\\
   - Case: Two cyclists plan to ride along a 60-km route, starting at opposite ends of the route at the same time. The speed of Cyclist A is  1.2v , and the speed of Cyclist B is  v .\\
   - Description: Two friends plan to walk along a 43-km trail, starting at opposite ends of the trail at the same time. \\
   - The description from 'answer1' tells us that Friend P's speed is  1.15v  (instead of 1.2v), and Friend Q's speed is  v .\\
   \\
   Answer:\\
   - The total distance is 43 kilometers.\\
   - The sum of the speeds is  v + 1.15v = 2.15v .\\
   - The encounter time is  t = Total distance / Sum of the speeds = 43/2.15v.\\
   \\
   answer2 = "43 / 2.15v"\\
\\
3. Third Call to LLM (sub\_question3):\\
   - Question: How many kilometers will the first person walk when they meet?\\
   - Case: The speed of Cyclist A is  1.2v  and the encounter time  t = 300/11v hours.\\
   - Description: The description from 'answer1' tells us the speed of Friend P is  1.15v , and 'answer2' gives the encounter time equation as  t = 43/2.15v.\\
   \\
   Answer:\\
   - The speed of Friend P is  1.15v  km/h.\\
   - The encounter time is  t = 43/2.15v hours.\\
   - The distance walked by Friend P is:\\
   \\
   Distance = Speed ×Time = 1.15v × 43/2.15v = 49.45 × 2.15 = 23  kilometers.\\
   \\
   answer3 = "23 kilometers"\\
\\
 Final Answer:\\
The first person (Friend P) will walk 23 kilometers when they meet.\\
\\
 Final Output: 23 kilometers\\

    \hline
  \end{tabular}
  \caption{\label{AQuAresult}
    The output of executing the pseudo-program knowledge graph in Table \ref{AQuA}.
  }
\end{table*}

\begin{table*}[t]
  \centering
  \small
  \begin{tabular}{p{\linewidth}}
    \hline
\textcolor[rgb]{0.72, 0.52, 0.04}{\#Given a program text, your role is the LLM function,which has three parameters: case, question, and description. The description is textual, the question requires you to answer based on the description, and the case involves processed scenarios or supplementary information or related considerations to assist you in answering the question. You need to strictly follow the program logic to execute. During execution, the LLM function will be called multiple times, which means you will answer corresponding questions based on different cases and descriptions. Please output the final result of this program text in natural language.}\\
\\
def LLM(case,question,description):\\
    \quad LLM\_answer=\textcolor{orange}{Given the question, the case is an example about the question. Please study this example and answer the question based on the description.}\\
    \quad return LLM\_answer\\
\\
\textcolor{blue}{sub\_question1}="How many times does the person write letters to each friend per week?"\\
\textcolor[rgb]{0.1333, 0.5451, 0.1333}{sub\_case1}=" \\
Description: Tom writes a letter to 4 different friends once a week.
Answer: Tom writes to each friend once a week."\\
\textcolor[rgb]{1, 0, 0}{sub\_description1}="\textcolor[rgb]{0.11, 0.56, 1}{James writes a letter to 2 different friends twice a week.}"\\
\textcolor[rgb]{1, 0.078, 0.57}{answer1}=LLM(sub\_case1,sub\_question1,sub\_description1)
\\
\textcolor{blue}{sub\_question2}="How many letters does the person write per week?"\\
\textcolor[rgb]{0.1333, 0.5451, 0.1333}{sub\_case2}="\\
Description:Tom writes a letter to 4 different friends once a week and he writes to each friend once a week.\\
Answer: Because there are 4 different friends and I write letters to each friend once a week, the total number of letters written per week is 4 x 1=4 letters."\\
\textcolor[rgb]{0.72, 0.52, 0.04}{\#Answering sub\_question2 requires relying on the answer of sub\_question1 , namely answer1}\\
\textcolor[rgb]{1, 0, 0}{sub\_description2}="\textcolor[rgb]{0.11, 0.56, 1}{James writes a letter to 2 different friends.}"+\textcolor[rgb]{1, 0.078, 0.57}{answer1}\\
\textcolor[rgb]{1, 0.078, 0.57}{answer2}=LLM(sub\_case2,sub\_question2,sub\_description2)\\
\\
\textcolor{blue}{sub\_question3}="What is the total number of pages written by the person per week?"\\
\textcolor[rgb]{0.1333, 0.5451, 0.1333}{sub\_case3}="\\
Description: The total number of letters written by Tom per week is 4 letters and each letter is 5 pages long.\\
Answer: Given that each letter has 5 pages, a total of 4 letters are written per week. Based on the total number of pages=the number of pages per letter x the total number of letters, the total number of pages written per week is 5 x 4=20 pages."\\
\textcolor[rgb]{0.72, 0.52, 0.04}{\#Answering sub\_question3 requires relying on the answers of sub\_question2, namely answer2}\\
\textcolor[rgb]{1, 0, 0}{sub\_description3}="\textcolor[rgb]{0.11, 0.56, 1}{Each letter is 3 pages long.}"+\textcolor[rgb]{1, 0.078, 0.57}{answer2}\\
\textcolor[rgb]{1, 0.078, 0.57}{answer3}=LLM(sub\_case3,sub\_question3,sub\_description3)\\
\\
\textcolor{blue}{sub\_question4}="What is the total number of pages written by the person a year?"\\
\textcolor[rgb]{0.1333, 0.5451, 0.1333}{sub\_case4}="\\
Description: One year is about 52 weeks and the total number of pages written by Tom per week is 20 pages.\\
Answer: Writing 20 pages per week, and knowing that there are approximately 52 weeks in a year, according to the total number of pages per year=pages per week x weeks per year, the total number of pages written in a year is 1040 pages."\\
\textcolor[rgb]{0.72, 0.52, 0.04}{\#Answering sub\_question4 requires relying on the answers of sub\_question3, namely answer3}\\
\textcolor[rgb]{1, 0, 0}{sub\_description4}=\textcolor[rgb]{1, 0.078, 0.57}{answer3}\\
\textcolor[rgb]{1, 0.078, 0.57}{answer4}=LLM(sub\_case4,sub\_question4,sub\_description4)\\
\\
\textcolor[rgb]{0.72, 0.52, 0.04}{\#The final answer}\\
\textcolor[rgb]{0, 0.39, 0}{final\_answer}=\textcolor[rgb]{1, 0.078, 0.57}{answer4}\\
print(final\_answer)\\

\hline
  \end{tabular}
  \caption{\label{GSM8K}
    The updated pseudo-program knowledge graph on GSM8K. (Question: James writes a 3-page letter to 2 different friends twice a week.  How many pages does he write a year?)
  }
\end{table*}

\begin{table*}[t]
  \centering
  \small
  \begin{tabular}{p{\linewidth}}
    \hline
Let's break down the program step by step and follow the logic to find the final answer.\\
\\
Step-by-Step Breakdown:\\
\\
1. First Call to LLM (sub\_question1):\\
   - Question: How many times does the person write letters to each friend per week?\\
   - Case: The case provides an example where Tom writes a letter to 4 different friends once a week.\\
   - Description: James writes a letter to 2 different friends twice a week.\\
\\
   Answer: Based on the description, James writes a letter to each friend twice a week.\\
\\
   answer1 = "twice"\\
\\
2. Second Call to LLM (sub\_question2):\\
   - Question: How many letters does the person write per week?\\
   - Case: The case tells us that Tom writes a letter to 4 different friends once a week.\\
   - Description: The description tells us James writes a letter to 2 different friends, and **twice** a week (from 'answer1').\\
   \\
   Calculation:\\
   - James writes 2 letters (one to each friend) twice a week, so the total number of letters James writes per week is:\\
   \\
   2 × 2  = 4 letters\\
\\
   answer2 = "4 letters"\\
\\
3. Third Call to LLM (sub\_question3):\\
   - Question: What is the total number of pages written by the person per week?\\
   - Case: This case mentions that Tom writes 4 letters per week, and each letter is 5 pages long.\\
   - Description: The description tells us that each letter is 3 pages long (adjusted for James based on 'answer2').\\
   \\
   Calculation:\\
   - James writes 4 letters per week, and each letter is 3 pages long.\\
   - Total pages per week:\\
   \\
   4 ×3 = 12 pages per week
\\
   answer3 = "12 pages"\\
\\
4. Fourth Call to LLM (sub\_question4):\\
   - Question: What is the total number of pages written by the person in a year?\\
   - Case: The case tells us that one year is approximately 52 weeks, and the total number of pages written per week is 12 pages (from 'answer3').\\
   - Description: We use the same number of pages per week (12), and multiply by the number of weeks in a year.\\
   \\
   Calculation:\\
   - Total pages per year:\\
   \\
   12 × 52  = 624 pages per year\\
\\
   answer4 = "624 pages"\\
\\
Final Answer:\\
The total number of pages James writes in a year is 624 pages.\\
\\
Final Output: 624 pages\\

    \hline
  \end{tabular}
  \caption{\label{GSM8Kresult}
    The output of executing the pseudo-program knowledge graph in Table \ref{GSM8K}.
  }
\end{table*}

\begin{table*}[t]
  \centering
  \small
  \begin{tabular}{p{\linewidth}}
    \hline
\textcolor[rgb]{0.72, 0.52, 0.04}{\#Given a program text, your role is the LLM function,which has three parameters: case, question, and description. The description is textual, the question requires you to answer based on the description, and the case involves processed scenarios or supplementary information or related considerations to assist you in answering the question. You need to strictly follow the program logic to execute. During execution, the LLM function will be called multiple times, which means you will answer corresponding questions based on different cases and descriptions. Please output the final result of this program text in natural language.}\\
\\
def LLM(case,question,description):\\
    \quad LLM\_answer=\textcolor{orange}{Given the question, the case is an example about the question. Please study this example and answer the question based on the description.}\\
    \quad return LLM\_answer\\
\\
\textcolor{blue}{sub\_question1}="What was the total number of shells initially found by the person?"\\
\textcolor[rgb]{0.1333, 0.5451, 0.1333}{sub\_case1}=" \\
Description: Lily picked up 80 seashells by the seaside.\\
Answer: The total number of shells initially found by Lily initially was 80."\\
\textcolor[rgb]{1, 0, 0}{sub\_description1}="\textcolor[rgb]{0.11, 0.56, 1}{Joan found 70 seashells on the beach.}"\\
\textcolor[rgb]{1, 0.078, 0.57}{answer1}=LLM(sub\_case1,sub\_question1,sub\_description1)\\
\\
\textcolor{blue}{sub\_question2}="How many shells does this person have left?"\\
\textcolor[rgb]{0.1333, 0.5451, 0.1333}{sub\_case2}="\\
Description: Lily has 32 seashells left.\\
Answer: Lily has 32 seashells left."\\
\textcolor[rgb]{1, 0, 0}{sub\_description2}="\textcolor[rgb]{0.11, 0.56, 1}{Joan has 27 seashell left.}"\\
\textcolor[rgb]{1, 0.078, 0.57}{answer2}=LLM(sub\_case2,sub\_question2,sub\_description2)\\
\\
\textcolor{blue}{sub\_question3}="How many shells did this person give to others?"\\
\textcolor[rgb]{0.1333, 0.5451, 0.1333}{sub\_case3}="\\
Description: The total number of shells initially found by Lily initially was 80 and she has 32 seashells left.\\
Answer: Subtracting the remaining number of shells from Lily's initial total number of shells gives her the number of shells she gives to others. That is to say, the number of shells given=the initial total number of shells - the remaining number of shells, that is, 80-32=48 (pieces)."\\
\textcolor[rgb]{0.72, 0.52, 0.04}{\#Answering sub\_question3 requires relying on the answers of sub\_question1 and sub\_question2, namely answer1 and answer2}\\
\textcolor[rgb]{1, 0, 0}{sub\_description3}=\textcolor[rgb]{1, 0.078, 0.57}{answer1}+\textcolor[rgb]{1, 0.078, 0.57}{answer2}\\
\textcolor[rgb]{1, 0.078, 0.57}{answer3}=LLM(sub\_case3,sub\_question3,sub\_description3)\\

\textcolor[rgb]{0.72, 0.52, 0.04}{\#The final answer}\\
\textcolor[rgb]{0, 0.39, 0}{final\_answer}=\textcolor[rgb]{1, 0.078, 0.57}{answer3}\\
print(final\_answer)\\

      \hline
  \end{tabular}
  \caption{\label{SingleEq}
    The updated pseudo-program knowledge graph on SingleEq. (Question: Joan found 70 seashells on the beach. she gave Sam some of her seashells. She has 27 seashell left. How many seashells did she give to Sam ?)
  }
\end{table*}

\begin{table*}[t]
  \centering
  \small
  \begin{tabular}{p{\linewidth}}
    \hline
Let's walk through the program step by step to determine the final answer.\\
\\
Step-by-Step Breakdown:\\
\\
1. First Call to LLM (sub\_question1):\\
   - Question: What was the total number of shells initially found by the person?\\
   - Case: The case provides an example where Lily initially found 80 seashells.\\
   - Description: Joan found 70 seashells on the beach.\\
\\
   Answer: The total number of shells initially found by Joan was 70, based on the description provided.\\
\\
   answer1 = "70"
\\
2. Second Call to LLM (sub\_question2):\\
   - Question: How many shells does this person have left?\\
   - Case: The case provides an example where Lily has 32 seashells left.
   - Description: Joan has 27 seashells left.\\
\\
   Answer: According to the description, Joan has 27 seashells left.\\
\\
   answer2 = "27"\\
\\
3. Third Call to LLM (sub\_question3):\\
   - Question: How many shells did this person give to others?\\
   - Case: This case involves calculating how many shells were given by Lily based on the initial number of shells and how many are left. The formula provided is:\\
\\
     Shells given = Initial shells - Remaining shells\\
    \\
   - Description: We are given the answers from previous questions: Joan initially found 70 shells and has 27 left.\\
\\
   Calculation:\\
   - Initially, Joan found 70 seashells.\\
   - She has 27 seashells left.\\
   - Therefore, Joan gave away 70 - 27 = 43 seashells.\\
\\
   answer3 = "43"\\
\\
4. Final Answer:\\
   The final answer is "43," indicating how many shells Joan gave to others.\\
\\
Final Output: 43\\

    \hline
  \end{tabular}
  \caption{\label{SingleEqresult}
    The output of executing the pseudo-program knowledge graph in Table \ref{SingleEq}.
  }
\end{table*}

\begin{table*}[t]
  \centering
  \small
  \begin{tabular}{p{\linewidth}}
    \hline
\textcolor[rgb]{0.72, 0.52, 0.04}{\#Given a program text, your role is the LLM function,which has three parameters: case, question, and description. The description is textual, the question requires you to answer based on the description, and the case involves processed scenarios or supplementary information or related considerations to assist you in answering the question. You need to strictly follow the program logic to execute. During execution, the LLM function will be called multiple times, which means you will answer corresponding questions based on different cases and descriptions. Please output the final result of this program text in natural language.}\\
\\
def LLM(case,question,description):\\
    \quad LLM\_answer=\textcolor{orange}{Given the question, the case is an example about the question. Please study this example and answer the question based on the description.}\\
    \quad return LLM\_answer\\
\\
\textcolor{blue}{sub\_question1}="Who was the member of the 'Mother Love Bone' band who passed away before the release of the album 'Apple'?"\\
\textcolor[rgb]{0.1333, 0.5451, 0.1333}{sub\_case1}=" Please focus on the correlation between the death of band members and the upcoming release of the album."\\
\textcolor[rgb]{1, 0, 0}{sub\_description1}="\textcolor[rgb]{0.11, 0.56, 1}{Frontman Andrew Wood’s personality and compositions helped to catapult the group to the top of the burgeoning late 1980s/early 1990s Seattle music scene.  Wood died only days before the scheduled release of the band’s debut album,“Apple”, thus ending the group’s hopes of success. The album was finally released a few months later.}"\\
\textcolor[rgb]{1, 0.078, 0.57}{answer1}=LLM(sub\_case1,sub\_question1,sub\_description1)\\
\\
\textcolor{blue}{sub\_question2}="What band has Andrew Wood joined (except for 'Mother Love Bone')?"\\
\textcolor[rgb]{0.1333, 0.5451, 0.1333}{sub\_case2}="Please pay attention to the description of the person's band experience."\\
\textcolor[rgb]{1, 0, 0}{sub\_description2}="\textcolor[rgb]{0.11, 0.56, 1}{Return to Olympus is the only album by the alterna tive rock band Malfunkshun. It was released after the band had broken up and after lead singer Andrew Wood (later of Mother Love Bone) had died of a drug overdose in 1990.}"\\
\textcolor[rgb]{1, 0.078, 0.57}{answer2}=LLM(sub\_case2,sub\_question2,sub\_description2)\\
\\
\textcolor{blue}{sub\_question3}="What band was Andrew Wood in before joining Mother Love Bone?"\\
\textcolor[rgb]{0.1333, 0.5451, 0.1333}{sub\_case3}="Please pay attention to the order in which the person joined the band."\\
\textcolor[rgb]{0.72, 0.52, 0.04}{\#Answering sub\_question3 requires relying on the answers of sub\_question1 and sub\_question2, namely answer1 and answer2}\\
\textcolor[rgb]{1, 0, 0}{sub\_description3}=\textcolor[rgb]{1, 0.078, 0.57}{answer1}+\textcolor[rgb]{1, 0.078, 0.57}{answer2}\\
\textcolor[rgb]{1, 0.078, 0.57}{answer3}=LLM(sub\_case3,sub\_question3,sub\_description3)\\
\\
\textcolor[rgb]{0.72, 0.52, 0.04}{\#The final answer}\\
\textcolor[rgb]{0, 0.39, 0}{final\_answer}=\textcolor[rgb]{1, 0.078, 0.57}{answer3}\\
print(final\_answer)\\

      \hline
  \end{tabular}
  \caption{\label{HotpotQA}
    The updated pseudo-program knowledge graph on HotpotQA. (
 Question: What was the former band of the member of Mother Love Bone who died just before the release of “Apple”?
Paragraph A, Return to Olympus:
 Return to Olympus is the only album by the alternative rock band Malfunkshun. It was released after the band had broken up and after lead singer Andrew Wood (later of Mother Love Bone) had died of a drug overdose in 1990. Stone Gossard, of Pearl Jam, had compiled the songs and released the album on his label, Loosegroove Records.
 Paragraph B, Mother Love Bone:
 Mother Love Bone was an American rock band that formed in Seattle, Washington in 1987. The band was active from 1987 to 1990. Frontman Andrew Wood’s personality and compositions helped to catapult the group to the top of the burgeoning late 1980s/early 1990s Seattle music scene. Wood died only days before the scheduled release of the band’s debut album,“Apple”, thus ending the group’s hopes of success. The album was finally released a few months later.)
  }
\end{table*}

\begin{table*}[t]
  \centering
  \small
  \begin{tabular}{p{\linewidth}}
    \hline
Let's break down the program and follow the steps logically to determine the final answer.\\
\\
Step-by-Step Breakdown:\\
\\
1. First Call to LLM (sub\_question1):\\
   - Question: Who was the member of the "Mother Love Bone" band who passed away before the release of the album "Apple"?\\
   - Case: The case asks us to focus on the correlation between the death of a band member and the upcoming release of the album.\\
   - Description: The description mentions that the frontman, Andrew Wood, passed away just days before the release of "Apple," which was the band's debut album.\\
   
   Answer: The band member who passed away before the release of the album "Apple" is Andrew Wood.\\
\\
   answer1 = "Andrew Wood"\\
\\
2. Second Call to LLM (sub\_question2):
   - Question: What band has Andrew Wood joined (except for "Mother Love Bone")?\\
   - Case: The case tells us to pay attention to the description of the person's band experience.\\
   - Description: The description talks about Andrew Wood being the lead singer of a band called Malfunkshun, which released the album "Return to Olympus" after the band broke up. This band existed before Andrew Wood joined "Mother Love Bone."\\
   \\
   Answer: Before joining "Mother Love Bone," Andrew Wood was part of the band Malfunkshun.\\
\\
   answer2 = "Malfunkshun"\\
\\
3. Third Call to LLM (sub\_question3):\\
   - Question: What band was Andrew Wood in before joining Mother Love Bone?\\
   - Case: This call asks to pay attention to the order in which Andrew Wood joined bands.\\
   - Description: The description includes the answers from previous questions. From answer1, we know Andrew Wood was in "Mother Love Bone." From answer2, we know he was in "Malfunkshun" before joining "Mother Love Bone."\\
   
   Answer: Before joining "Mother Love Bone," Andrew Wood was in "Malfunkshun."\\
\\
   answer3 = "Malfunkshun"\\
\\
4. Final Answer:\\
   The final answer is "Malfunkshun."\\
\\
Final Output: Malfunkshun

   \\ \hline
  \end{tabular}
  \caption{\label{HotpotQAresult}
    The output of executing the pseudo-program knowledge graph in Table \ref{HotpotQA}.
  }
\end{table*}

\begin{table*}[t]
  \centering
  \small
  \begin{tabular}{p{\linewidth}}
    \hline
\textcolor[rgb]{0.72, 0.52, 0.04}{\#Given a program text, your role is the LLM function,which has three parameters: case, question, and description. The description is textual, the question requires you to answer based on the description, and the case involves processed scenarios or supplementary information or related considerations to assist you in answering the question. You need to strictly follow the program logic to execute. During execution, the LLM function will be called multiple times, which means you will answer corresponding questions based on different cases and descriptions. Please output the final result of this program text in natural language.}\\
\\
def LLM(case,question,description):\\
    \quad LLM\_answer=\textcolor{orange}{Given the question, the case is an example about the question. Please study this example and answer the question based on the description.}\\
    \quad return LLM\_answer\\
\\
\textcolor{blue}{sub\_question1}="What are the general characteristics of the area represented by each option in terms of housing distribution and isolation?"\\
\textcolor[rgb]{0.1333, 0.5451, 0.1333}{sub\_case1}=" \\
Neighborhood: The houses in the neighborhood are relatively compact, and there may be a mix of different types of houses. The boundaries are usually not particularly closed, and many houses in the neighborhood may not have obvious isolation facilities between them. They are closely connected to the surrounding streets and other buildings.
\\
Subdivision: Zoning is usually a planned area, and some zones may have a unified architectural style and management regulations. Some high-end partitions may have access control facilities to ensure residents' privacy and safety, isolating themselves from the outside world.
\\
City: A city is a large area that includes various functional areas and a wide variety of housing types. Most houses in the city are open plan, closely connected to various commercial facilities, public transportation, etc., and few houses can be completely isolated from other places.
\\
Suburbs: Houses in the suburbs are relatively scattered, but generally do not have strong isolation. The suburbs are mainly for residential purposes, although there may be some independent residences, they are not primarily characterized by isolation.
\\
Street: The street is mainly a transportation channel, and the houses on both sides of the street are mainly distributed along the road, usually without overall isolation measures."\\
\textcolor[rgb]{1, 0, 0}{sub\_description1}="\textcolor[rgb]{0.11, 0.56, 1}{Answer Choices: (A) neighborhood (B) subdivision (C) city (D) suburbs (E) street}"\\
\textcolor[rgb]{1, 0.078, 0.57}{answer1}=LLM(sub\_case1,sub\_question1,sub\_description1)\\
\\
\textcolor{blue}{sub\_question2}="Which area is most likely to provide houses with access control facilities and isolation from other areas?"\\
\textcolor[rgb]{0.1333, 0.5451, 0.1333}{sub\_case2}="Pay attention to comparing the characteristics of each region"\\
\textcolor[rgb]{0.72, 0.52, 0.04}{\#Answering sub\_question2 requires relying on the answers of sub\_question1, namely answer1}\\
\textcolor[rgb]{1, 0, 0}{sub\_description2}=\textcolor[rgb]{1, 0.078, 0.57}{answer1}\\
\textcolor[rgb]{1, 0.078, 0.57}{answer2}=LLM(sub\_case2,sub\_question2,sub\_description2)\\
\\
\textcolor[rgb]{0.72, 0.52, 0.04}{\#The final answer}\\
\textcolor[rgb]{0, 0.39, 0}{final\_answer}=\textcolor[rgb]{1, 0.078, 0.57}{answer2}\\
print(final\_answer)\\

      \hline
  \end{tabular}
  \caption{\label{CSQA}
    The updated pseudo-program knowledge graph on CSQA, where \textit{sub-case} is derived from the LLM-decomposed \textit{Knowledge case} of the decision tree built by experts. (Question: He wanted a house that was gated off from other places, where should he start looking?  Answer Choices: (A) neighborhood (B) subdivision (C) city (D) suburbs (E) street)
  }
\end{table*}

\begin{table*}[t]
  \centering
  \small
  \begin{tabular}{p{\linewidth}}
    \hline
Let's walk through the steps of the program and how it works logically to produce the final result.\\
\\
Step-by-Step Breakdown:\\
\\
1. First Call to LLM (sub\_question1):\\
   - Question: What are the general characteristics of the area represented by each option in terms of housing distribution and isolation?\\
   - Case: The case provides descriptions of different areas: neighborhood, subdivision, city, suburbs, and street. Each description highlights the characteristics of housing distribution and isolation.\\
   - Description: The answer choices include (A) neighborhood, (B) subdivision, (C) city, (D) suburbs, and (E) street.\\
\\
   Analysis: The focus of the question is on the characteristics of each option in terms of housing distribution and isolation.\\
   \\
   - Neighborhood: Houses are relatively compact with no clear isolation between them.\\
   - Subdivision: Has unified architectural styles and may have access control facilities and isolation, especially in high-end areas.\\
   - City: Houses are open plan and closely connected to other buildings and commercial areas; few houses have isolation.\\
   - Suburbs: Houses are scattered but generally lack strong isolation.\\
   - Street: Houses are distributed along the road and generally have no isolation.\\
\\
   Answer: Based on the description, the "subdivision" is most likely to provide houses with access control facilities and isolation from other areas, especially in high-end zones.\\
\\
   answer1 = "B" (subdivision)\\
\\
2. Second Call to LLM (sub\_question2):\\
   - Question: Which area is most likely to provide houses with access control facilities and isolation from other areas?\\
   - Case: The case asks to compare the characteristics of each area to determine which is most likely to have access control and isolation.\\
   - Description: The description relies on the answer from sub\_question1, which identified the subdivision as the area most likely to have access control facilities and isolation.\\
\\
   Answer: The subdivision is the most likely area to provide houses with access control and isolation based on the characteristics described earlier.\\
\\
   answer2 = "B" (subdivision)\\
\\
3. Final Answer:\\
   The final answer is "B", corresponding to **subdivision**.\\
\\
Final Output: B

 \\   \hline
  \end{tabular}
  \caption{\label{CSQAresult}
    The output of executing the pseudo-program knowledge graph in Table \ref{CSQA}.
  }
\end{table*}

\begin{table*}[t]
  \centering
  \small
  \begin{tabular}{p{\linewidth}}
    \hline
\textcolor[rgb]{0.72, 0.52, 0.04}{\#Given a program text, your role is the LLM function,which has three parameters: case, question, and description. The description is textual, the question requires you to answer based on the description, and the case involves processed scenarios or supplementary information or related considerations to assist you in answering the question. You need to strictly follow the program logic to execute. During execution, the LLM function will be called multiple times, which means you will answer corresponding questions based on different cases and descriptions. Please output the final result of this program text in natural language.}\\
\\
def LLM(case,question,description):\\
    \quad LLM\_answer=\textcolor{orange}{Given the question, the case is an example about the question. Please study this example and answer the question based on the description.}\\
    \quad return LLM\_answer\\
\\
\textcolor{blue}{sub\_question1}="Is this behavior directly related to understanding and answering the teacher's questions?"\\
\textcolor[rgb]{0.1333, 0.5451, 0.1333}{sub\_case1}=" \\
Asking for a gold star is a behavior where one may expect to receive a reward after answering a question, and it is not something that needs to be done before understanding and answering the question."\\
\textcolor[rgb]{1, 0, 0}{sub\_description1}="\textcolor[rgb]{0.11, 0.56, 1}{Behavior: ask for a gold star}"\\
\textcolor[rgb]{1, 0.078, 0.57}{answer1}=LLM(sub\_case1,sub\_question1,sub\_description1)\\
\\
\textcolor{blue}{sub\_question2}="Is this behavior helpful for understanding and answering the teacher's questions?"\\
\textcolor[rgb]{0.1333, 0.5451, 0.1333}{sub\_case2}="Skipping class means not being in the classroom and not being able to hear the teacher's questions"\\
\textcolor[rgb]{1, 0, 0}{sub\_description2}="\textcolor[rgb]{0.11, 0.56, 1}{Behavior: skip her class}"\\
\textcolor[rgb]{1, 0.078, 0.57}{answer2}=LLM(sub\_case2,sub\_question2,sub\_description2)\\
\\
\textcolor{blue}{sub\_question3}="Does this behavior help to understand and answer the teacher's questions?"\\
\textcolor[rgb]{0.1333, 0.5451, 0.1333}{sub\_case3}="Only when a person knows the information related to the problem, can they understand the content of the teacher's question well and accurately answer the question based on the information they have mastered"\\
\textcolor[rgb]{1, 0, 0}{sub\_description3}="\textcolor[rgb]{0.11, 0.56, 1}{Behavior: know the information}"\\
\textcolor[rgb]{1, 0.078, 0.57}{answer3}=LLM(sub\_case3,sub\_question3,sub\_description3)\\

\textcolor{blue}{sub\_question4}="Which behavior is most likely to occur?"\\
\textcolor[rgb]{0.1333, 0.5451, 0.1333}{sub\_case4}="Please choose the behavior that best fits your understanding and answers the teacher's question in this scenario."\\
\textcolor[rgb]{0.72, 0.52, 0.04}{\#Answering sub\_question4 requires relying on the answers of sub\_question1,sub\_question2 and sub\_question3, namely answer1, answer2 and answer3}\\
\textcolor[rgb]{1, 0, 0}{sub\_description4}=\textcolor[rgb]{1, 0.078, 0.57}{answer1}+\textcolor[rgb]{1, 0.078, 0.57}{answer2}+\textcolor[rgb]{1, 0.078, 0.57}{answer3}\\
\textcolor[rgb]{1, 0.078, 0.57}{answer4}=LLM(sub\_case3,sub\_question3,sub\_description3)\\
\\
\textcolor[rgb]{0.72, 0.52, 0.04}{\#The final answer}\\
\textcolor[rgb]{0, 0.39, 0}{final\_answer}=\textcolor[rgb]{1, 0.078, 0.57}{answer4}\\
print(final\_answer) \\

    \hline
  \end{tabular}
  \caption{\label{SIQA}
    The updated pseudo-program knowledge graph on SIQA. (Question: The teacher asked the class a question and they seemed puzzled. Aubrey understood the question well and answered. What does Aubrey need to do before this?  Answer Choices: (A) ask for a gold star (B) skip her class (C) know the information)
  }
\end{table*}

\begin{table*}[t]
  \centering
  \small
  \begin{tabular}{p{\linewidth}}
    \hline
Let's break down the logic and the steps for this program:\\
\\
Step-by-Step Breakdown:\\
\\
1. First Call to LLM (sub\_question1):\\
   - Question: Is this behavior directly related to understanding and answering the teacher's questions?\\
   - Case: Asking for a gold star is a behavior that does not directly contribute to understanding the question. It's more about receiving a reward after the question is answered.\\
   - Description: The behavior is to "ask for a gold star."\\
   \\
   Answer: Since asking for a gold star doesn't directly relate to understanding or answering the question, the answer is "No."\\
\\

2. Second Call to LLM (sub\_question2):\\
   - Question: Is this behavior helpful for understanding and answering the teacher's questions?\\
   - Case: Skipping class means not being present to hear the teacher's questions, which would make it impossible to understand or answer the question.\\
   - Description: The behavior is "skip her class."\\
   \\
   Answer: Since skipping class means missing the teacher's questions, the answer is "No."\\

\\
3. Third Call to LLM (sub\_question3):\\
   - Question: Does this behavior help to understand and answer the teacher's questions?\\
   - Case: Knowing the information related to the problem is essential to understanding the teacher's question and answering it accurately.\\
   - Description: The behavior is "know the information."\\
   \\
   Answer: Knowing the information is crucial for understanding and answering the question, so the answer is "Yes."\\
\\
4. Fourth Call to LLM (sub\_question4):\\
   - Question: Which behavior is most likely to occur?\\
   - Case: This question asks to select the behavior that best fits the context of understanding and answering the teacher's question.\\
   - Description: The description includes the answers from the previous questions: "No," "No," and "Yes."\\
   \\
   Answer: The behavior most likely to occur is the one that helps understand and answer the teacher's question, which is knowing the information ("Yes").\\
\\
 Final Answer: know the information\\

     \hline
  \end{tabular}
  \caption{\label{SIQAresult}
    The output of executing the pseudo-program knowledge graph in Table \ref{SIQA}.
  }
\end{table*}

\begin{table*}[t]
  \centering
  \small
  \begin{tabular}{p{\linewidth}}
    \hline
\textcolor[rgb]{0.72, 0.52, 0.04}{\#Given a program text, your role is the LLM function,which has three parameters: case, question, and description. The description is textual, the question requires you to answer based on the description, and the case involves processed scenarios or supplementary information or related considerations to assist you in answering the question. You need to strictly follow the program logic to execute. During execution, the LLM function will be called multiple times, which means you will answer corresponding questions based on different cases and descriptions. Please output the final result of this program text in natural language.}\\
\\
def LLM(case,question,description):\\
    \quad LLM\_answer=\textcolor{orange}{Given the question, the case is an example about the question. Please study this example and answer the question based on the description.}\\
    \quad return LLM\_answer\\
\\
\textcolor{blue}{sub\_question1}="What is the last letter of this word?"\\
\textcolor[rgb]{0.1333, 0.5451, 0.1333}{sub\_case1}=" \\
Description:  Pretty\\
Answer: The last letter of 'Pretty' is' y '."\\
\textcolor[rgb]{1, 0, 0}{sub\_description1}="\textcolor[rgb]{0.11, 0.56, 1}{Gavin}"\\
\textcolor[rgb]{1, 0.078, 0.57}{answer1}=LLM(sub\_case1,sub\_question1,sub\_description1)\\
\\
\textcolor{blue}{sub\_question2}="What is the last letter of this word?"\\
\textcolor[rgb]{0.1333, 0.5451, 0.1333}{sub\_case2}="\\
Description: Jada\\
Answer: The last letter of 'Jada' is' a '."\\
\textcolor[rgb]{1, 0, 0}{sub\_description2}="\textcolor[rgb]{0.11, 0.56, 1}{Neha}"\\
\textcolor[rgb]{1, 0.078, 0.57}{answer2}=LLM(sub\_case2,sub\_question2,sub\_description2)\\
\\
\textcolor{blue}{sub\_question3}="What is the last letter of this word?"\\
\textcolor[rgb]{0.1333, 0.5451, 0.1333}{sub\_case3}="\\
Description: Sarita\\
Answer: The last letter of 'Sarita' is' a '."\\
\textcolor[rgb]{1, 0, 0}{sub\_description3}="\textcolor[rgb]{0.11, 0.56, 1}{Asha}"\\
\textcolor[rgb]{1, 0.078, 0.57}{answer3}=LLM(sub\_case3,sub\_question3,sub\_description3)\\
\\
\textcolor{blue}{sub\_question4}="What is the last letter of this word?"\\
\textcolor[rgb]{0.1333, 0.5451, 0.1333}{sub\_case4}="\\
Description: Allen\\
Answer: The last letter of 'Allen' is' n '."\\
\textcolor[rgb]{1, 0, 0}{sub\_description4}="\textcolor[rgb]{0.11, 0.56, 1}{Baltazar}"\\
\textcolor[rgb]{1, 0.078, 0.57}{answer4}=LLM(sub\_case4,sub\_question4,sub\_description4)\\
\\
\textcolor{blue}{sub\_question5}="What is the result of concatenating the last letter obtained from each word in order?"\\
\textcolor[rgb]{0.1333, 0.5451, 0.1333}{sub\_case5}="\\
Description: The last letter of 'Pretty' is' y ', the last letter of 'Jada' is' a ', the last letter of 'Sarita' is' a ' and the last letter of 'Allen' is' n '.\\
Answer: The last letter of the four words is' y ',' a ',' a ',' n ', and when they are concatenated in order, the result is' yaan'."\\
\textcolor[rgb]{0.72, 0.52, 0.04}{\#Answering sub\_question5 requires relying on the answers of sub\_question1,sub\_question2,sub\_question3 and sub\_question4, namely answer1, answer2, answer3 and answer4}\\
\textcolor[rgb]{1, 0, 0}{sub\_description5}=\textcolor[rgb]{1, 0.078, 0.57}{answer1}+\textcolor[rgb]{1, 0.078, 0.57}{answer2}+\textcolor[rgb]{1, 0.078, 0.57}{answer3}+\textcolor[rgb]{1, 0.078, 0.57}{answer4}\\
\textcolor[rgb]{1, 0.078, 0.57}{answer5}=LLM(sub\_case5,sub\_question5,sub\_description5)\\
\\
\textcolor[rgb]{0.72, 0.52, 0.04}{\#The final answer}\\
\textcolor[rgb]{0, 0.39, 0}{final\_answer}=\textcolor[rgb]{1, 0.078, 0.57}{answer5}\\
print(final\_answer)\\

    \hline
  \end{tabular}
  \caption{\label{Lastletter}
    The updated pseudo-program knowledge graph on Last Letter Concatenation. (Question: Take the last letters of each words in \"Gavin Neha Asha Baltazar\" and concatenate them.)
  }
\end{table*}

\begin{table*}[t]
  \centering
  \small
  \begin{tabular}{p{\linewidth}}
    \hline
Let's break down the program step by step to determine the final output:\\
\\
Step-by-Step Breakdown:\\
\\
1. First Call to LLM (sub\_question1):\\
   - Question: What is the last letter of this word?\\
   - Case: The description indicates that the last letter of "Pretty" is "y".\\
   - Description: The word is "Gavin".\\
   \\
   Answer: The last letter of "Gavin" is "n".\\
\\
   answer1 = "n"\\
\\
2. Second Call to LLM (sub\_question2):\\
   - Question: What is the last letter of this word?\\
   - Case: The description indicates that the last letter of "Jada" is "a".\\
   - Description: The word is "Neha".\\
   \\
   Answer: The last letter of "Neha" is "a".\\
\\
   answer2 = "a"\\
\\
3. Third Call to LLM (sub\_question3):\\
   - Question: What is the last letter of this word?\\
   - Case: The description indicates that the last letter of "Sarita" is "a".\\
   - Description: The word is "Asha".\\
   \\
   Answer: The last letter of "Asha" is "a".\\
\\
   answer3 = "a"\\
\\
4. Fourth Call to LLM (sub\_question4):\\
   - Question: What is the last letter of this word?\\
   - Case: The description indicates that the last letter of "Allen" is "n".\\
   - Description: The word is "Baltazar".\\
   \\
   Answer: The last letter of "Baltazar" is "r".\\
\\
   answer4 = "r"\\
\\
5. Fifth Call to LLM (sub\_question5):\\
   - Question: What is the result of concatenating the last letter obtained from each word in order?\\
   - Case: The case gives the last letters of the words: "n" (from Gavin), "a" (from Neha), "a" (from Asha), and "r" (from Baltazar).\\
   - Description: The concatenation of the last letters "n", "a", "a", and "r" results in "naar".\\
\\
   Answer: The concatenated result of the last letters is "naar".\\
\\
   answer5 = "naar"\\
\\
Final Answer: naar\\

     \hline
  \end{tabular}
  \caption{\label{Lastletterresult}
    The output of executing the pseudo-program knowledge graph in Table \ref{Lastletter}.
  }
\end{table*}

\begin{table*}[t]
  \centering
  \small
  \begin{tabular}{p{\linewidth}}
    \hline
\textcolor[rgb]{0.72, 0.52, 0.04}{\#Given a program text, your role is the LLM function,which has three parameters: case, question, and description. The description is textual, the question requires you to answer based on the description, and the case involves processed scenarios or supplementary information or related considerations to assist you in answering the question. You need to strictly follow the program logic to execute. During execution, the LLM function will be called multiple times, which means you will answer corresponding questions based on different cases and descriptions. Please output the final result of this program text in natural language.}\\
\\
def LLM(case,question,description):\\
    \quad LLM\_answer=\textcolor{orange}{Given the question, the case is an example about the question. Please study this example and answer the question based on the description.}\\
    \quad return LLM\_answer\\
\\
\textcolor{blue}{sub\_question1}="What is the initial state of the coin?"\\
\textcolor[rgb]{0.1333, 0.5451, 0.1333}{sub\_case1}="Pay attention to the initial state of the coin at this moment."\\
\textcolor[rgb]{1, 0, 0}{sub\_description1}="\textcolor[rgb]{0.11, 0.56, 1}{A coin is heads up}"\\
\textcolor[rgb]{1, 0.078, 0.57}{answer1}=LLM(sub\_case1,sub\_question1,sub\_description1)\\
\\
\textcolor{blue}{sub\_question2}="What is the current state of the coin?"\\
\textcolor[rgb]{0.1333, 0.5451, 0.1333}{sub\_case2}="\\
Pay attention to whether this operation changes the original state of the coin and outputs the state after this operation."\\
\textcolor[rgb]{0.72, 0.52, 0.04}{\#Answering sub\_question2 requires relying on the answers of sub\_question1, namely answer1}\\
\textcolor[rgb]{1, 0, 0}{sub\_description2}=\textcolor[rgb]{1, 0.078, 0.57}{answer1}+"\textcolor[rgb]{0.11, 0.56, 1}{Sager does not flip the coin}"\\
\textcolor[rgb]{1, 0.078, 0.57}{answer2}=LLM(sub\_case2,sub\_question2,sub\_description2)\\
\\
\textcolor{blue}{sub\_question3}="What is the current state of the coin?"\\
\textcolor[rgb]{0.1333, 0.5451, 0.1333}{sub\_case3}="\\
Pay attention to whether this operation changes the original state of the coin and outputs the state after this operation"\\
\textcolor[rgb]{0.72, 0.52, 0.04}{\#Answering sub\_question3 requires relying on the answers of sub\_question2, namely answer2}\\
\textcolor[rgb]{1, 0, 0}{sub\_description3}=\textcolor[rgb]{1, 0.078, 0.57}{answer2}+"\textcolor[rgb]{0.11, 0.56, 1}{Zyheir flips the coin}"\\
\textcolor[rgb]{1, 0.078, 0.57}{answer3}=LLM(sub\_case3,sub\_question3,sub\_description3)\\
\\
\textcolor{blue}{sub\_question4}="Is the coin still heads up?"\\
\textcolor[rgb]{0.1333, 0.5451, 0.1333}{sub\_case4}="Pay attention to the coin's transition at this moment"\\
\textcolor[rgb]{0.72, 0.52, 0.04}{\#Answering sub\_question4 requires relying on the answers of sub\_question3, namely answer3}\\
\textcolor[rgb]{1, 0, 0}{sub\_description4}=\textcolor[rgb]{1, 0.078, 0.57}{answer3}\\
\textcolor[rgb]{1, 0.078, 0.57}{answer4}=LLM(sub\_case4,sub\_question4,sub\_description4)\\
\\
\textcolor[rgb]{0.72, 0.52, 0.04}{\#The final answer}\\
\textcolor[rgb]{0, 0.39, 0}{final\_answer}=\textcolor[rgb]{1, 0.078, 0.57}{answer4}\\
print(final\_answer)\\

    \hline
  \end{tabular}
  \caption{\label{CoinFlip}
    The updated pseudo-program knowledge graph on Coin Flip, where \textit{sub-case} is derived from the LLM-decomposed \textit{Knowledge case} of the decision tree built by experts. (Question: A coin is heads up. sager does not flip the coin. zyheir flips the coin.  Is the coin still heads up?)
  }
\end{table*}

\begin{table*}[t]
  \centering
  \small
  \begin{tabular}{p{\linewidth}}
    \hline
Let's analyze the program step-by-step and follow the execution logic to determine the final output:\\
\\
Step-by-Step Breakdown:\\
\\
1. First Call to LLM (sub\_question1):\\
   - Question: What is the initial state of the coin?\\
   - Case: Pay attention to the initial state of the coin at this moment.\\
   - Description: The coin is heads up.\\
   \\
   Answer: Based on the description, the initial state of the coin is "heads up."\\
\\
   answer1 = "heads up"\\
\\
2. Second Call to LLM (sub\_question2):\\
   - Question: What is the current state of the coin?\\
   - Case: Pay attention to whether this operation changes the original state of the coin and output the state after the operation.\\
   - Description: The description includes the answer from 'sub\_question1' ("heads up") and adds the information that Sager does not flip the coin.\\
   \\
   Answer: Since Sager does not flip the coin, the state remains the same as "heads up."\\
\\
   answer2 = "heads up"\\
\\
3. Third Call to LLM (sub\_question3):\\
   - Question: What is the current state of the coin?\\
   - Case: Pay attention to whether this operation changes the original state of the coin and output the state after the operation.\\
   - Description: The description includes the answer from 'sub\_question2' ("heads up") and adds the information that Zyheir flips the coin.\\
   \\
   Answer: Since Zyheir flips the coin, the state changes from "heads up" to "tails up."\\
\\
   answer3 = "tails up"\\
\\
4. Fourth Call to LLM (sub\_question4):\\
   - Question: Is the coin still heads up?\\
   - Case: Pay attention to the coin's transition at this moment.\\
   - Description: The description includes the answer from 'sub\_question3'  ("tails up"), so we check if the coin is still heads up.\\
   \\
   Answer: Since the coin is "tails up" (after Zyheir flipped it), it is no longer heads up.\\
\\
   answer4 = "No"\\
\\
5. Final Answer:\\
   The final answer is "No" because after the flip, the coin is not heads up.\\

     \hline
  \end{tabular}
  \caption{\label{CoinFlipresult}
    The output of executing the pseudo-program knowledge graph in Table \ref{CoinFlip}.
  }
\end{table*}

\begin{table*}[t]
  \centering
  \small
  \begin{tabular}{p{\linewidth}}
    \hline
\textcolor[rgb]{0.72, 0.52, 0.04}{\#Given a program text, your role is the LLM function,which has three parameters: case, question, and description. The description is textual, the question requires you to answer based on the description, and the case involves processed scenarios or supplementary information or related considerations to assist you in answering the question. You need to strictly follow the program logic to execute. During execution, the LLM function will be called multiple times, which means you will answer corresponding questions based on different cases and descriptions. Please output the final result of this program text in natural language.}\\
\\
def LLM(case,question,description):\\
    \quad LLM\_answer=\textcolor{orange}{Given the question, the case is an example about the question. Please study this example and answer the question based on the description.}\\
    \quad return LLM\_answer\\
\\
\textcolor{blue}{sub\_question1}="What is the amount of cash stolen in this theft?"\\
\textcolor[rgb]{0.1333, 0.5451, 0.1333}{sub\_case1}="Pay attention to accurately finding sentences that describe the theft and the corresponding amount of stolen cash, to avoid confusion with related descriptions of other theft events."\\
\textcolor[rgb]{1, 0, 0}{sub\_description1}="\textcolor[rgb]{0.11, 0.56, 1}{At about 11:00 on January 2, 2016, the defendant Liu XX, Song XX, and Hu Mou took the Ouyang special bus and passed a certain section of a certain town in Jianyang District, Nanping City. Hu Mou and Liu XX watched, and Song XX used a blade to cut the right pocket of the victim Liu Mou's coat and stole the small yellow envelope containing 1,500 yuan in cash.}"\\
\textcolor[rgb]{1, 0.078, 0.57}{answer1}=LLM(sub\_case1,sub\_question1,sub\_description1)\\
\\
\textcolor{blue}{sub\_question2}="What is the amount of cash stolen in this theft?"\\
\textcolor[rgb]{0.1333, 0.5451, 0.1333}{sub\_case2}="Pay attention to accurately finding sentences that describe the theft and the corresponding amount of stolen cash, to avoid confusion with related descriptions of other theft events."\\
\textcolor[rgb]{1, 0, 0}{sub\_description2}="\textcolor[rgb]{0.11, 0.56, 1}{At about 11:00 on January 20, 2016, the defendants Liu XX, Song XX, and Hu Mou took the Ouyang special bus to a certain section of the road. They saw the victim Zhang Mou 1 sleeping, so Hu Mou and Liu XX watched, and Song XX cut open Zhang Mou 1's coat pocket with a blade, and stole 7,000 yuan in cash tied with cow tendons in the pocket.}"\\
\textcolor[rgb]{1, 0.078, 0.57}{answer2}=LLM(sub\_case2,sub\_question2,sub\_description2)\\
\\
\textcolor{blue}{sub\_question3}="What is the total amount involved in all theft crimes?"\\
\textcolor[rgb]{0.1333, 0.5451, 0.1333}{sub\_case3}="Pay attention to performing correct addition operations on all previous results to ensure that there are no errors in the calculation process, and do not omit or double calculate the amount."\\
\textcolor[rgb]{0.72, 0.52, 0.04}{\#Answering sub\_question3 requires relying on the answers of sub\_question1 and sub\_question2, namely answer1 and answer2}\\
\textcolor[rgb]{1, 0, 0}{sub\_description3}=\textcolor[rgb]{1, 0.078, 0.57}{answer1}+\textcolor[rgb]{1, 0.078, 0.57}{answer2}\\
\textcolor[rgb]{1, 0.078, 0.57}{answer3}=LLM(sub\_case3,sub\_question3,sub\_description3)\\
\\
\textcolor[rgb]{0.72, 0.52, 0.04}{\#The final answer}\\
\textcolor[rgb]{0, 0.39, 0}{final\_answer}=\textcolor[rgb]{1, 0.078, 0.57}{answer3}\\
print(final\_answer)\\

    \hline
  \end{tabular}
  \caption{\label{LawBench}
    The updated pseudo-program knowledge graph on LawBench. (Question: What is the total amount of crime involved in the document?
Document: After investigation, it was found that during January 2016, the defendant Liu XX, Song XX, and Hu (both of whom have been sentenced) agreed to commit theft on the Ouyang special bus, and agreed that whoever did it would get half of the stolen property, and the other two would share the other half. The specific criminal facts are as follows: 1. At about 11:00 on January 2, 2016, the defendant Liu XX, Song XX, and Hu Mou took the Ouyang special bus and passed a certain section of a certain town in Jianyang District, Nanping City. Hu Mou and Liu XX watched, and Song XX used a blade to cut the right pocket of the victim Liu Mou's coat and stole the small yellow envelope containing 1,500 yuan in cash. After getting off the bus, Song XX got 750 yuan of the stolen money, and Hu Mou and Liu XX each got 375 yuan of the stolen money. 2. At about 11:00 on January 20, 2016, the defendants Liu XX, Song XX, and Hu Mou took the Ouyang special bus to a certain section of the road. They saw the victim Zhang Mou 1 sleeping, so Hu Mou and Liu XX watched, and Song XX cut open Zhang Mou 1's coat pocket with a blade, and stole 7,000 yuan in cash tied with cow tendons in the pocket. Afterwards, Song XX received 3,500 yuan, and Hu Mou and Liu XX each received 1,750 yuan.)
  }
\end{table*}

\begin{table*}[t]
  \centering
  \small
  \begin{tabular}{p{\linewidth}}
    \hline
To solve this, let's break down the program logic and how it works step by step:\\
\\
Step-by-step breakdown:\\
\\
1. First Call to LLM (sub\_question1):\\
   - Question: What is the amount of cash stolen in this theft?\\
   - Case: Focus on accurately finding the amount of stolen cash in the description, ensuring no confusion with other theft events.\\
   - Description: The theft took place on January 2, 2016, where Song XX cut open the victim's coat pocket and stole a small yellow envelope containing 1,500 yuan.\\
   Answer: Based on the description, the amount of cash stolen is 1,500 yuan.\\
\\
   answer1 = 1500\\
\\
2. Second Call to LLM (sub\_question2):\\
   - Question: What is the amount of cash stolen in this theft?\\
   - Case: Again, accurately find the amount of stolen cash from the description, ensuring no confusion with other theft events.\\
   - Description: On January 20, 2016, Song XX stole 7,000 yuan from the victim Zhang Mou 1's coat pocket.\\
   Answer: Based on the description, the amount of cash stolen is 7,000 yuan.\\
\\
   answer2 = 7000\\
\\
3. Third Call to LLM (sub\_question3):\\
   - Question: What is the total amount involved in all theft crimes?\\
   - Case: Perform the addition operation correctly to avoid errors and omissions, using the amounts obtained from the previous answers.\\
   - Description: The description contains both 'answer1' (1,500 yuan) and 'answer2' (7,000 yuan).\\
   \\
   Answer: To calculate the total amount, we add 'answer1' and 'answer2':  \\
   1,500 + 7,000 = 8,500 yuan.\\
   answer3 = 8500\\
\\
4. Final Answer: The total amount involved in all theft crimes is 8,500 yuan.\\

     \hline
  \end{tabular}
  \caption{\label{LawBenchresult}
    The output of executing the pseudo-program knowledge graph in Table \ref{LawBench}.
  }
\end{table*}

\begin{table*}[t]
  \centering
  \small
  \begin{tabular}{p{\linewidth}}
    \hline
\textcolor[rgb]{0.72, 0.52, 0.04}{\#Given a program text, your role is the LLM function,which has three parameters: case, question, and description. The description is textual, the question requires you to answer based on the description, and the case involves processed scenarios or supplementary information or related considerations to assist you in answering the question. You need to strictly follow the program logic to execute. During execution, the LLM function will be called multiple times, which means you will answer corresponding questions based on different cases and descriptions. Please output the final result of this program text in natural language.}\\
\\
def LLM(case,question,description):\\
    \quad LLM\_answer=\textcolor{orange}{Given the question, the case is an example about the question. Please study this example and answer the question based on the description.}\\
    \quad return LLM\_answer\\
\\
\textcolor{blue}{sub\_question1}="Does the contract mention any content regarding usage permissions?"\\
\textcolor[rgb]{0.1333, 0.5451, 0.1333}{sub\_case1}="Pay attention to carefully reading the text and find the guiding statements related to content usage permissions."\\
\textcolor[rgb]{1, 0, 0}{sub\_description1}="\textcolor[rgb]{0.11, 0.56, 1}{See the Permission to use your content section for more about your rights in your content, and how your content is used in our services.}"\\
\textcolor[rgb]{1, 0.078, 0.57}{answer1}=LLM(sub\_case1,sub\_question1,sub\_description1)\\
\\
\textcolor{blue}{sub\_question2}="Does the contract mention the removal of content?"\\
\textcolor[rgb]{0.1333, 0.5451, 0.1333}{sub\_case2}="Pay attention to the guidance prompts in the text, accurately locate and remove the relevant parts of the content, and do not confuse them with other parts."\\
\textcolor[rgb]{1, 0, 0}{sub\_description2}="\textcolor[rgb]{0.11, 0.56, 1}{See the Removing your content section to learn why and how we might remove user-generated content from our services.}"\\
\textcolor[rgb]{1, 0.078, 0.57}{answer2}=LLM(sub\_case2,sub\_question2,sub\_description2)\\
\\
\textcolor{blue}{sub\_question3}="How is intellectual property infringement stipulated in the contract?"\\
\textcolor[rgb]{0.1333, 0.5451, 0.1333}{sub\_case3}="Pay attention to the relevant statements regarding the handling measures after intellectual property infringement, and understand the prescribed infringement handling process and methods."\\
\textcolor[rgb]{0.72, 0.52, 0.04}{\#Answering sub\_question3 requires relying on the answers of sub\_question1 and sub\_question2, namely answer1 and answer2}\\
\textcolor[rgb]{1, 0, 0}{sub\_description3}=\textcolor[rgb]{1, 0.078, 0.57}{answer1}+\textcolor[rgb]{1, 0.078, 0.57}{answer2}+"\textcolor[rgb]{0.11, 0.56, 1}{If you think that someone is infringing your intellectual property rights, you can send us notice of the infringement and well take appropriate action. For example, we suspend or close the Google Accounts of repeat copyright infringers as described in our Copyright Help Centre.}"\\
\textcolor[rgb]{1, 0.078, 0.57}{answer3}=LLM(sub\_case3,sub\_question3,sub\_description3)\\
\\
\textcolor{blue}{sub\_question4}="Will Google help me deal with situations where content is used without permission?"\\
\textcolor[rgb]{0.1333, 0.5451, 0.1333}{sub\_case4}="Pay attention to accurately understanding Google's measures for handling infringement in sub\_question3."\\
\textcolor[rgb]{0.72, 0.52, 0.04}{\#Answering sub\_question4 requires relying on the answers of sub\_question3, namely answer3}\\
\textcolor[rgb]{1, 0, 0}{sub\_description4}=\textcolor[rgb]{1, 0.078, 0.57}{answer3}\\
\textcolor[rgb]{1, 0.078, 0.57}{answer4}=LLM(sub\_case4,sub\_question4,sub\_description4)\\
\\
\textcolor[rgb]{0.72, 0.52, 0.04}{\#The final answer}\\
\textcolor[rgb]{0, 0.39, 0}{final\_answer}=\textcolor[rgb]{1, 0.078, 0.57}{answer4}\\
print(final\_answer)\\

    \hline
  \end{tabular}
  \caption{\label{LegalBench}
    The updated pseudo-program knowledge graph on LegalBench.  (Question:  Will Google help me if I think someone has taken and used content Ive created without my permission?
Contract: Some of our services give you the opportunity to make your content publicly available . For example, you might post a product or restaurant review that you wrote, or you might upload a blog post that you created.
See the Permission to use your content section for more about your rights in your content, and how your content is used in our services
See the Removing your content section to learn why and how we might remove user-generated content from our services
If you think that someone is infringing your intellectual property rights, you can send us notice of the infringement and well take appropriate action. For example, we suspend or close the Google Accounts of repeat copyright infringers as described in our Copyright Help Centre.)
  }
\end{table*}

\begin{table*}[t]
  \centering
  \small
  \begin{tabular}{p{\linewidth}}
    \hline
Let's go through the program step by step, answering each question based on the cases and descriptions provided:\\
\\
Step-by-step Execution:\\
sub\_question1\\
Question: Does the contract mention any content regarding usage permissions?\\
Case: "Pay attention to carefully reading the text and find the guiding statements related to content usage permissions."\\
Description: "See the Permission to use your content section for more about your rights in your content, and how your content is used in our services."\\
Analysis for sub\_question1: The description directly refers to a "Permission to use your content" section, which indicates that usage permissions are mentioned in the contract.\\
Answer1: "Yes, the contract mentions content usage permissions."\\
\\
sub\_question2\\
Question: Does the contract mention the removal of content?\\
Case: "Pay attention to the guidance prompts in the text, accurately locate and remove the relevant parts of the content, and do not confuse them with other parts."\\
Description: "See the Removing your content section to learn why and how we might remove user-generated content from our services."\\
Analysis for sub\_question2: The description references a "Removing your content" section, indicating that the contract does address the removal of content.\\
Answer2: "Yes, the contract mentions the removal of content."\\
\\
 sub\_question3\\
Question: How is intellectual property infringement stipulated in the contract?\\
Case: "Pay attention to the relevant statements regarding the handling measures after intellectual property infringement, and understand the prescribed infringement handling process and methods."\\
Description: This description relies on the answers to sub\_question1 and sub\_question2. Combining those with the following statement: "If you think that someone is infringing your intellectual property rights, you can send us notice of the infringement and we'll take appropriate action. For example, we suspend or close the Google Accounts of repeat copyright infringers as described in our Copyright Help Centre."\\
Analysis for sub\_question3: The description outlines the process for handling intellectual property infringement, including the option to report infringement and the potential actions (like suspension of accounts). This answers how intellectual property infringement is dealt with in the contract.\\
Answer3: "The contract specifies that intellectual property infringement is handled by allowing users to report infringement and take actions such as suspending or closing accounts of repeat offenders."\\
\\
 sub\_question4
Question: Will Google help me deal with situations where content is used without permission?\\
Case: "Pay attention to accurately understanding Google's measures for handling infringement in sub\_question3."\\
Description: This relies on answer3: "The contract specifies that intellectual property infringement is handled by allowing users to report infringement and take actions such as suspending or closing accounts of repeat offenders."\\
Analysis for sub\_question4: The description indicates that Google provides help in dealing with content usage without permission through its procedures for reporting and addressing infringement.\\
Answer4: "Yes, Google will help you deal with situations where content is used without permission by taking actions such as suspending or closing accounts of repeat infringers."\\
\\
Final Answer:\\
The final answer, based on the previous analysis, is:\\
"Yes, Google will help you deal with situations where content is used without permission by taking actions such as suspending or closing accounts of repeat infringers."\\

     \hline
  \end{tabular}
  \caption{\label{LegalBenchresult}
    The output of executing the pseudo-program knowledge graph in Table \ref{LegalBench}.
  }
\end{table*}

\begin{table*}[t]
  \centering
  \small
  \begin{tabular}{p{\linewidth}}
    \hline
\textcolor[rgb]{0.72, 0.52, 0.04}{\#Given a program text, your role is the LLM function,which has three parameters: case, question, and description. The description is textual, the question requires you to answer based on the description, and the case involves processed scenarios or supplementary information or related considerations to assist you in answering the question. You need to strictly follow the program logic to execute. During execution, the LLM function will be called multiple times, which means you will answer corresponding questions based on different cases and descriptions. Please output the final result of this program text in natural language.}\\
\\
def LLM(case,question,description):\\
    \quad LLM\_answer=\textcolor{orange}{Given the question, the case is an example about the question. Please study this example and answer the question based on the description.}\\
    \quad return LLM\_answer\\
\\
\textcolor{blue}{sub\_question1}="Is the description related to the question 'Can I still buy new energy vehicles'?"\\
\textcolor[rgb]{0.1333, 0.5451, 0.1333}{sub\_case1}="Market inquiry refers to inquiries about the overall situation of the securities market. This includes broad market indices like the Shanghai Composite Index or Shenzhen Component Index, which reflect the performance and conditions of the entire stock market."\\
\textcolor[rgb]{1, 0, 0}{sub\_description1}="\textcolor[rgb]{0.11, 0.56, 1}{Market inquiry}"\\
\textcolor[rgb]{1, 0.078, 0.57}{answer1}=LLM(sub\_case1,sub\_question1,sub\_description1)\\
\\
\textcolor{blue}{sub\_question2}="Is the description related to the question 'Can I still buy new energy vehicles'?"\\
\textcolor[rgb]{0.1333, 0.5451, 0.1333}{sub\_case2}="Industry sectors refer to the various parts of the securities market that are classified by industry, such as the financial sector, technology sector, new energy sector, etc. Industry sector inquiries mainly focus on the overall situation of a specific industry, including industry development trends, competition patterns within the industry, the impact of policies on the industry, and the industry's valuation in the market, among many other aspects."\\
\textcolor[rgb]{1, 0, 0}{sub\_description2}="\textcolor[rgb]{0.11, 0.56, 1}{Industry sector inquiry}"\\
\textcolor[rgb]{1, 0.078, 0.57}{answer2}=LLM(sub\_case2,sub\_question2,sub\_description2)\\
\\
\textcolor{blue}{sub\_question3}="Is the description related to the question 'Can I still buy new energy vehicles'?"\\
\textcolor[rgb]{0.1333, 0.5451, 0.1333}{sub\_case3}="Individual stock inquiry refers to inquiries related to the stocks of a specific company. This may include questions about the company’s financial health, performance, management changes, and stock price trends, as well as technical analysis related to stock price movements."\\
\textcolor[rgb]{1, 0, 0}{sub\_description3}="\textcolor[rgb]{0.11, 0.56, 1}{Individual stock inquiry}"\\
\textcolor[rgb]{1, 0.078, 0.57}{answer3}=LLM(sub\_case3,sub\_question3,sub\_description3)\\
\\
\textcolor{blue}{sub\_question4}="Is the description related to the question 'Can I still buy new energy vehicles'?"\\
\textcolor[rgb]{0.1333, 0.5451, 0.1333}{sub\_case4}="Fund inquiry pertains to inquiries about investment funds, which pool money from multiple investors for collective investment. This includes questions about the fund's type (e.g., equity, bond, or money market funds), investment strategy, performance, risk characteristics, management team, and subscription or redemption rules."\\
\textcolor[rgb]{1, 0, 0}{sub\_description4}="\textcolor[rgb]{0.11, 0.56, 1}{Fund inquiry}"\\
\textcolor[rgb]{1, 0.078, 0.57}{answer4}=LLM(sub\_case4,sub\_question4,sub\_description4)\\
\\
\textcolor{blue}{sub\_question5}="Is the description related to the question 'Can I still buy new energy vehicles'?"\\
\textcolor[rgb]{0.1333, 0.5451, 0.1333}{sub\_case5}="Customer service inquiry refers to inquiries made by investors to customer service departments of financial institutions regarding issues related to securities trading. These can involve problems with account operations (e.g., account opening, password retrieval), transaction rules (e.g., fees, time), product services, and handling of transaction disputes."\\
\textcolor[rgb]{1, 0, 0}{sub\_description5}="\textcolor[rgb]{0.11, 0.56, 1}{Customer service inquiry}"\\
\textcolor[rgb]{1, 0.078, 0.57}{answer5}=LLM(sub\_case5,sub\_question5,sub\_description5)\\
\\
\textcolor{blue}{sub\_question6}="Which one best fits the intention of the question 'Can I still buy new energy vehicles'?"\\
\textcolor[rgb]{0.1333, 0.5451, 0.1333}{sub\_case6}="Pay attention to comprehensively considering the previous analysis results to ensure accurate and error free judgment, without missing any key information."\\
\textcolor[rgb]{0.72, 0.52, 0.04}{\#Answering sub\_question6 requires relying on the answers of sub\_question1, sub\_question2, sub\_question3, sub\_question4, and sub\_question5, namely answer1, answer2, answer3, answer4 and answer5}\\
\textcolor[rgb]{1, 0, 0}{sub\_description6}=\textcolor[rgb]{1, 0.078, 0.57}{answer1}+\textcolor[rgb]{1, 0.078, 0.57}{answer2}+\textcolor[rgb]{1, 0.078, 0.57}{answer3}+\textcolor[rgb]{1, 0.078, 0.57}{answer4}+\textcolor[rgb]{1, 0.078, 0.57}{answer5}\\
\textcolor[rgb]{1, 0.078, 0.57}{answer6}=LLM(sub\_case6,sub\_question6,sub\_description6)\\
\\
\textcolor[rgb]{0.72, 0.52, 0.04}{\#The final answer}\\
\textcolor[rgb]{0, 0.39, 0}{final\_answer}=\textcolor[rgb]{1, 0.078, 0.57}{answer6}\\
print(final\_answer)\\

    \hline
  \end{tabular}
  \caption{\label{CFBenchmark}
    The updated pseudo-program knowledge graph on CFBenchmark.  (Question: You are an intention emotion assistant. Please analyze the intention of the following query: [market inquiry, industry sector inquiry, individual stock inquiry, fund inquiry, customer service inquiry]? Query: Can I still buy new energy vehicles?
  }
\end{table*}

\begin{table*}[t]
  \centering
  \small
  \begin{tabular}{p{\linewidth}}
    \hline
Let's go through the execution of the program step-by-step, based on the provided logic and descriptions:\\
\\
Step-by-step Execution:\\
sub\_question1\\
Question: Is the description related to the question 'Can I still buy new energy vehicles'?\\
Case: "Market inquiry refers to inquiries about the overall situation of the securities market. This includes broad market indices like the Shanghai Composite Index or Shenzhen Component Index, which reflect the performance and conditions of the entire stock market."\\
Description: "Market inquiry"\\
\\
Analysis for sub\_question1: The description focuses on market indices and the securities market, which is unrelated to the question about buying new energy vehicles. The question is focused on vehicle availability, not the stock market.\\
Answer1: "No, this description is not related to the question."\\
\\
sub\_question2\\
Question: Is the description related to the question 'Can I still buy new energy vehicles'?\\
Case: "Industry sectors refer to the various parts of the securities market that are classified by industry, such as the financial sector, technology sector, new energy sector, etc. Industry sector inquiries mainly focus on the overall situation of a specific industry, including industry development trends, competition patterns within the industry, the impact of policies on the industry, and the industry's valuation in the market, among many other aspects."\\
Description: "Industry sector inquiry"\\
Analysis for sub\_question2: The description discusses the new energy sector, which is relevant to the question about new energy vehicles. While the focus is on the broader industry trends, it still relates to the availability of vehicles in the new energy sector.\\
Answer2: "Yes, this description is related to the question."\\
\\
sub\_question3\\
Question: Is the description related to the question 'Can I still buy new energy vehicles'?\\
Case: "Individual stock inquiry refers to inquiries related to the stocks of a specific company. This may include questions about the company’s financial health, performance, management changes, and stock price trends, as well as technical analysis related to stock price movements."\\
Description: "Individual stock inquiry"\\
Analysis for sub\_question3: The description focuses on individual company stocks, which is unrelated to the question about buying new energy vehicles. The question pertains to vehicle availability, not individual stock performance.\\
Answer3: "No, this description is not related to the question."\\
\\
sub\_question4\\
Question: Is the description related to the question 'Can I still buy new energy vehicles'?\\
Case: "Fund inquiry pertains to inquiries about investment funds, which pool money from multiple investors for collective investment. This includes questions about the fund's type (e.g., equity, bond, or money market funds), investment strategy, performance, risk characteristics, management team, and subscription or redemption rules."\\
Description: "Fund inquiry"\\
\\
Analysis for sub\_question4: The description concerns investment funds, which is not directly relevant to the question about buying new energy vehicles. The question is focused on vehicles, not investment funds.\\
Answer4: "No, this description is not related to the question."\\
\\
sub\_question5\\
Question: Is the description related to the question 'Can I still buy new energy vehicles'?\\
Case: "Customer service inquiry refers to inquiries made by investors to customer service departments of financial institutions regarding issues related to securities trading. These can involve problems with account operations (e.g., account opening, password retrieval), transaction rules (e.g., fees, time), product services, and handling of transaction disputes."\\
Description: "Customer service inquiry"\\
Analysis for sub\_question5: The description is about customer service inquiries related to financial services, which is unrelated to the question about buying new energy vehicles.\\
Answer5: "No, this description is not related to the question."\\
\\
sub\_question6\\
Question: Which one best fits the intention of the question 'Can I still buy new energy vehicles'?\\
Case: "Pay attention to comprehensively considering the previous analysis results to ensure accurate and error-free judgment, without missing any key information."\\
Description: The description is a combination of the answers from sub\_question1, sub\_question2, sub\_question3, sub\_question4, and sub\_question5:\\
- "No, this description is not related to the question."\\
- "Yes, this description is related to the question."\\
- "No, this description is not related to the question."\\
- "No, this description is not related to the question."\\
- "No, this description is not related to the question."\\
Analysis for sub\_question6: The only description that is related to the question is from sub\_question2, which mentions the new energy sector. Therefore, this is the most relevant response.\\
Answer6: "The description in sub\_question2 best fits the intention of the question."\\
\\
Final Answer:\\
The description "Industry sector inquiry" in sub\_question2, which relates to the new energy sector, best fits the intention of the question "Can I still buy new energy vehicles?"\\

     \hline
  \end{tabular}
  \caption{\label{CFBenchmarkresult}
    The output of executing the pseudo-program knowledge graph in Table \ref{CFBenchmark}.
  }
\end{table*}

\begin{table*}[t]
  \centering
  \small
  \begin{tabular}{p{\linewidth}}
    \hline
\textcolor[rgb]{0.72, 0.52, 0.04}{\#Given a program text, your role is the LLM function,which has three parameters: case, question, and description. The description is textual, the question requires you to answer based on the description, and the case involves processed scenarios or supplementary information or related considerations to assist you in answering the question. You need to strictly follow the program logic to execute. During execution, the LLM function will be called multiple times, which means you will answer corresponding questions based on different cases and descriptions. Please output the final result of this program text in natural language.}\\
\\
def LLM(case,question,description):\\
    \quad LLM\_answer=\textcolor{orange}{Given the question, the case is an example about the question. Please study this example and answer the question based on the description.}\\
    \quad return LLM\_answer\\
\\
\textcolor{blue}{sub\_question1}="Is this description consistent with the expression 'Ancient Greek philosophers said that life without reflection is worth less'?"\\
\textcolor[rgb]{0.1333, 0.5451, 0.1333}{sub\_case1}="It implies that introspection is a prerequisite for a life to have value. The phrase "Only after" emphasizes that without introspection, life has no value. In other words, it presents a rather absolute condition where the value of life is directly tied to the act of introspection and doesn't consider the possibility of value in other contexts that don't involve this specific act."\\
\textcolor[rgb]{1, 0, 0}{sub\_description1}="\textcolor[rgb]{0.11, 0.56, 1}{Only after introspection can life be of value.}"\\
\textcolor[rgb]{1, 0.078, 0.57}{answer1}=LLM(sub\_case1,sub\_question1,sub\_description1)\\
\\
\textcolor{blue}{sub\_question2}="Is this description consistent with the expression 'Ancient Greek philosophers said that life without reflection is worth less'?"\\
\textcolor[rgb]{0.1333, 0.5451, 0.1333}{sub\_case2}=" In order for life to possess value, it is necessary to engage in the act of reflection regularly. The use of "To...we must..." indicates a causal or purpose - driven relationship. It emphasizes that reflection is an important activity that one needs to do in order to achieve a valuable life, without completely ruling out that there might be other factors also involved in a valuable life."\\
\textcolor[rgb]{1, 0, 0}{sub\_description2}="\textcolor[rgb]{0.11, 0.56, 1}{To be valuable in life, we must reflect on it from time to time.}"\\
\textcolor[rgb]{1, 0.078, 0.57}{answer2}=LLM(sub\_case2,sub\_question2,sub\_description2)\\
\\
\textcolor{blue}{sub\_question3}="Is this description consistent with the expression 'Ancient Greek philosophers said that life without reflection is worth less'?"\\
\textcolor[rgb]{0.1333, 0.5451, 0.1333}{sub\_case3}=" It simply means that a person remains in a state of confusion throughout their entire life. There is no mention of the concepts of reflection or the value of life, as the focus is solely on the state of being confused."\\
\textcolor[rgb]{1, 0, 0}{sub\_description3}="\textcolor[rgb]{0.11, 0.56, 1}{I'm confused for a lifetime.}"\\
\textcolor[rgb]{1, 0.078, 0.57}{answer3}=LLM(sub\_case3,sub\_question3,sub\_description3)\\
\\
\textcolor{blue}{sub\_question4}="Is this description consistent with the expression 'Ancient Greek philosophers said that life without reflection is worth less'?"\\
\textcolor[rgb]{0.1333, 0.5451, 0.1333}{sub\_case4}="This implies that the purpose of people's lives should be to gain understanding. It doesn't touch on the ideas of reflection and the worth of life as presented in the original maxim. Instead, it centers around the concept of living with the goal of understanding something, without any reference to the importance of reflecting on one's life for its value."\\
\textcolor[rgb]{1, 0, 0}{sub\_description4}="\textcolor[rgb]{0.11, 0.56, 1}{People should live to understand.}"\\
\textcolor[rgb]{1, 0.078, 0.57}{answer4}=LLM(sub\_case4,sub\_question4,sub\_description4)\\
\\
\textcolor{blue}{sub\_question5}="Which one best fits the expression 'Ancient Greek philosophers said that life without reflection is worth less'?"\\
\textcolor[rgb]{0.1333, 0.5451, 0.1333}{sub\_case5}="Pay attention to considering the previous analysis results comprehensively and select the best one. Ensure accurate judgment without missing any key information."\\
\textcolor[rgb]{0.72, 0.52, 0.04}{\#Answering sub\_question6 requires relying on the answers of sub\_question1, sub\_question2, sub\_question3,  and sub\_question4, namely answer1, answer2, answer3 and answer4} \\
\textcolor[rgb]{1, 0, 0}{sub\_description5}=\textcolor[rgb]{1, 0.078, 0.57}{answer1}+\textcolor[rgb]{1, 0.078, 0.57}{answer2}+\textcolor[rgb]{1, 0.078, 0.57}{answer3}+\textcolor[rgb]{1, 0.078, 0.57}{answer4}\\
\textcolor[rgb]{1, 0.078, 0.57}{answer5}=LLM(sub\_case6,sub\_question6,sub\_description6)\\
\\
\textcolor[rgb]{0.72, 0.52, 0.04}{\#The final answer}\\
\textcolor[rgb]{0, 0.39, 0}{final\_answer}=\textcolor[rgb]{1, 0.078, 0.57}{answer5}\\
print(final\_answer)\\

    \hline
  \end{tabular}
  \caption{\label{AGIEval}
    The updated pseudo-program knowledge graph on AGIEval. (Question:Ancient Greek philosophers said that life without reflection is worthless. Which of the following options is the closest to the meaning of this maxim?Answer Choices: (A) Only after introspection can life be of value. (B) To be valuable in life, we must reflect on it from time to time. (C) I'm confused for a lifetime. (D) People should live to understand.)
}
\end{table*}

\begin{table*}[t]
  \centering
  \small
  \begin{tabular}{p{\linewidth}}
    \hline
Let's go through the execution step by step based on the provided program logic. We are essentially evaluating a series of descriptions to see how well they align with the expression "Ancient Greek philosophers said that life without reflection is worth less." Then, we summarize the findings to determine which description fits best with that expression.
\\
Step-by-step Execution:\\
sub\_question1\\
Question: Is this description consistent with the expression 'Ancient Greek philosophers said that life without reflection is worth less'?\\
Case: "It implies that introspection is a prerequisite for a life to have value. The phrase 'Only after' emphasizes that without introspection, life has no value. In other words, it presents a rather absolute condition where the value of life is directly tied to the act of introspection and doesn't consider the possibility of value in other contexts that don't involve this specific act."\\
Description: "Only after introspection can life be of value."\\
\\
Analysis for sub\_question1: The description directly correlates with the notion that introspection (or reflection) is essential for life to have value, aligning closely with the expression about life being less valuable without reflection.\\
\\
Answer1: "Yes, this description is consistent with the expression."\\
\\
sub\_question2\\
Question: Is this description consistent with the expression 'Ancient Greek philosophers said that life without reflection is worth less'?\\
Case: "In order for life to possess value, it is necessary to engage in the act of reflection regularly. The use of 'To...we must...' indicates a causal or purpose-driven relationship. It emphasizes that reflection is an important activity that one needs to do in order to achieve a valuable life, without completely ruling out that there might be other factors also involved in a valuable life."\\
Description: "To be valuable in life, we must reflect on it from time to time."\\

Analysis for sub\_question2: The description suggests that reflection is a necessary but not exclusive condition for life to have value. This is slightly different from the original expression, which implies reflection as the sole or primary requirement for value.\\
\\
Answer2: "Yes, this description is consistent with the expression, but it introduces the possibility of other factors contributing to a valuable life."\\
\\
sub\_question3\\
Question: Is this description consistent with the expression 'Ancient Greek philosophers said that life without reflection is worth less'?\\
Case: "It simply means that a person remains in a state of confusion throughout their entire life. There is no mention of the concepts of reflection or the value of life, as the focus is solely on the state of being confused."\\
Description: "I'm confused for a lifetime."\\
Analysis for sub\_question3: The description focuses on confusion, with no mention of reflection or life’s value. It is not aligned with the original expression.\\
Answer3: "No, this description is not consistent with the expression."\\
\\
sub\_question4\\
Question: Is this description consistent with the expression 'Ancient Greek philosophers said that life without reflection is worth less'?\\
Case: "This implies that the purpose of people's lives should be to gain understanding. It doesn't touch on the ideas of reflection and the worth of life as presented in the original maxim. Instead, it centers around the concept of living with the goal of understanding something, without any reference to the importance of reflecting on one's life for its value."\\
Description: "People should live to understand."\\
Analysis for sub\_question4: The description speaks about understanding as the goal of life, but it doesn't emphasize reflection or introspection as the path to valuing life, making it inconsistent with the original expression.\\
Answer4: "No, this description is not consistent with the expression."\\
\\
sub\_question5\\
Question: Which one best fits the expression 'Ancient Greek philosophers said that life without reflection is worth less'?\\
Case: "Pay attention to considering the previous analysis results comprehensively and select the best one. Ensure accurate judgment without missing any key information."\\
Description: The description is a combination of the answers from sub\_question1, sub\_question2, sub\_question3, and sub\_question4:  "Yes, this description is consistent with the expression. Yes, this description is consistent with the expression, but it introduces the possibility of other factors contributing to a valuable life. No, this description is not consistent with the expression. No, this description is not consistent with the expression."\\
Analysis for sub\_question5: The best match comes from sub\_question1 and sub\_question2, which both emphasize the importance of reflection in a valuable life. Sub\_question1 suggests a direct, strong connection between reflection and life’s value, while sub\_question2 allows for other factors but still ties value to reflection.\\
Answer5: "The descriptions from sub\_question1 and sub\_question2 best fit the expression, but sub\_question1 aligns more directly with the original maxim."\\

Final Answer:\\
The description that best fits the expression "Ancient Greek philosophers said that life without reflection is worth less" is the one in sub\_question1, as it strongly links introspection to life's value without introducing any exceptions or other factors.\\

     \hline
  \end{tabular}
  \caption{\label{AGIEvalresult}
    The output of executing the pseudo-program knowledge graph in Table \ref{AGIEval}.
  }
\end{table*}

\begin{table*}[t]
  \centering
  \small
  \begin{tabular}{p{\linewidth}}
    \hline
    \textcolor[rgb]{0.72, 0.52, 0.04}{\#Given a program text, your role is the LLM function,which has three parameters: case, question, and description. The description is textual, the question requires you to answer based on the description, and the case involves processed scenarios or supplementary information or related considerations to assist you in answering the question. You need to strictly follow the program logic to execute. During execution, the LLM function will be called multiple times, which means you will answer corresponding questions based on different cases and descriptions. Please output the final result of this program text in natural language.}\\
    \\
\#include <iostream>\\
\#include <string>\\
using namespace std;\\
\\
string LLM(string case, string question, string description) \{\\
    \quad string LLM\_answer = \textcolor{orange}{Given the question, the case is an example about the question. Please study this example and answer the question based on the description};\\
    \quad return LLM\_answer;\\
\}\\
\\
int main() \{\\
    \quad string \textcolor{blue}{sub\_question1} = "How many lunch trays can the person carry at once?";\\
    \quad string \textcolor[rgb]{0.1333, 0.5451, 0.1333}{sub\_case1} = "Description:Tom is helping the restaurant staff collect dishes, but he can only carry 6 plates at a time.\\
    \quad Answer:Tom can carry 6 plates at a time.";\\
    \quad string \textcolor[rgb]{1, 0, 0}{sub\_description1} = "\textcolor[rgb]{0.11, 0.56, 1}{Roger can only carry 4 trays at a time.}";\\
    \quad string \textcolor[rgb]{1, 0.078, 0.57}{answer1} = LLM(sub\_case1, sub\_question1, sub\_description1);\\
\\
    \quad string \textcolor{blue}{sub\_question2} = "How many trays does the person need to pick up from the first table?";\\
    \quad string \textcolor[rgb]{0.1333, 0.5451, 0.1333}{sub\_case2} = "Description:Tom needs to pick up 15 plates from one table and 9 plates from another.\\
    \quad Answer:Tom needs to pick up 15 plates from the first table.";\\
    \quad string \textcolor[rgb]{1, 0, 0}{sub\_description2} = "\textcolor[rgb]{0.11, 0.56, 1}{Roger had to pick up 10 trays from one table and 2 trays from another.}";\\
    \quad string \textcolor[rgb]{1, 0.078, 0.57}{answer2} = LLM(sub\_case2, sub\_question2, sub\_description2);\\
\\
    \quad string \textcolor{blue}{sub\_question3} = "How many trays does the person need to pick up from the second table?";\\
    \quad string \textcolor[rgb]{0.1333, 0.5451, 0.1333}{sub\_case3} = "Description:Tom needs to pick up 15 plates from one table and 9 plates from another.\\
    \quad Answer:Tom needs to pick up 9 plates from the second table.";\\
    \quad string \textcolor[rgb]{1, 0, 0}{sub\_description3} = "\textcolor[rgb]{0.11, 0.56, 1}{Roger had to pick up 10 trays from one table and 2 trays from another.}";\\
    \quad string \textcolor[rgb]{1, 0.078, 0.57}{answer3} = LLM(sub\_case3, sub\_question3, sub\_description3);\\
\\
    \quad string \textcolor{blue}{sub\_question4} = "How many trays does the person need to carry in total?";\\
    \quad string \textcolor[rgb]{0.1333, 0.5451, 0.1333}{sub\_case4} = "Description:Tom needs to pick up 15 plates from the first table and 9 plates from the second table.\\
    \quad Answer: 15+9=24. Therefore Tom needs to carry 24 plates in total.";\\
    \textcolor[rgb]{0.72, 0.52, 0.04}{// Answering sub\_question4 requires relying on the answers of sub\_question2 and sub\_question3}\\
    \quad string \textcolor[rgb]{1, 0, 0}{sub\_description4} = \textcolor[rgb]{1, 0.078, 0.57}{answer2} + \textcolor[rgb]{1, 0.078, 0.57}{answer3};\\
    \quad string \textcolor[rgb]{1, 0.078, 0.57}{answer4} = LLM(sub\_case4, sub\_question4, sub\_description4);\\
\\
    \quad string \textcolor{blue}{sub\_question5} = "How many trips will the person need to make to carry all the trays?";\\
    \quad string \textcolor[rgb]{0.1333, 0.5451, 0.1333}{sub\_case5} = "Description:Tom can carry 6 plates at a time and Tom needs to carry 24 plates in total.\\
    \quad Answer: 24/6=4. Therefore Tom needs to carry 24 plates in total.";\\
    \textcolor[rgb]{0.72, 0.52, 0.04}{// Answering sub\_question5 requires relying on the answers of sub\_question1 and sub\_question4}\\
    \quad string \textcolor[rgb]{1, 0, 0}{sub\_description5} = \textcolor[rgb]{1, 0.078, 0.57}{answer1} + \textcolor[rgb]{1, 0.078, 0.57}{answer4};\\
    \quad string \textcolor[rgb]{1, 0.078, 0.57}{answer5} = LLM(sub\_case5, sub\_question5, sub\_description5);\\
\\
    \textcolor[rgb]{0.72, 0.52, 0.04}{// The final answer}\\
    \quad string \textcolor[rgb]{0, 0.39, 0}{final\_answer} = \textcolor[rgb]{1, 0.078, 0.57}{answer5};\\
    \quad cout << final\_answer << endl;\\
\\
    \quad return 0;\\
\}\\

    \hline
  \end{tabular}
  \caption{\label{C++}
    The updated pseudo-program knowledge graph in the form of \textbf{C++} on MultiArith. (Question: Roger was helping the cafeteria workers pick up lunch trays, but he could only carry 4 trays at a time. If he had to pick up 10 trays from one table and 2 trays from another, how many trips will he make?)
  }
\end{table*}

\begin{table*}[t]
  \centering
  \small
  \begin{tabular}{p{\linewidth}}
    \hline
Let's walk through the program to understand how the 'LLM' function is used to answer the questions and derive the final answer.\\
\\
 Breakdown of Execution:\\
\\
1. First Call to LLM (sub\_question1):\\
   - Question: How many lunch trays can the person carry at once?\\
   - Case: Tom can carry 6 plates at a time.\\
   - Description: Roger can carry 4 trays at a time.\\
   - Answer: Since Roger can carry 4 trays at a time, the answer is simply 4.\\
\\
   answer1 = "4"\\
\\
2. Second Call to LLM (sub\_question2):\\
   - Question: How many trays does the person need to pick up from the first table?\\
   - Case: Tom needs to pick up 15 plates from one table.\\
   - Description: Roger needs to pick up 10 trays from one table.\\
   - Answer: Roger needs to pick up 10 trays from the first table.\\
\\
   answer2 = "10"\\
\\
3. Third Call to LLM (sub\_question3):\\
   - Question: How many trays does the person need to pick up from the second table?\\
   - Case: Tom needs to pick up 9 plates from another table.\\
   - Description: Roger needs to pick up 2 trays from another table.\\
   - Answer: Roger needs to pick up 2 trays from the second table.\\
\\
   answer3 = "2"\\
\\
4. Fourth Call to LLM (sub\_question4):\\
   - Question: How many trays does the person need to carry in total?\\
   - Case: Tom needs to pick up 15 plates from one table and 9 plates from the second table.\\
   - Description: The total number of trays that Roger needs to carry is the sum of the trays picked up from both tables. From the previous answers:\\
     - 10 trays from the first table\\
     - 2 trays from the second table\\
     - Total = 10 + 2 = 12 trays\\
\\
   answer4 = "12"\\
\\
5. Fifth Call to LLM (sub\_question5):\\
   - Question: How many trips will the person need to make to carry all the trays?\\
   - Case: Tom can carry 6 plates at a time and needs to carry 24 plates in total.\\
   - Description: Roger can carry 4 trays at a time and needs to carry 12 trays in total. To calculate the number of trips, we divide the total number of trays by the number of trays he can carry at a time:\\
     - 12 / 4  = 3 trips\\
\\
   answer5 = "3"\\
\\
Final Answer: Roger needs to make 3 trips to carry all the trays.\\
\\
Output: 3\\

     \hline
  \end{tabular}
  \caption{\label{C++result}
    The output of executing the pseudo-program knowledge graph in Table \ref{C++}.
  }
\end{table*}

\begin{table*}[t]
  \centering
  \small
  \begin{tabular}{p{\linewidth}}
    \hline
    \textcolor[rgb]{0.72, 0.52, 0.04}{\#Given a program text, your role is the LLM function,which has three parameters: case, question, and description. The description is textual, the question requires you to answer based on the description, and the case involves processed scenarios or supplementary information or related considerations to assist you in answering the question. You need to strictly follow the program logic to execute. During execution, the LLM function will be called multiple times, which means you will answer corresponding questions based on different cases and descriptions. Please output the final result of this program text in natural language.}\\
    \\
public class Main \{\\
\\
    \quad public static String LLM(String case, String question, String description) \{\\
        \quad \quad String LLM\_answer = \textcolor{orange}{Given the question, the case is an example about the question. Please study this example and answer the question based on the description;}\\
        \quad \quad return LLM\_answer;\\
   \quad \}\\
\\
    \quad public static void main(String[] args) \{\\
\\
        \quad String \textcolor{blue}{sub\_question1} = "How many lunch trays can the person carry at once?";\\
        \quad String \textcolor[rgb]{0.1333, 0.5451, 0.1333}{sub\_case1} = "Description:Tom is helping the restaurant staff collect dishes, but he can only carry 6 plates at a time.\\
        \quad Answer:Tom can carry 6 plates at a time.";\\
        \quad String \textcolor[rgb]{1, 0, 0}{sub\_description1} = "\textcolor[rgb]{0.11, 0.56, 1}{Roger can only carry 4 trays at a time.}";\\
        \quad String \textcolor[rgb]{1, 0.078, 0.57}{answer1} = LLM(sub\_case1, sub\_question1, sub\_description1);\\
\\
        \quad String \textcolor{blue}{sub\_question2} = "How many trays does the person need to pick up from the first table?";\\
        \quad String \textcolor[rgb]{0.1333, 0.5451, 0.1333}{sub\_case2} = "Description:Tom needs to pick up 15 plates from one table and 9 plates from another.\\
        \quad Answer:Tom needs to pick up 15 plates from the first table.";\\
        \quad String \textcolor[rgb]{1, 0, 0}{sub\_description2} = "\textcolor[rgb]{0.11, 0.56, 1}{Roger had to pick up 10 trays from one table and 2 trays from another.}";\\
        \quad String \textcolor[rgb]{1, 0.078, 0.57}{answer2} = LLM(sub\_case2, sub\_question2, sub\_description2);\\
\\
        \quad String \textcolor{blue}{sub\_question3} = "How many trays does the person need to pick up from the second table?";\\
        \quad String \textcolor[rgb]{0.1333, 0.5451, 0.1333}{sub\_case3} = "Description:Tom needs to pick up 15 plates from one table and 9 plates from another.\\
        \quad Answer:Tom needs to pick up 9 plates from the second table.";\\
        \quad String \textcolor[rgb]{1, 0, 0}{sub\_description3} = "\textcolor[rgb]{0.11, 0.56, 1}{Roger had to pick up 10 trays from one table and 2 trays from another.}";\\
        \quad String \textcolor[rgb]{1, 0.078, 0.57}{answer3} = LLM(sub\_case3, sub\_question3, sub\_description3);\\
\\
        \quad String \textcolor{blue}{sub\_question4} = "How many trays does the person need to carry in total?";\\
        \quad String \textcolor[rgb]{0.1333, 0.5451, 0.1333}{sub\_case4} = "Description:Tom needs to pick up 15 plates from the first table and 9 plates from the second table.\\
        \quad Answer: 15+9=24. Therefore Tom needs to carry 24 plates in total.";\\
        \quad \textcolor[rgb]{0.72, 0.52, 0.04}{// Answering sub\_question4 requires relying on the answers of sub\_question2 and sub\_question3}\\
        \quad String \textcolor[rgb]{1, 0, 0}{sub\_description4} = \textcolor[rgb]{1, 0.078, 0.57}{answer2} + \textcolor[rgb]{1, 0.078, 0.57}{answer3};\\
        \quad String \textcolor[rgb]{1, 0.078, 0.57}{answer4} = LLM(sub\_case4, sub\_question4, sub\_description4);\\
\\
        \quad String \textcolor{blue}{sub\_question5} = "How many trips will the person need to make to carry all the trays?";\\
        \quad String \textcolor[rgb]{0.1333, 0.5451, 0.1333}{sub\_case5} = "Description:Tom can carry 6 plates at a time and Tom needs to carry 24 plates in total.\\
        \quad Answer: 24/6=4. Therefore Tom needs to carry 24 plates in total.";\\
        \textcolor[rgb]{0.72, 0.52, 0.04}{// Answering sub\_question5 requires relying on the answers of sub\_question1 and sub\_question4}\\
        \quad String \textcolor[rgb]{1, 0, 0}{sub\_description5} = \textcolor[rgb]{1, 0.078, 0.57}{answer1} + \textcolor[rgb]{1, 0.078, 0.57}{answer4};\\
        \quad String \textcolor[rgb]{1, 0.078, 0.57}{answer5} = LLM(sub\_case5, sub\_question5, sub\_description5);\\
\\
        \textcolor[rgb]{0.72, 0.52, 0.04}{// The final answer}\\
        \quad String string \textcolor[rgb]{0, 0.39, 0}{final\_answer} = \textcolor[rgb]{1, 0.078, 0.57}{answer5};\\
\\
        \quad System.out.println(final\_answer);\\
    \quad \}\\
\}\\

    \hline
  \end{tabular}
  \caption{\label{Java}
    The updated pseudo-program knowledge graph in the form of \textbf{Java} on MultiArith. (Question: Roger was helping the cafeteria workers pick up lunch trays, but he could only carry 4 trays at a time. If he had to pick up 10 trays from one table and 2 trays from another, how many trips will he make?)
  }
\end{table*}

\begin{table*}[t]
  \centering
  \small
  \begin{tabular}{p{\linewidth}}
    \hline
Let's break down the program step-by-step and understand how the 'LLM'function is used to answer the questions and derive the final result.
\\
Breakdown of Execution:\\
\\
1. First Call to LLM (sub\_question1):\\
   - Question: How many lunch trays can the person carry at once?\\
   - Case: Tom can carry 6 plates at a time.\\
   - Description: Roger can carry 4 trays at a time.\\
   - Answer: Based on the description, Roger can carry 4 trays at a time.\\
   \\
   So, answer1 = "4 trays"\\
\\
2. Second Call to LLM (sub\_question2):\\
   - Question: How many trays does the person need to pick up from the first table?\\
   - Case: Tom needs to pick up 15 plates from one table.\\
   - Description: Roger needs to pick up 10 trays from one table and 2 trays from another.\\
   - Answer: Roger needs to pick up 10 trays from the first table.\\
   \\
   So, answer2 = "10 trays"\\
\\
3. Third Call to LLM (sub\_question3):\\
   - Question: How many trays does the person need to pick up from the second table?\\
   - Case: Tom needs to pick up 15 plates from one table and 9 plates from another.\\
   - Description: Roger needs to pick up 10 trays from one table and 2 trays from another.\\
   - Answer: Roger needs to pick up 2 trays from the second table.\\
   \\
   So, answer3 = "2 trays"\\
\\
4. Fourth Call to LLM (sub\_question4):\\
   - Question: How many trays does the person need to carry in total?\\
   - Case: Tom needs to pick up 15 plates from the first table and 9 plates from the second table.\\
   - Description: By adding up the trays from the first and second tables, we get the total number of trays Roger needs to carry:\\
     - 10 trays from the first table\\
     - 2 trays from the second table\\
     - Total = 10 + 2 = 12 trays\\
\\
   So, answer4 = "12 trays"\\
\\
5. Fifth Call to LLM (sub\_question5):\\
   - Question: How many trips will the person need to make to carry all the trays?\\
   - Case: Tom can carry 6 plates at a time and needs to carry 24 plates in total.\\
   - Description: Roger can carry 4 trays at a time and needs to carry 12 trays in total. To calculate the number of trips, we divide the total number of trays by the number of trays he can carry at a time:\\
     - 12 / 4  = 3 trips\\
\\
   So, answer5 = "3 trips"\\
\\
 Final Answer: Roger needs to make 3 trips to carry all the trays.\\
\\
Output: 3 trips\\
     \hline
  \end{tabular}
  \caption{\label{Javaresult}
    The output of executing the pseudo-program knowledge graph in Table \ref{Java}.
  }
\end{table*}

\begin{table*}[t]
  \centering
  \small
  \begin{tabular}{p{\linewidth}}
    \hline
You are tasked with generating new questions that share the same underlying reasoning logic as a given original question but differ in content. Follow these steps carefully:\\

1. \textbf{Input} : You will be provided with an original question from an open-source dataset.\\
2. \textbf{Task} : For each original question, generate \textbf{4 new questions} that meet the following criteria:
\begin{itemize}
    \item \vspace{-1em} \textbf{Same Reasoning Logic}: The new questions must require the same reasoning steps or problem-solving approach as the original.
    \item \vspace{-1em}\textbf{Different Content}: The surface content (e.g., entities, scenarios, or context) should be altered to create distinct variations.
    \item \vspace{-1em}\textbf{High Quality}: Ensure the new questions are clear, grammatically correct, and logically valid.
\end{itemize}
\vspace{-1em} 3. \textbf{Output Format}: Return the generated questions in this structured format: \\
\quad \quad Original Question: [Insert original question here]  \\
\quad \quad New Questions:  \\
\quad \quad a. [Generated question 1]  \\
\quad \quad b. [Generated question 2]  \\
\quad \quad c. [Generated question 3]  \\
\quad \quad d. [Generated question 4]  \\

4. \textbf{Example}:\\
\quad \quad Original Question: "If a train travels 300 miles in 5 hours, what is its average speed?"\\
\quad \quad New Questions:\\

\quad \quad a. "If a car travels 240 kilometers in 4 hours, what is its average speed?"\\

\quad \quad b. "A cyclist covers 45 miles in 1.5 hours. What is their average speed?"\\

\quad \quad c. "A plane flies 1,800 miles in 6 hours. Calculate its average speed."\\

\quad \quad d. "A runner completes a 10-kilometer race in 50 minutes. What is their average speed in km/h?"
  \\   \hline
  \end{tabular}
  \caption{\label{prompt}
    The prompt for using LLMs to assist in generating datasets.
  }
\end{table*}
\end{document}